\pdfoutput=1
\documentclass[10pt,onecolumn,times]{article}

\usepackage[pagenumbers]{cvpr} 

\definecolor{cvprblue}{rgb}{0.21,0.49,0.74}
\usepackage[pagebackref,breaklinks,colorlinks,allcolors=cvprblue]{hyperref}

\def\paperID{9529} 
\def\confName{CVPR}
\def\confYear{2025}
\usepackage{hyperref}
\usepackage{url}
\usepackage[utf8]{inputenc} 
\usepackage[T1]{fontenc}    
\usepackage{booktabs}       
\usepackage{amsfonts}       
\usepackage{nicefrac}       
\usepackage{microtype}      

\usepackage{amsmath}
\usepackage{amssymb}
\usepackage{mathtools}
\usepackage{amsthm}

\usepackage[capitalize,noabbrev]{cleveref}

\usepackage{thm-restate}

\theoremstyle{plain}

\theoremstyle{definition}

\theoremstyle{remark}

\usepackage[textsize=tiny]{todonotes}
\usepackage{adjustbox}
\usepackage{arydshln}
\usepackage{booktabs}
\usepackage{siunitx}
\usepackage{makecell}
\usepackage{algorithm}
\usepackage[noend]{algpseudocode}
\usepackage{wrapfig}
\usepackage{pifont} 
\usepackage{colortbl} 
\usepackage{verbatim}
\usepackage{multirow}
\usepackage{multicol}
\usepackage{graphicx}
\usepackage{booktabs, multirow} 
\usepackage{soul}
\usepackage{changepage,threeparttable} 
\usepackage{bm}

\definecolor{tbgray}{gray}{.92}

\newcommand{\name}[1]{\textsc{MMVideo}}

\usepackage{tcolorbox}

\definecolor{deepblue}{RGB}{33, 150, 243}
\definecolor{lightgray}{RGB}{240, 240, 245}
\definecolor{highlightblue}{RGB}{0, 123, 167}
\definecolor{darkgreen}{RGB}{34, 139, 34}
\definecolor{lightcream}{RGB}{255, 252, 240}
\definecolor{darkcream}{RGB}{245, 245, 220}

\title{STIV: Scalable Text and Image Conditioned Video Generation}

\author{Zongyu Lin$^{1}$$^\star$\thanks{This work was done during an internship at Apple. \\ \quad $^{1}$First authors \quad $^{2}$Core authors \quad $^\diamond$Senior authors} \and
Wei Liu$^{1}$$^\star$ \and
Chen Chen$^{2}$$^\star$ \and
Jiasen Lu$^{2}$$^\star$ \and 
Wenze Hu$^{2}$$^\star$ \and
Tsu-Jui Fu$^{2}$$^\star$ \and
Jesse Allardice$^{2}$$^\star$ \and 
Zhengfeng Lai$^{2}$$^\star$ \and
Liangchen Song$^{2}$$^\star$ \and
Bowen Zhang$^{2}$$^\star$ \and
Cha Chen$^{2}$$^\star$ \and
Yiran Fei$^\star$ \and 
Lezhi Li$^\star$ \and
\quad Yizhou Sun$^\diamond$$^\dagger$ \quad Kai-Wei Chang$^\diamond$$^\dagger$ \quad Yinfei Yang$^\diamond$$^\star$\\[1em]
$^\star$Apple \quad $^\dagger$University of California, Los Angeles
}

\begin{document}
\maketitle
\begin{abstract}
The field of video generation has made remarkable advancements, yet there remains a pressing need for a clear, systematic recipe that can guide the development of robust and scalable models. In this work, we present a comprehensive study that systematically explores the interplay of model architectures, training recipes, and data curation strategies, culminating in a simple and scalable text-image-conditioned video generation method, named STIV.
Our framework integrates image condition into a Diffusion Transformer (DiT) through frame replacement, while incorporating text conditioning via a joint image-text conditional classifier-free guidance. This design enables STIV to perform both text-to-video (T2V) and text-image-to-video (TI2V) tasks simultaneously. Additionally, STIV can be easily extended to various applications, such as video prediction, frame interpolation, multi-view generation, and long video generation, etc. 
With comprehensive ablation studies on T2I, T2V, and TI2V, STIV demonstrate strong performance, despite its simple design. An 8.7B model with \(512^2\) resolution achieves 83.1 on VBench T2V, surpassing both leading open and closed-source models like CogVideoX-5B, Pika, Kling, and Gen-3. The same-sized model also achieves a state-of-the-art result of 90.1 on VBench I2V task at \(512^2\) resolution. By providing a transparent and extensible recipe for building cutting-edge video generation models, we aim to empower future research and accelerate progress toward more versatile and reliable video generation solutions.
\end{abstract}
\section{Introduction}
\label{sec:introduction}

The field of video generation has witnessed a significant progress with the introduction of Sora~\citep{openai2024sora}, a video generation model based on Diffusion Transformer (DiT)~\citep{peebles2023scalable} architecture. Researchers have been actively exploring optimal methods to incorporate text and other conditions into the DiT architecture. For example, PixArt-$\alpha$~\citep{chen2023pixart} leverages cross attention, while SD3~\citep{esser2024scaling} concatenates text with the noised patches and applies self-attention using the MMDiT block. Several video generation models~\citep{yang2024cog-video-x, polyak2024movie, gao2024lumina} adopt similar approaches and have made substantial progress in the text-to-video (T2V) task. Pure T2V approaches often struggle with producing coherent and realistic videos, as their outputs are not grounded in external references or contextual constraints~\citep{chu2024sora}. To address this limitation, text-image-to-video (TI2V) introduce an initial image frame along with the textual prompt, providing a more concrete grounding for the generated video.

\begin{wrapfigure}{r}{0.48\textwidth}
  \begin{center}
    \vspace{-0.4cm}
    \includegraphics[width=.48\textwidth]{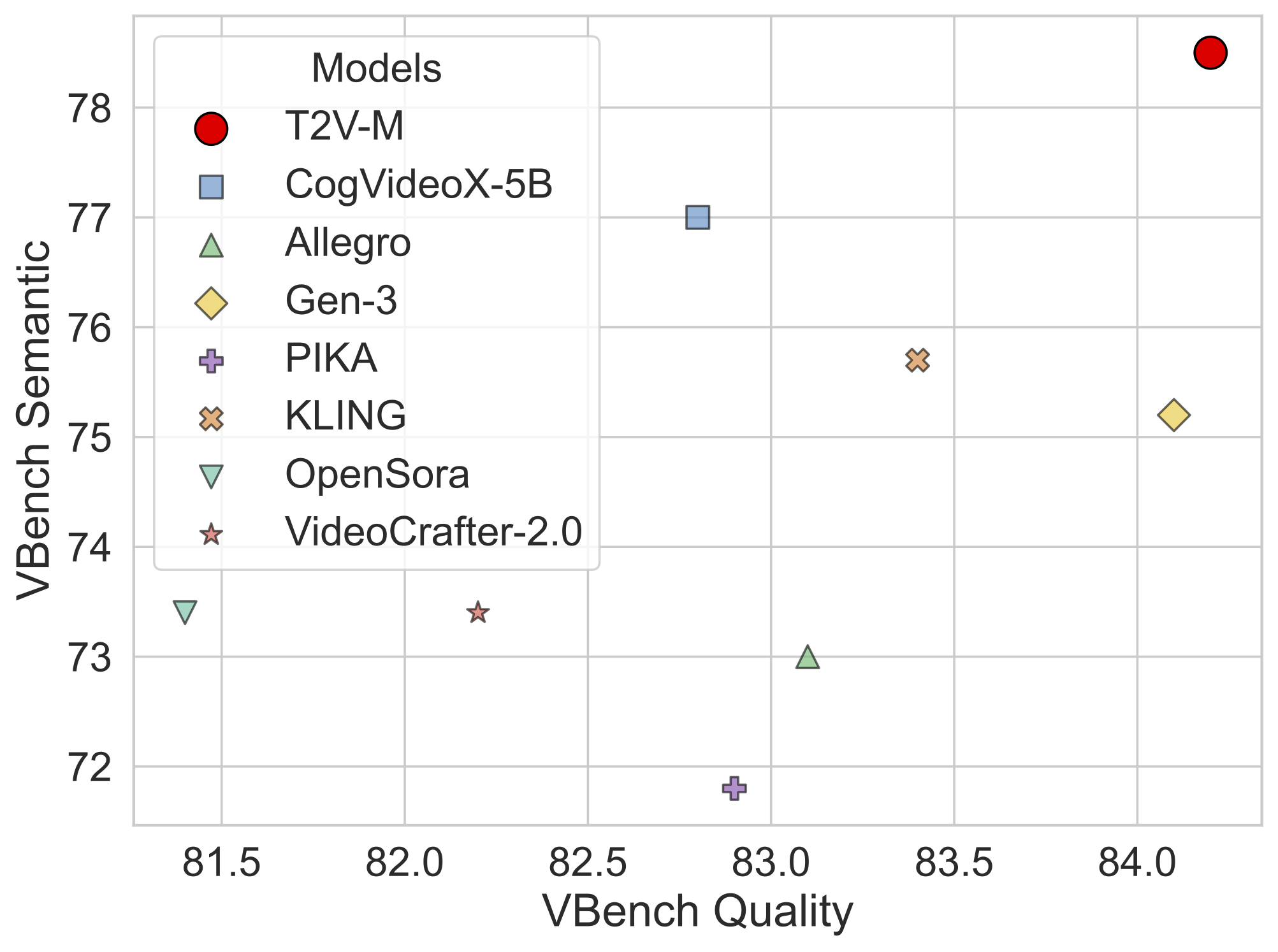}
  \end{center}
  \vspace{-0.7cm}
  \caption{Performance comparison of our Text-to-Video model against both open-source and closed-source state-of-the-art models on VBench~\cite{huang2024vbench}.}
  \label{fig:combined_figures}
\end{wrapfigure}
Despite substantial progress in video generation, achieving Sora-level performance for T2V and TI2V remains challenging.
A central challenge is how to seamlessly integrate image-based conditions into the DiT architecture, calling for innovative techniques blend visual inputs smoothly with textual cues.
Meanwhile, there is a pressing need for stable, efficient large-scale training strategies, as well as improving the overall quality of training datasets. 
To address these issues, a comprehensive, step-by-step ``recipe'' would greatly assist in developing unified models that handle both T2V and TI2V task under one framework. 
Overcoming these challenges is essential for advancing the field and fully realizing the potential of video generation models.

Although various studies~\cite{xing2023dynamicrafter, blattmann2023stable, chen2023videocrafter1, chen2023seine, ren2024consisti2v, zhang2023i2vgen, dai2023animateanything} have examined methods of integrating image conditions into the U-Net architectures, how to effectively incorporate such conditions into the DiT architecture remains unsolved. 
Moreover, existing studies in video generation often focuses on individual aspects independently, overlooking the how their collective impact on overall performance. 
For instance, while stability tricks like QK-norm~\citep{henry2020qk-norm,esser2024scaling} have been introduced, they prove insufficient as models scale to larger sizes~\citep{team2024chameleon}, and no existing approach has successfully unified T2V and TI2V capabilities within a single model. This lack of systematic, holistic research limits progress toward more efficient and versatile video generation solutions.

In this work, we first present a comprehensive study of model architectures and training strategies to establish a robust foundation for T2V.  
Our analysis reveals three key insights: 
(1) stability techniques such as QK-norm and sandwich-norm~\citep{gong2022sandwich-norm,ding2021cogview} are critical for effectively scaling larger video generation models; 
(2) employing factorized spatial-temporal attention~\cite{bertasius2021space}, MaskDiT~\cite{zheng2023fast}, and switching to AdaFactor~\cite{shazeer2018adafactor} significantly improve training efficiency and reduce memory usage with minimal impact on performance loss; 
(3) progressive training, where spatial and temporal layers are initialized from separate models, outperforms using a single model under the same compute constraints. Starting from a PixArt-$\alpha$ baseline architecture, we address scaling challenges with these stability and efficiency measures, and further enhance performance with Flow Matching~\citep{ma2024sit}, RoPE~\citep{su2024roformer}, and micro conditions~\citep{podell2023sdxl}. As a result, our largest T2V model (8.7B parameters) achieves state-of-the-art semantic alignment and a VBench score of 83.1.

\begin{figure*}[htbp]
    \centering
    \begin{minipage}{\textwidth}
        \centering
        \fcolorbox{darkgreen}{lightcream}{
            \parbox{\textwidth}{
                \centering \textbf{\textcolor{darkgreen}{Text-to-Video}}
            }
        } \\[1.0em]

        \textbf{Prompt:} \textit{An adorable kangaroo wearing blue jeans and a white t shirt taking a pleasant stroll in Johannesburg South Africa during a beautiful sunset.} \\[0.5em]
        \includegraphics[width=0.22\textwidth]{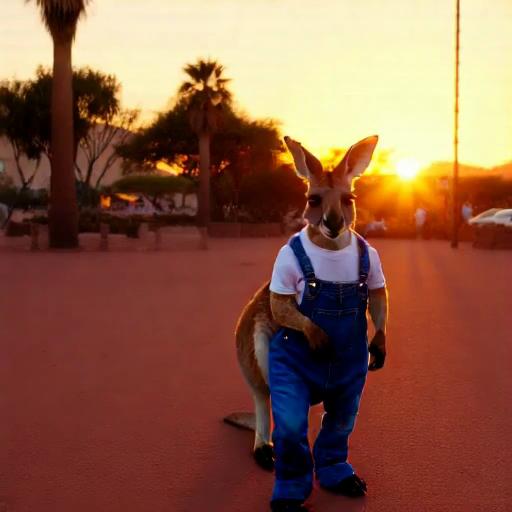}
        \includegraphics[width=0.22\textwidth]{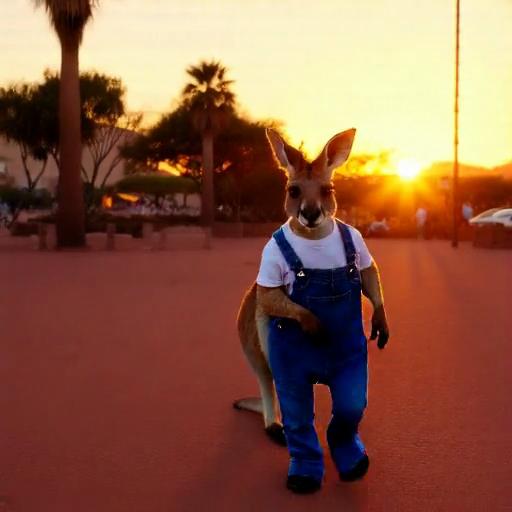}
        \includegraphics[width=0.22\textwidth]{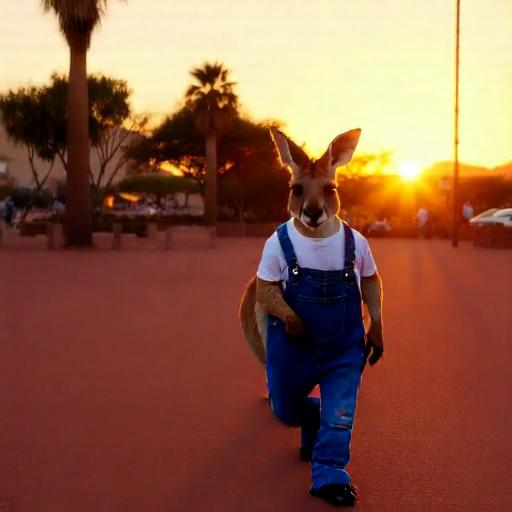}
        \includegraphics[width=0.22\textwidth]{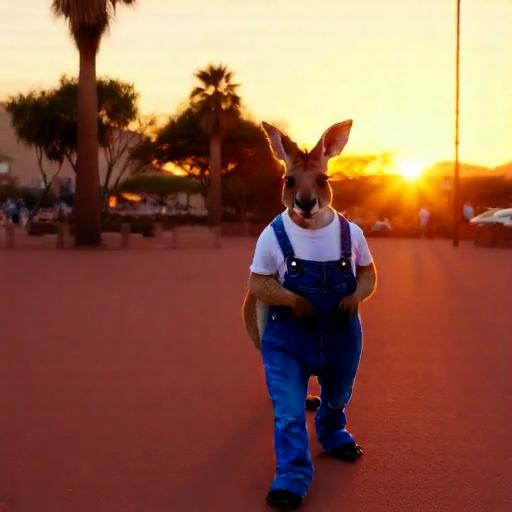}
        \\[1em]
        \textbf{Prompt:} \textit{A swan with wings tipped in gold gliding across a misty lake, leaving a trail of soft, shimmering light that fades as the sun rises.} \\[0.5em]
        \includegraphics[width=0.21\textwidth]{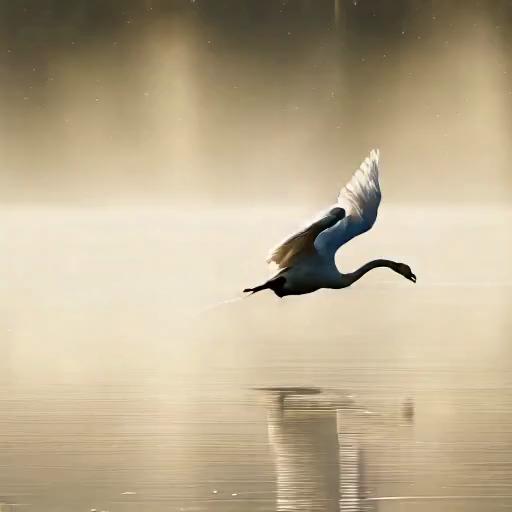}
        \includegraphics[width=0.21\textwidth]{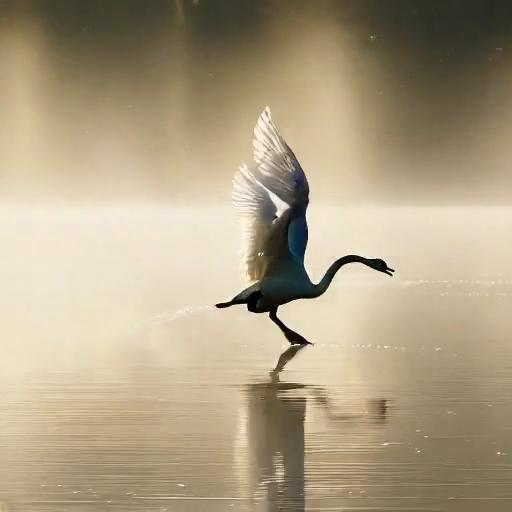}
        \includegraphics[width=0.21\textwidth]{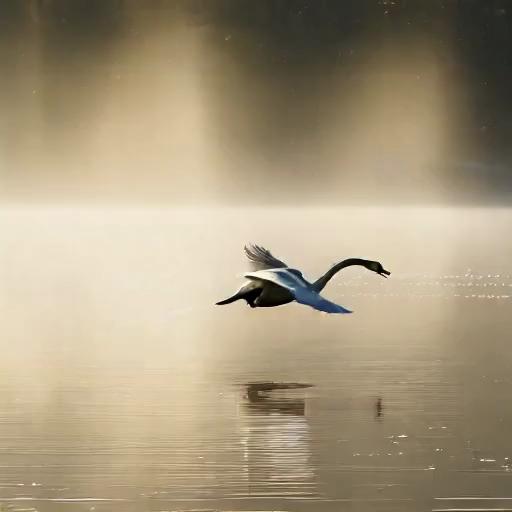}
        \includegraphics[width=0.21\textwidth]{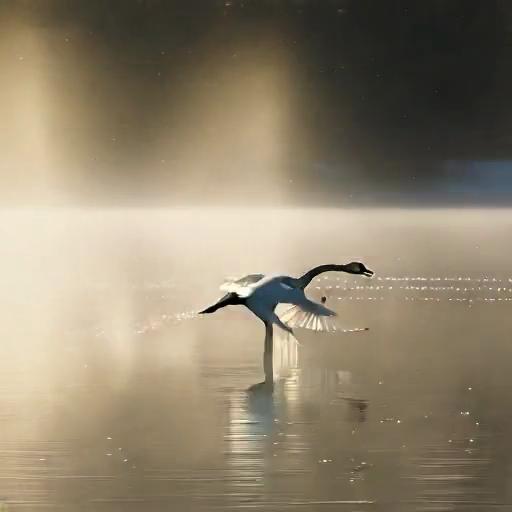}
        \\[1.0em]

    \end{minipage}
    \\[1.5em]
    \centering
    \fcolorbox{darkgreen}{lightcream}{
        \parbox{\textwidth}{
            \centering \textbf{\textcolor{darkgreen}{Text-Image-to-Video}}
        }
    }
    \\[1.0em]


    \centering
    \textbf{Prompt:} \textit{The video presents a sequence of frames that depict a space scene with a large, green and yellow planet at the center, surrounded by smaller celestial bodies. The background is a deep blue, speckled with stars.} \\[0.5em]
    \begin{minipage}{\textwidth}
        \centering
        \begin{minipage}[t]{0.18\textwidth}
            \centering
            \fcolorbox{blue}{white}{\includegraphics[width=\linewidth]{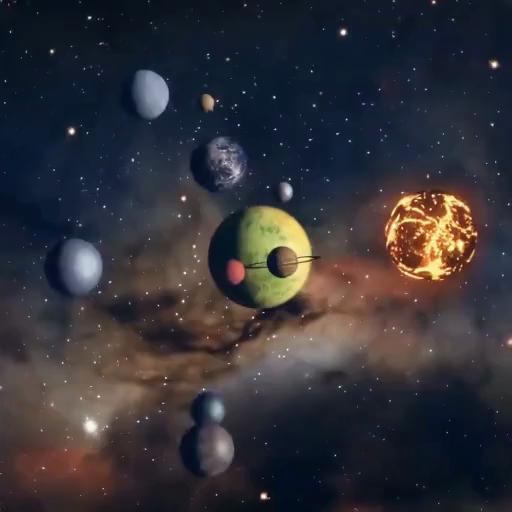}}
            \\[0.5em]
        \end{minipage}
        \hfill
        \includegraphics[width=0.18\textwidth]{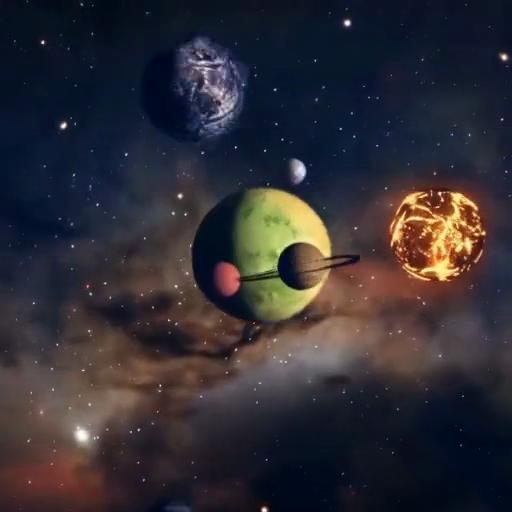}
        \includegraphics[width=0.18\textwidth]{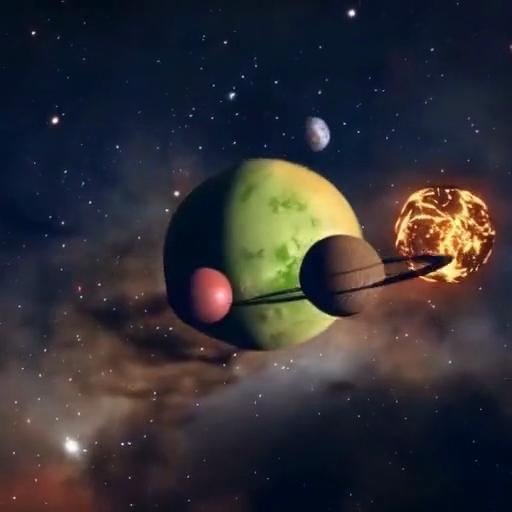}
        \includegraphics[width=0.18\textwidth]{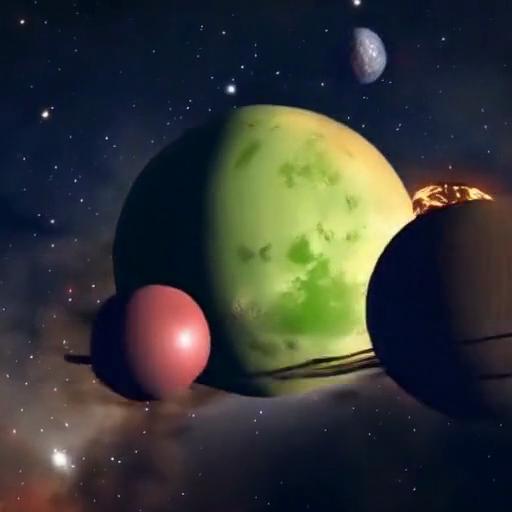}
        \includegraphics[width=0.18\textwidth]{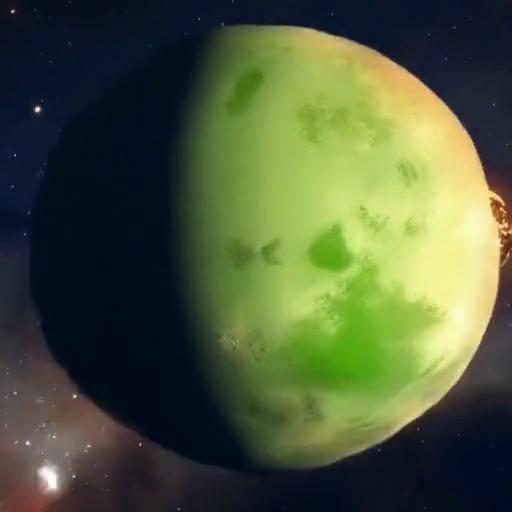}
    \end{minipage}
    \\[1.0em]

    \centering
    \textbf{Prompt:} \textit{Robots move efficiently through a futuristic laboratory, adjusting holographic displays and conducting experiments, while scientists observe and interact with the high-tech equipment.} \\[0.5em]
    \begin{minipage}{\textwidth}
        \centering
        \begin{minipage}[t]{0.18\textwidth}
            \centering
            \fcolorbox{blue}{white}{\includegraphics[width=\linewidth]{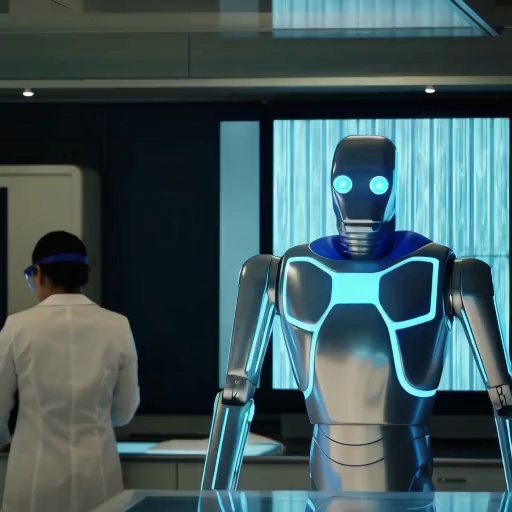}}
            \\[0.5em]
        \end{minipage}
        \hfill
        \includegraphics[width=0.18\textwidth]{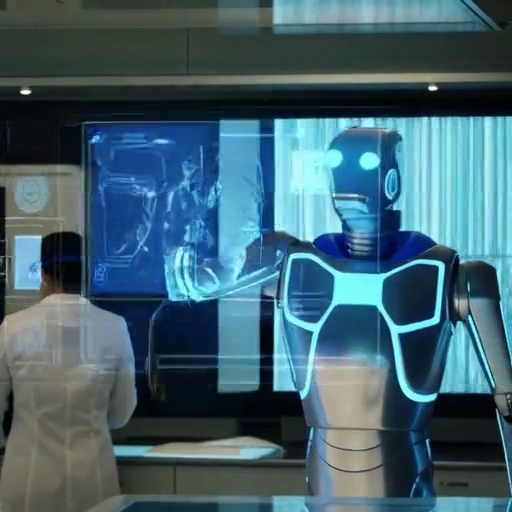}
        \includegraphics[width=0.18\textwidth]{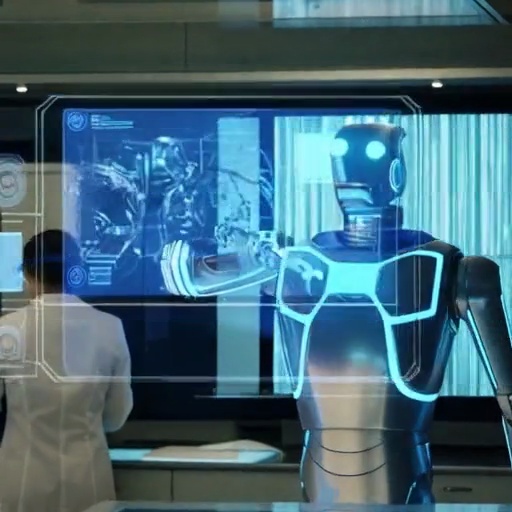}
        \includegraphics[width=0.18\textwidth]{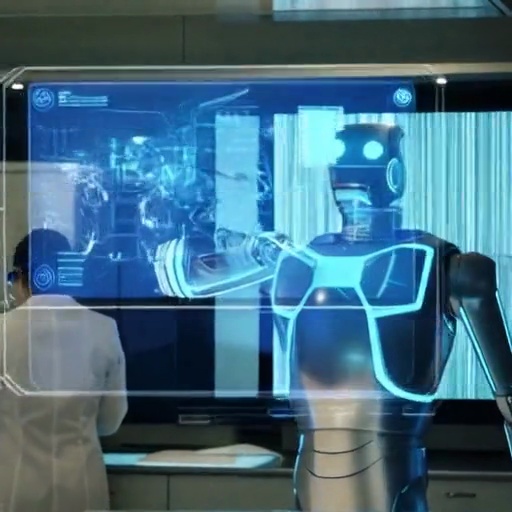}
        \includegraphics[width=0.18\textwidth]{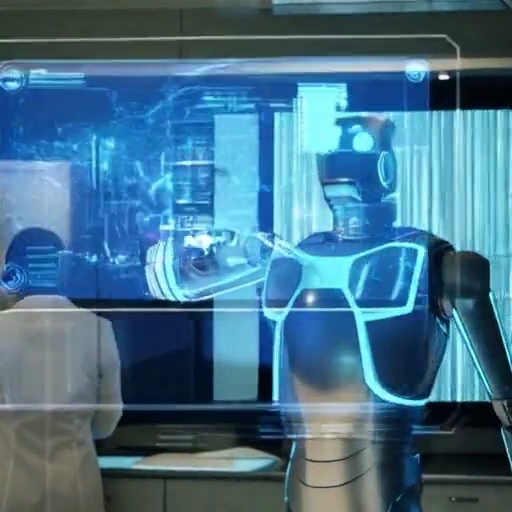}
    \end{minipage}
    \vspace{-0.5em}
    \caption{Text-to-Video and Text-Image-to-Video generation samples by T2V and STIV models. The text prompts and first frame image conditions are borrowed from Sora's demos~\cite{openai2024sora} and MovieGenBench~\cite{polyak2024movie}.}
    \label{fig:demo_intro}
\end{figure*}

We then identify the optimal model architecture and hyperparameters established in the T2V setting and apply them to the TI2V task. 
Our results show that simply replacing the first noised latent frame with the un-noised image condition latent yields strong performance. 
Although ConsistI2V~\cite{ren2024consisti2v} introduced a similar idea in a U-Net setting, it required spatial self-attention for each frame and window-based temporal self-attention to match our quality.
In contrast, the DiT architecture natively propagates the image-conditioned first frame through stacked spatial-temporal attention layers, eliminating the need for these additional operations.
However, as we scale up spatial resolution, we observe the model producing slow or nearly static motion. To solve this, we introduce random dropout of the image condition during training and apply joint image-text conditional classifier-free guidance (JIT-CFG) for both text and image conditions during inference. This strategy resolves the motion issue and also enables a single model to excel at both T2V and TI2V tasks.

With all these changes, we finalize our model and scale it up from 600M to 8.7B parameters. Our best STIV model achieves a state-of-the-art result of 90.1 in the VBench I2V task at $512^2$ resolution.
Beyond enhancing video generation quality, we demonstrate the potential of extending our framework to various downstream applications, including video prediction, frame interpolation, multi-view generation and long video generation. These results validate the scalability and versatility of our approach, showcasing its ability to address diverse video generation challenges. We summarize our contributions as follows:
\begin{itemize}
    \item We present STIV, a single model capable of performing both T2V and TI2V tasks. At its core, we replace the noised latent with the un-noised image condition latent and introduce joint image-text conditioned CFG.
    \item We conduct a systematic study for T2I, T2V and TI2V, covering model architectures, efficient and stable training techniques, and progressive training recipes to scale up the model size, spatial resolution, and duration.
    \item These design features make it easy to train and adaptable to various tasks, including video prediction, frame interpolation, and long video generation.
    \item Our experiments include detailed ablation studies on different design choices and hyperparameters, evaluated on VBench, VBench-I2V and MSRVTT. These studies demonstrate the effectiveness of the proposed model compared with a range of recent state-of-the-art open-source and closed-source video generation models. Some of the generated videos are shown in Fig.~\ref{fig:demo_intro}. More examples can be found in the Sec.~\ref{sec:more_examples} in the Appendix.
\end{itemize}
\section{Basics for STIV}
\label{sec:method}

\begin{figure*}[t]
    \centering
    \includegraphics[width=1.0\linewidth]{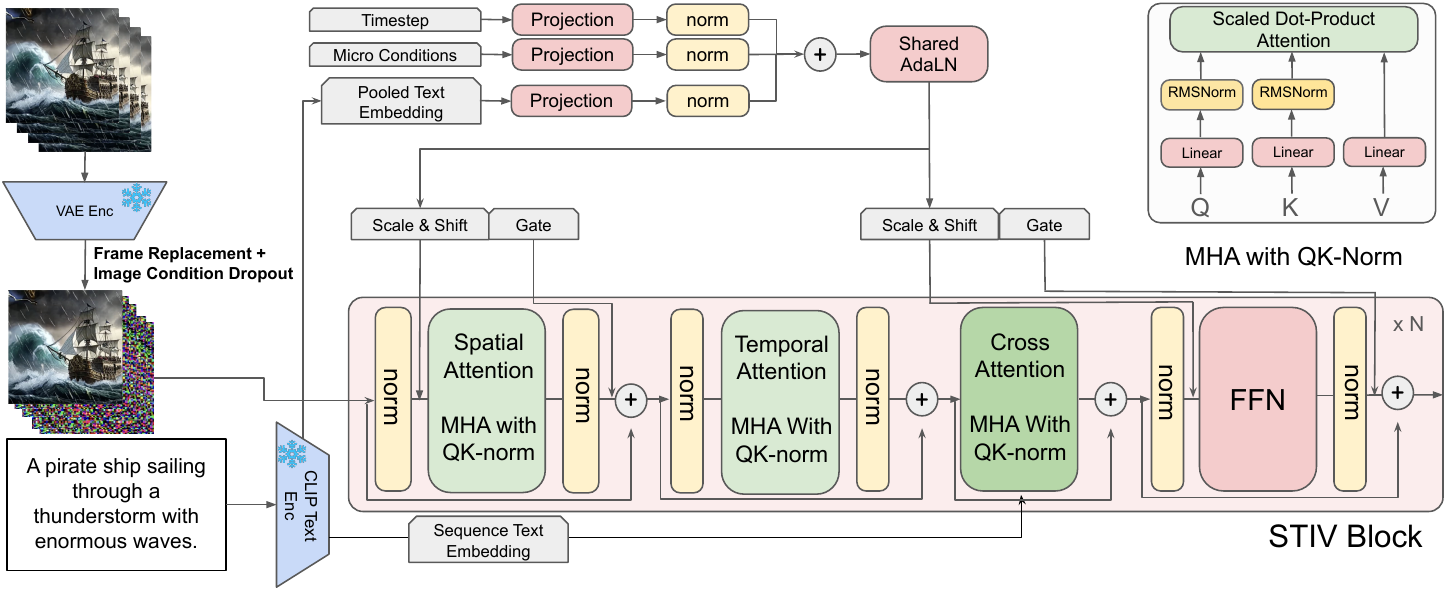}
    \caption{We replace the first frame of the noised video latents with the ground truth latent and randomly drop out the image condition. We use cross attention to incorporate the text embedding, and use QK-norm in multi-head attention, the sandwich-norm in both attention and feedforward, and stateless layernorm after singleton conditions to stabilize the training.}
    \label{fig:mmvideo-framework}
\end{figure*}

This section describes our key components of our proposed STIV method for text-image-to-video (TI2V) generation, which is illustrated in Fig.~\ref{fig:mmvideo-framework}. Afterward, Sec.~\ref{sec:ablation} and \ref{sec:exp} presents detailed experimental results.

\subsection{Base Model Architecture}
The STIV model is based on PixArt-$\alpha$~\citep{chen2023pixart}, which converts the input frames into spatial and temporal latent embeddings using a frozen Variational Autoencoder (VAE). These embeddings are then processed by a stack of learnable DiT-like blocks. We employ the T5~\cite{raffel2020exploring} tokenizer and an internally trained CLIP~\cite{radford2021learning} text encoder to process text prompts. The overall framework is illustrated in Fig.~\ref{fig:mmvideo-framework}. For more details, please refer to the appendix. The other significant architectural changes are outlined below.
\vspace{0.2cm}

\noindent\textbf{Spatial-Temporal Attention}
We employ factorized spatial and temporal attention~\cite{bertasius2021space} to handle video frames. We first fold the temporal dimension into the batch dimension and perform spatial self-attention on spatial tokens. Then, we permute the outputs and fold the spatial dimension into the batch dimension to perform temporal self-attention on temporal tokens. By using factorized spatial and temporal attention, we can easily preload weights from a text-to-image (T2I) model, as images are a special case of videos with only one temporal token and only need spatial attention.
\vspace{0.2cm}

\noindent\textbf{Singleton Condition}
We use the original image resolution, crop coordinates, sampling stride, and number of frames as micro conditions to encode the meta information of the training data. We first use a sinusoidal embedding layer to encode these properties, followed by an MLP to project them into a d-dimensional embedding space. These micro condition embeddings, along with the diffusion timestep embedding and the last text token embedding from the last layer of the CLIP model, are added to form a singleton condition. We also apply stateless layer normalization to each singleton embedding and then add them together. This singleton condition is used to produce shared scale-shift-gate parameters that are utilized in the spatial attention and feed-forward layers of each Transformer layer.
\vspace{0.2cm}

\noindent\textbf{Rotary Positional Embedding} 
Rotary Positional Embeddings (RoPE) \citep{su2024roformer} are used so that the model has a strong inductive bias for processing relative temporal and spatial relationships. Additionally, RoPE can be made compatible with the masking methods used in high compute applications and are highly adaptable to variations in resolution \citep{zhuo2024lumina}.
We apply 2D RoPE \citep{lu2024unified} for the spatial attention and 1D RoPE for the temporal attention inside the factorized Spatial-Temporal attention.
\vspace{0.2cm}

\noindent\textbf{Flow Matching}
Instead of employing the conventional diffusion loss, we opt for a Flow Matching training objective. This objective defines a conditional optimal transport between two examples drawn from a source and target distribution. In our case, we assume the source distribution to be Gaussian and utilize linear interpolates \cite{ma2024sit} to achieve this.
\begin{equation}
    \bm{x}_t = t \cdot \bm{x}_1 + (1 - t) \cdot \bm{\epsilon}.
\end{equation}
The training objective is then formulated as
\begin{equation}
    \min_\theta \mathbb{E}_{\bm{x}, \bm{\epsilon} \in \mathcal{N}(\bm{0}, \bm{I}), \bm{c}, t} \left[ \left\| \bm{F}_\theta(\bm{x}_t, \bm{c}, t) - \bm{v}_t \right\|_2^2 \right]
\end{equation}
where the velocity vector field \(\bm{v}_t = \bm{x_1} - \bm{\epsilon}\).

In inference time, we solve the corresponding reverse-time SDE, from timestep 0 to 1, to generate images from a randomly sampled Gaussian noise $\bm{\epsilon}$.

\subsection{Model Scaling}
\label{sec:model_scaling}
As we scale up the model, we encounter training instability and infrastructure challenges in fitting larger models into memory. In this section, we outline the methods to stabilize the training and enhance training efficiency.
\vspace{0.2cm}

\noindent\textbf{Stable Training Recipes}
We discovered that QK-Norm --- applying RMSNorm~\cite{zhang2019root} to the query and key vectors prior to computing attention logits --- significantly stabilizes training. This finding aligns with the results reported in SD3 \cite{esser2024scaling}. We also change from pre-norm to sandwich-norm~\citep{ding2021cogview} for both MHA and FFN, which involves adding pre-norm and post-norm with stateless layer normalization~\cite{lei2016layer} to both the layers within the STIV block.
\[
    \begin{aligned}
       \text{MHA}(x) &= x + \text{gate} \cdot \text{norm}\left(\text{Attn}\left(\text{scale} \cdot \text{norm}(x)  + \text{shift}\right)\right) \\
       \text{FFN}(x) &= x + \text{gate} \cdot \text{norm}\left(\text{MLP}\left(\text{scale} \cdot \text{norm}(x) + \text{shift}\right)\right)
    \end{aligned}
\]

\noindent\textbf{Efficient DiT Training}
We follow MaskDiT~\cite{zheng2023fast} by randomly masking 50\% of spatial tokens before passing them into the major DiT blocks. After unmasking, we add two additional DiT blocks. We also switch from AdamW to AdaFactor optimizer and employ gradient checkpointing to only store the self-attention outputs. These modifications significantly enhance efficiency and reduce memory consumption, enabling the training of larger models at higher resolution and longer duration.
\vspace{0.2cm}

\subsection{Image Conditioning}
\subsubsection{Frame Replacement}
During training, we replace the noised first frame latent with the un-noised latent of the image condition before passing the latents into the STIV blocks, and masking out the loss of the replaced frame. During inference, we use the un-noised latent of the original image condition for the first frame at each TI2V diffusion step.

The frame replacement strategy offers flexibility in extending STIV to various applications. For instance, if \(\bm{c}_I =\varnothing\), it defaults to text-to-video (T2V) generation. Conversely, if \(\bm{c}_I\) is the initial frame, it becomes the typical text-image-to-video (TI2V) generation. Moreover, if multiple frames as \(\bm{c}_I\) are provided, they can be used for video prediction even without \(\bm{c}_T\). Additionally, supplying the first and last frames as  \(\bm{c}_I\) enables the model to learn a frame interpolation, generating frames between them. Furthermore, combining  T2V and frame interpolation allows for the generation of long-duration videos: T2V generates keyframes, and frame interpolation frames then fills in frames between each pair of consecutive keyframes. Ultimately, a single model can be trained to perform all tasks by randomly selecting the appropriate conditioning strategy.
\vspace{0.1cm}

\subsubsection{Image Condition Dropout}
As discussed previously, the frame replacement strategy offers substantial flexibility for training various types of models. Here, we demonstrate a specific application in which we train a model to perform both T2V and TI2V tasks. In this case, we randomly drop out  \(\bm{c}_I\) and  \(\bm{c}_T\) during training, similar to how T2V models employ random dropout to text condition alone.
\vspace{0.1cm}

Classifier-free guidance (CFG), commonly used in text-to-image generation, has proven to be highly beneficial in enhancing the quality of generated images by directing the probability mass toward the high-likelihood regions given the condition. Building on this concept, we introduce a \textbf{Joint Image-Text Classifier-Free Guidance (JIT-CFG)} approach, which leverages both text and image conditions. It modifies the velocity estimates as
\[
    \begin{split}
\hat{\bm{F}}_\theta(\bm{x}_t,\bm{c}_T,   & \bm{c}_I, t)= {\bm{F}}_\theta(\bm{x}_t, \varnothing, \varnothing, t) \\
& \quad + s \cdot \left( \bm{F}_\theta(\bm{x}_t, \bm{c}_T, \bm{c}_I, t)  - {\bm{F}}_\theta(\bm{x}_t, \varnothing, \varnothing, t) \right)
\end{split}
\]
where \( s \) is the guidance scale. When \(\bm{c}_I =\varnothing\), it reduces to standard CFG for T2V generation. Although it is possible to introduce two separate guidance scales, as done in \citep{brooks2023instructpix2pix}, to balance the strength of the image and text conditions, we found that our two-pass approach yields strong results. Additionally, using two scales would require three forward passes, increasing the inference cost.

Empirical observations~\ref{sec:effectiveness_img_condition} suggest that applying image condition dropout with JIT-CFG effectively not only achieves multi-task training in a natural way, but also resolves the staleness issue for a $512^2$ STIV model. We hypothesize that image condition dropout prevents the model from passively overfitting to the image condition, allowing it to more effectively capture the motion information from the underlying video training data.

\subsection{Progressive Training Recipe}
\begin{figure}[hbtp]
    \centering
    \includegraphics[width=0.7\linewidth]{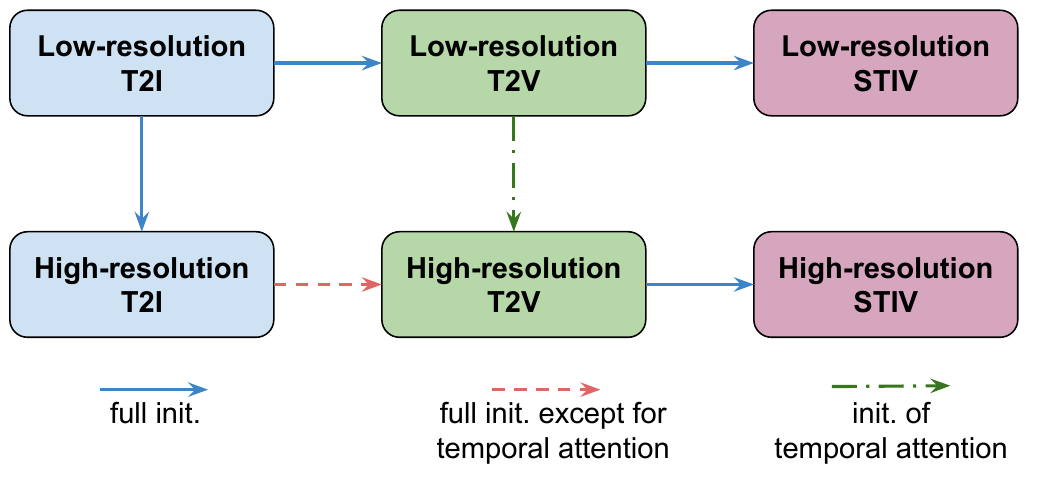}
    \caption{Progressive training pipeline of the STIV model. The T2I model is first trained to initialize the T2V model, which then initializes the STIV model at both low and high resolutions. Notably, the high-res T2V model is initialized using both the high-res T2I model and the low-res T2V model.}
    \label{fig:traince_receipe}
\end{figure}

We employ a progressive training recipe as illustrated in Figure \ref{fig:traince_receipe}. The process begins by training a text-to-image (T2I) model, which serves to initialize a text-to-video (T2V) model. Next, the T2V model serves as the starting point for initializing the STIV model. To facilitate rapid adaptation to higher resolutions and longer durations training, we incorporate interpolated RoPE embeddings in both the spatial and temporal dimensions, while initializing the model weights using those from the lower-resolution, shorter-duration models. 

\section{Recipe Study for STIV}
\label{sec:ablation}
\begin{figure}[hbtp]
    \centering
    \includegraphics[width=1.0\linewidth]{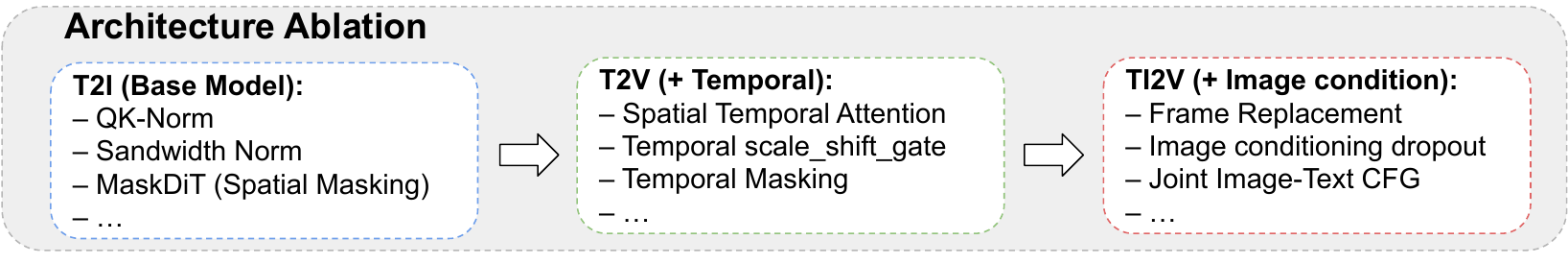}
    \caption{Ablation study of the STIV model, from the base T2I model to the temporally-aware T2V model, and finally to the image-conditioned TI2V model.}
    \label{fig:stiv_study_arch}
\end{figure}

\subsection{Basic Setup}
Before we dive into the studies of architecture and data for video generation models, we first introduce the training, data and evaluation setup before introducing our model and studies as follows:
\vspace{0.2cm}

\noindent\textbf{Training}
Unless otherwise specified, we use the AdaFactor optimizer ($\beta_1$ = 0.9, $\beta_2$ = 0.999)~\cite{shazeer2018adafactor} without any weight decay. We also clip the gradient norm if the gradient norm exceeds 1.0.
We use a constant learning rate schedule with a 1k step linear warmup with a maximum learning rate of $2 \times 10^{-4}$. 
For T2I models, we train each model for 400k steps with a batch size of 4,096. This is approximately 1.4 epochs on our internal T2I datasets. For T2V and TI2V models, we train each model for 400k steps with a batch size of 1,024. This is roughly 5.5 epochs on our internal video datasets.
For all models, exponential moving average weights are gathered by a decay rate of 0.9999 and are then used for evaluation. When MaskDiT is used, we train with 50\% spatial random masking during the initial 400k steps. Subsequently, we perform unmasked fine-tuning using all tokens. We use 50k steps of unmasked fine-tuning for T2I models and 100k steps for T2V and TI2V models.
\vspace{0.2cm}

\noindent\textbf{Data} We build a video data engine pipeline that includes video pre-processing, captioning, and filtering to accelerate the model's development when handling large-scale videos is required. Specifically, we apply PySceneDetect~\footnote{\href{https://github.com/Breakthrough/PySceneDetect}{https://github.com/Breakthrough/PySceneDetect}} to analyze video frames, detect and segment scenes based on abrupt transitions and gradual fades. This segmentation is followed by the feature extractions for filtering, including motion score, aesthetic score, text area, frame dimensions, clarity score, temporal consistency, and video orientation, among others. For each video segment, we perform dense captioning and categorization to gain a comprehensive understanding of video distribution. To further enhance caption quality, we adapt DSG~\cite{Cho2024DSG} and propose DSG-Video, a metric designed to assess hallucination rates and overall caption quality. This data engine is integral in filtering videos and preparing tailored datasets for different training stages: our data sources include Panda-70M~\cite{chen2024panda} and an internally curated high-quality dataset of 42M videos. Using our data engine, we curate over 90M high-quality video-caption pairs~\footnote{The details and effectiveness of our data engine are studied in the Appendix.}.
Next we are going to dive into more fine-grained modulation studies. As illustrated in figure \ref{fig:stiv_study_arch}, We follow the principle of studying from base T2I model to the temporally-aware T2V model, and finally to the TI2V model by adding image conditioning.
\vspace{0.2cm}

\noindent\textbf{Evaluation}
We mainly use VBench \cite{huang2024vbench}, VBench-I2V and MSRVTT~\cite{xu2016msr} to evaluate T2V and TI2V models. For VBench, we mainly report Quality (temporal quality and frame-wise quality), Semantic (semantic alignment with different perspectives of the input text prompt) and Total score (weighted average of Quality and Semantic), and they can be actually decomposed into 16 dimensions in total. VBench-I2V builds upon the VBench with three new Video-Image Alignment metrics: Subject Consistency, Background Consistency, and Camera Motion Control. These additional metrics provide a more comprehensive evaluation by focusing on how well the generated video aligns with the input image and specified prompt instructions. More details about the detailed dimensions are presented in Section~\ref{sec:detailed_vbench}.
We present three model scales: XL, XXL, and M with their configuration detailed in \cref{tab:configuration}. In the following section, we use the notation X-S-T to represent an X-size model with an $S^2$ resolution and T frames. If unspecified, the default configuration is a \(256^2\) resolution with 20 frames. More detailed model and training configurations are provided in the Appendix.

\begin{table}[h!]
\centering
\small
\begin{tabular}{l|c|c|c}
\Xhline{1pt}
Model Size & \# of STIV Blocks & Hidden Dim & \# of Attn Heads \\
\Xhline{1pt}
XL (600M) & 28 & 1,152 & 18 \\
XXL (1.5B) & 38 & 1,536 & 24 \\
M (8.7B) & 46 & 3,072 & 48 \\
\Xhline{1pt}
\end{tabular}
\caption{Model Configurations}
\label{tab:configuration}
\end{table}

\begin{table}[tbp!]
\centering
\begin{tabular}{l|ccccccc}
\Xhline{1pt}
 \multirow{2}{*}{Model} 
& \multicolumn{1}{c}{COCO} & \multicolumn{1}{c}{COCO} & \multicolumn{1}{c}{COCO} 
& \multicolumn{1}{c}{Gen} & \multicolumn{1}{c}{DSG} & \multicolumn{1}{c}{HPSv2} & \multicolumn{1}{c}{Image} \\
& \multicolumn{1}{c}{FID\textdownarrow} & \multicolumn{1}{c}{PICK\textuparrow} & \multicolumn{1}{c}{CLIP\textuparrow} 
& \multicolumn{1}{c}{Eval\textuparrow} & \multicolumn{1}{c}{Eval\textuparrow} & \multicolumn{1}{c}{Eval\textuparrow} & \multicolumn{1}{c}{Reward\textuparrow} \\
\Xhline{1pt}
Baseline & 26.17 & 20.91 & 32.03 & 0.358 & 0.571 & 26.33 & -0.25 \\
+ QK norm & 25.60 & 20.92 & 32.08 & 0.372 & 0.574 & 26.32 & -0.22 \\
+ Sandwich norm & 25.76 & 20.97 & 32.13 & 0.366 & 0.577 & 26.32 & -0.23 \\
+ Cond. norm & 25.58 & 21.05 & 32.27 & 0.393 & 0.583 & 26.43 & -0.22 \\
+ LR to 2E-4 & 26.35 & 21.03 & 32.28 & 0.379 & 0.586 & 26.40 & -0.12 \\
\hline
+ Flow & 24.96 & 21.45 & 32.90 & 0.457 & 0.639 & 26.95 & 0.15 \\
+ Renorm & 21.16 & 21.46 & 32.93 & 0.471 & 0.668 & 27.27 & 0.32 \\
\hline
+ AdaFactor & 20.26 & 21.47 & 32.97 & 0.474 & 0.661 & 27.26 & 0.32 \\
+ MaskDiT & 23.85 & 21.51 & 33.07 & 0.499 & 0.663 & 27.28 & 0.30 \\
+ Shared AdaLN & 22.83 & 21.44 & 33.12 & 0.496 & 0.658 & 27.27 & 0.24 \\
\hline
+ Micro cond. & 20.02 & 21.50 & 33.09 & 0.498 & 0.673 & 27.27 & 0.41 \\
+ RoPE & 18.40 & 21.46 & 33.11 & 0.502 & 0.680 & 27.26 & 0.48 \\
\hline
+ Internal VAE & 19.57 & 21.79 & 33.26 & 0.492 & 0.668 & 27.26 & 0.52 \\
+ Internal CLIP & \textbf{17.97} & 21.89 & 33.62 & 0.607 & 0.717 & 27.40 & 0.65 \\
+ Synth. captions & 18.04 & \textbf{22.10} & \textbf{33.65} & \textbf{0.685} & \textbf{0.751} & \textbf{27.65} & \textbf{0.81} \\
\Xhline{1pt}
\end{tabular}
\caption{Text-to-image model ablation studies.}
\label{tab:t2i_metrics}
\end{table}
\subsection{Ablation Studies for Key Changes on T2I}
\noindent We conduct a comprehensive ablation study to understand the impact of various model architecture designs and training strategies mentioned in Sec.~\ref{sec:method} on the text-to-image generation task. To evaluate generation quality, we use a suite of popular automated metrics, including FID score \cite{FID}, Pick Score \cite{kirstain2023pick}, CLIP Score, GenEval \cite{ghosh2024geneval}, and DSGEval \cite{Cho2024DSG}, Human Preference Score (HPSv2) \cite{wu2023human}, Image Reward \cite{xu2023imagereward}.


We began with a base T2I-XL model, a DiT \cite{peebles2023scalable} model augmented with cross-attention layers to integrate with text embeddings. Initially, we applied a series of stabilization techniques, including QK-norm, sandwich-norm and singleton condition norm, which yielded comparable results to the baseline. Notably, these techniques enabled us to train models stably even with a learning rate increased from 1e-4 to 2e-4. We demonstrated that incorporating Flow Matching during training and employing CFG-Renormalization\footnote{Detailed description in Appendix.} during inference improved all the metrics substantially. Subsequently, we explored techniques to reduce training memory, such as AdaFactor Optimizer, MaskDiT, and Shared AdaLN, which maintained similar performance. Utilizing micro conditions and RoPE further reduced the FID score and improved DSGEval and Image Reward. Finally, incorporating an internally trained bigG CLIP model improved on all metrics even more. Notably, combining synthetic recaption with original caption following \cite{lai2024revisit} achieved the best results in almost all metrics. For more details, refer to the Appendix~\ref{appendix:t2i_ablation}.

We use the optimal model architecture and training hyperparameters based on the T2I ablation study as our starting point for the remaining T2V and TI2V experiments.

\subsection{Ablation Studies on Key Designs for T2V}
\noindent\textbf{Key Modulation}
We make some design choices in our model based on the evaluations on VBench, as shown in Fig.~\ref{tab:modulate_t2v_vbench}.
The base model uses a temporal path size of 2, non-causal temporal attention, and a spatial masking ratio of 0.5. As expected, the model with temporal patch=1 performs the best, but it is only slightly better with 2x compute. However, the model with temporal patch=4 leads to a noticeable performance drop.
Using causal temporal attention also results in a significant drop in both quality and total scores. Adding a scale-shift-gate to the temporal attention layer~\footnote{See Fig. \ref{fig:mmvideo-framework} for the diagram of the model.} is slightly worse than the baseline, despite having more parameters.
Furthermore, removing the spatial masking results in a slight decrease in the Semantic score and an improvement in the Quality and Total scores. However, this comes at the cost of requiring more compute as the length of tokens are doubled. On the other hand, using temporal masking significantly degrades model performance, with large drops observed in the VBench quality and final scores.
\vspace{0.2cm}

\begin{figure*}
\begin{minipage}{.53\textwidth}
\begin{subfigure}[b]{\linewidth}
\centering
\small
\resizebox{\linewidth}{!}{%
\begin{tabular}{l|c|c|c}
\Xhline{1pt}
\multirow{2}{*}{Module} & \multicolumn{3}{c}{VBench} \\
\cline{2-4}
& Quality $\uparrow$ & Semantic $\uparrow$ & Total $\uparrow$ \\
\Xhline{1pt}
Base model & 80.19 & 70.51 & 78.25 \\
\hspace{1mm} w/ temp. patch=1 & \textbf{80.92} & 71.69 & \textbf{79.07} \\
\hspace{1mm} w/ temp. patch=4 & 79.72 & 69.15 & 77.61 \\
\hspace{1mm} w/ causal temp.\_atten & 74.59 & \textbf{73.13} & 74.30 \\
\hspace{1mm} + temp. scale\_shift\_gate & 80.32 & 68.94 & 78.04 \\
\hspace{1mm} + temp. mask & 77.58 & 65.95 & 75.25 \\
\hspace{1mm} - spatial mask & 80.57 & 70.31 & 78.52 \\
\Xhline{1pt}
\end{tabular}
}
\vspace{0.2cm}
\caption{Ablation Study of T2V model design using T2V-XL. The base model uses temporal path size 2, non-causal temporal attention, spatial masking ratio 0.5, and no temporal masking.}
\vspace{0.2cm}
\label{tab:modulate_t2v_vbench}
\end{subfigure}

\end{minipage}
\hfill
\begin{minipage}{.45\linewidth}
\begin{subfigure}[b]{\linewidth}
    \small
    \centering
\resizebox{\linewidth}{!}{%
    \begin{tabular}{p{1.5cm}|c|c|c|c}
    \Xhline{1pt}
    \multirow{2}{*}{Init.} & \multicolumn{1}{c|}{MSRVTT} & \multicolumn{3}{c}{VBench} \\
    \cline{2-5}
     & FVD $\downarrow$ & Quality $\uparrow$ & Semantic $\uparrow$ & Total $\uparrow$ \\
    \Xhline{1pt}
    Scratch             & 417.98 & 80.27 & 67.84 & 77.78 \\
    T2V-256             & 415.63 & 80.28 & 71.29  & 78.49 \\
    T2I-512             & \textbf{401.83} & 79.77 & 71.58 & 78.13 \\
    Both               & 405.14 & \textbf{80.45} & \textbf{72.37} & \textbf{78.83} \\
    \Xhline{1pt}
    \end{tabular}
}
    \caption{Different model initialization for T2V-XL-512.}
    \label{tab:t2v_high_res_init}
\vspace{0.2cm}
\end{subfigure}
\hfill
\begin{subfigure}[b]{\linewidth}
    \small
    \centering
\resizebox{\linewidth}{!}{%
    \begin{tabular}{l|c|c|c|c}
    \Xhline{1pt}
    \multirow{2}{*}{Init.} & \multicolumn{1}{c|}{MSRVTT} & \multicolumn{3}{c}{VBench} \\
    \cline{2-5}
    & FVD $\downarrow$ & Quality $\uparrow$ & Semantic $\uparrow$ & Total $\uparrow$ \\
    \Xhline{1pt}
    T2I                     & 549.13 & 78.71 & 65.69 & 76.10 \\
    T2V (inter.)           & 407.86 & 79.56 & 65.42 & 76.73 \\
    T2V (extra.)           & \textbf{397.90} & 79.18 & 64.63 & 76.27 \\
    T2V 2x (inter.)        & 401.94 & \textbf{79.59} & \textbf{66.24} & \textbf{76.92} \\
    \Xhline{1pt}
    \end{tabular}
}
    \caption{Different initialization for T2V-XL 40 frames. }
    \label{tab:t2v_high_frame_count_init}
\end{subfigure}
\end{minipage}

\caption{Ablation studies of key designs for T2V.}
\end{figure*}

\noindent\textbf{Model Initialization}
\label{sec:t2v init}
We investigate how initialization impacts the performance of T2V-XL models. We train $512^2$ T2V models by four different paths under a controlled total FLOPs setting: from scratch, initializing from a lower resolution T2V-256 model, initializing from a T2I-512 model, and loading both the temporal and spatial weights from T2V-256 and T2I-512 models respectively (Fig.~\ref{tab:t2v_high_res_init}). We find that jointly initializing from both a low resolution T2V model and a high resolution T2I model can achieve better VBench metrics. This joint initialization method yields slightly improved FVD values compared to training from scratch and offers benefits in terms of efficient experimentation and cost when low resolution models are already present. Under a similar methodology we additionally explore the effects of training T2V models with higher numbers of frames (40 frames) by initializing from shorter T2V models (20 frames). Fig.~\ref{tab:t2v_high_frame_count_init} shows that when training models with a higher number of frames initializing from a low frame count model achieves improved metrics over initializing directly from a T2I model. Using interpolation of the RoPE embeddings yields improved VBench scores compared to extrapolation. Additionally we find that initializing the high frame count training from a T2V model trained with a proportionally lower frame rate (higher frame sub-sampling stride) can improve the VBench metrics, particularly the motion smoothness and dynamic degree.

\vspace{-1mm}
\subsection{Ablation Studies on Key Designs for TI2V}
\label{sec:image_cond}
To integrate the image condition with the text condition, we reformulate the model as \(\bm{F}_\theta(\bm{x}_t, \bm{c}_T, \bm{c}_I, t)\), where \(\bm{c}_T\) and \(\bm{c}_I\) are the text and image conditions. Then, we studied each design component in TI2V framework and tackled multi-task learning and staleness issue encountered when training high resolution TI2V models.

\subsubsection{The Effectiveness of Frame Replacement}
\label{sec:effectiveness_frame}

We ablate several model variants for TI2V on STIV-XL model, by combining the following key components: Frame Replacement (FR), Cross Attention (CA), Large Projection (LP), and First Frame Loss (FFL)~\footnote{(1) FR uses frame replacement alone for strong image-video alignment. (2) CA uses cross attention alone to align features between the input image and generated video. (3) FFL removes the first frame loss mask introduced in Section~\ref{sec:image_cond} to constrain the initial frame of the video. (4) LP employs a more powerful ResNet 2D encoder as the projection layer here.}. As shown in Tab.~\ref{tab:ti2v_ablation_metrics}, notably, adding a large projection layer enhances the information passed by the cross-attention, resulting in improvements in both subject and background consistency. However, this approach may overly constrain the model, as evidenced by a reduction in the dynamic degree score (22.36 for FR + CA + LP compared to 35.4 for FR + CA), indicating that the model might exert excessive control over the generated output. Additionally, adding a first-frame loss, though seemingly beneficial, has shown to reduce overall scores, particularly in aspects of motion quality, suggesting that this loss might inadvertently constrain the model’s temporal dynamics. In contrast, frame replacement alone has proven to be a robust and effective approach, yielding consistent improvements without negatively impacting other dimensions of video quality. The frame replacement (FR) model achieves high scores in I2V average scores (the average of I2V Subject, I2V Background and Camera Motion) and total average scores. These results underline the advantage of frame replacement as a foundational component, providing a stable backbone for maintaining quality across diverse dimensions.
\begin{table*}[htbp]
\small
\centering
\resizebox{\linewidth}{!}{
\begin{tabular}{l|c|c|c|c|c|c|c|c|c|c|c|c}
\Xhline{1pt}
\multirow{2}{*}{\textbf{Models}} & \textbf{Subj} & \textbf{Bg} & \textbf{Temp} & \textbf{Mot} & \textbf{Dyn} & \textbf{Aesth} & \textbf{Img} & \textbf{I2V} & \textbf{I2V} & \textbf{Cam} & \textbf{I2V Avg} & \textbf{Avg} \\
& \textbf{Cons} & \textbf{Cons} & \textbf{Flick} & \textbf{Smooth} & \textbf{Deg} & \textbf{Qual} & \textbf{Qual} & \textbf{Subj} & \textbf{Bg} & \textbf{Mot} & \textbf{Scores} & \textbf{Scores} \\
\Xhline{1pt}
CA & 82.2 & 92.8 & 95.7 & 96.3 & \textbf{42.4} & 48.8 & 65.5 & 88.9 & 90.9 & 26.9 & 68.2 & 73.0 \\
CA + FFL & 84.5 & 95.6 & 96.1 & 96.7 & 29.7 & 48.7 & 64.7 & 91.5 & 94.7 & 17.6 & 67.2 & 72.0 \\
CA + LP & 95.2 & \textbf{98.7} & \textbf{97.4} & 98.1 & 22.2 & 57.3 & \textbf{66.8} & 96.9 & 97.3 & 22.7 & 72.3 & 75.3 \\
FR & 94.5 & 98.3 & 96.6 & 97.8 & 36.6 & \textbf{58.0} & 66.1 & 96.8 & 97.1 & \textbf{31.5} & \textbf{75.8} & \textbf{77.3} \\
FR + CA & 95.1 & 98.6 & 97.0 & 98.1 & 35.4 & \textbf{58.0} & 66.2 & 96.9 & 97.3 & 28.8 & 74.4 & 77.1 \\
FR + CA + LP & \textbf{95.3} & 98.5 & 97.3 & \textbf{98.2} & 22.4 & 57.3 & 66.3 & \textbf{97.0} & \textbf{97.4} & 25.8 & 73.4 & 75.6 \\
FR + CA + LP + FFL & 95.2 & \textbf{98.7} & \textbf{97.4} & 98.1 & 22.2 & 57.3 & \textbf{66.8} & 96.9 & 97.3 & 22.7 & 72.3 & 75.3 \\
\Xhline{1pt}
\end{tabular}
}
\caption{Ablation Study Results for Different Model Components for Text-Image-To-Video~(TI2V) task on VBench-I2V.}
\label{tab:ti2v_ablation_metrics}
\end{table*}
\subsubsection{The Effectiveness of Image Condition Dropout}
\label{sec:effectiveness_img_condition}
Our experiments show that  image condition dropout with JIT-CFG not only supports multi-task training but also resolves staleness in a $512^2$ STIV model. 
\vspace{0.2cm}

\noindent\textbf{Multi-task training}
By using image-conditioning dropout during STIV training, we effectively enable both T2V and TI2V capability. As shown in Tab.~\ref{tab:multi_task}, models trained exclusively on T2V or TI2V task alone cannot perform the other task, while  STIV with image condition dropout can easily handles both two task well, achieving performance comparable to the best single-task models. 

\begin{table}[htbp]
\centering
\small
\begin{tabular}{l|ccc|ccc}
\Xhline{1pt}
\multirow{2}{*}{\textbf{Model}} & \multicolumn{3}{c|}{\textbf{VBench-T2V}} & \multicolumn{3}{c}{\textbf{VBench-I2V}} \\
\cline{2-7}
 & \textbf{Q} $\uparrow$ & \textbf{S} $\uparrow$ & \textbf{T} $\uparrow$ & \textbf{I} $\uparrow$ & \textbf{Q} $\uparrow$ & \textbf{T} $\uparrow$ \\
\Xhline{1pt}
T2V-M-512 & 82.2 & 77.0 & \textbf{81.2} & / & / & / \\
STIV-M-512 & 74.6 & 31.9 & 66.1 & \textbf{98.0} & 82.1 & \textbf{90.1} \\
STIV-M-512-JIT & 82.3 & \textbf{74.1} & 80.7 & 97.6 & 81.9 & 89.8 \\
STIV-M-512-JIT-TUP & \textbf{83.0} & 73.1 & 81.0 & 97.2 & \textbf{82.3} & 89.7 \\
\Xhline{1pt}
\end{tabular}
\caption{Comparison of T2V, STIV and STIV with JIT-CFG on VBench and VBench-I2V \textbf{I}2V Score, \textbf{Q}uality, \textbf{T}otal scores.}
\label{tab:multi_task}
\end{table}

\noindent\textbf{Greater motion}
\label{sec:greater_motion}
In practice, we have observed that while STIV-M-512 performs well on VBench-I2V, it sometimes generates static frames. The VBench-I2V metric tends to favor videos with less motion, prioritizing smoothness and consistency. 
As shown in Tab.~\ref{tab:model_comparison}, STIV with JIT-CFG achieves higher dynamic degrees at the cost of a slight reduction in consistency and smoothness scores. We also show visual comparisons from Fig.~\ref{fig:dpnodp1} to Fig.~\ref{fig:dpnodp3} in the Appendix.

\begin{table}[htbp]
\centering
\small
\begin{tabular}{l|c|c|c|c}
\Xhline{1pt}
\multirow{2}{*}{\textbf{Model}} & \textbf{Dynamic} & \textbf{Motion} & \textbf{Temporal} & \textbf{Background}  \\
& \textbf{Degree} & \textbf{Smoothness} & \textbf{Consistency} & \textbf{Flickering} \\
\Xhline{1pt}
STIV-M-512 & 10.2 & \textbf{99.6} & \textbf{99.3} & \textbf{99.1} \\
STIV-M-512-JIT & \textbf{24.0} & 99.1 & 98.6 & 98.6 \\
\Xhline{1pt}
\end{tabular}
\caption{Effect of JIT-CFG on motion-related scores.}
\label{tab:model_comparison}
\end{table}

\noindent\textbf{JIT-CFG and the Variant}
It is natural to think about extending the traditional classifier-free guidance (CFG) to a three-copy weighted approach, where three conditioning modes are considered: 
(1) \textbf{Null condition}: Both the image (\( \bm{c}_I \)) and text (\( \bm{c}_T \)) conditions are null (\( \varnothing \)).
(2) \textbf{Image-only condition}: The image condition is the source image, while the text condition is null.
(3) \textbf{Joint condition}: Both the image and text conditions are provided. The velocity estimates are combined as:
\[
\begin{split}
\hat{\bm{F}}_\theta(\bm{x}_t, \bm{c}_T, \bm{c}_I, t) = & {\bm{F}}_\theta(\bm{x}_t, \varnothing, \varnothing, t) \\
& \quad + s_1 \cdot \left( \bm{F}_\theta(\bm{x}_t, \bm{c}_I, \varnothing, t) - {\bm{F}}_\theta(\bm{x}_t, \varnothing, \varnothing, t) \right) \\
& \quad + s_2 \cdot \left( \bm{F}_\theta(\bm{x}_t, \bm{c}_I, \bm{c}_T, t) - {\bm{F}}_\theta(\bm{x}_t, \bm{c}_I, \varnothing, t) \right).
\end{split}
\]

Here, \(s_1\) and \(s_2\) are guidance scales for the image-only (CFG-I) and joint conditions (CFG-T), respectively. We dub it separate image and text classifier-free guidance (SIT-CFG). We achieved \(FVD=94.1\) on the MSRVTT test set using STIV-M-512-JIT by setting \(s=7.5\) in JIT-CFG. Meanwhile, we conducted experiments on our STIV-M-512-JIT, performing a grid search on \(s_1\) and \(s_2\) by taking the Cartesian product \((1.1, 1.5, 4.5, 7.5, 10.5) \times (1.1, 1.5, 4.5, 7.5, 10.5)\).

\begin{figure}[hbtp]
    \centering
    \begin{minipage}{0.53\textwidth}
        \quad As shown in the Fig.~\ref{fig:heatmap_cfg}, we observed: 
        (1) Fixing CFG-T, as CFG-I increases, FVD first decreases and then increases; 
        (2) Fixing CFG-I, as CFG-T increases, FVD continuously decreases, except when CFG-I is very small (1.1, 1.5), where it first decreases and then increases; 
        (3) The best configuration occurs at \(CFG-T=7.5\) and \(CFG-I=1.5\), yielding \(FVD=95.2\). 
        However, overall, SIT-CFG does not show significant advantages compared to JIT-CFG, and using two copies for inference is significantly less efficient. Note that this search was optimized for MSRVTT, and for other prompts requiring stronger dependence on the first-frame subject, a larger CFG-I might be needed.
    \end{minipage}
    \hfill
    \begin{minipage}{0.45\textwidth}
        \includegraphics[width=\linewidth]{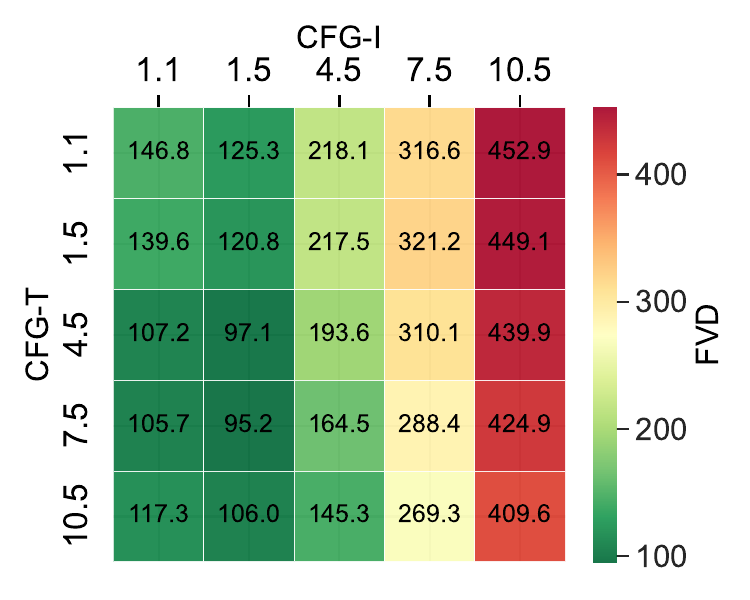}
        \caption{Grid search the optimal FVD for CFG-I \(s_1\) and CFG-T \(s_2\) on MSRVTT~\citep{xu2016msr} test set.}
        \label{fig:heatmap_cfg}
    \end{minipage}
\end{figure}

\newpage
\subsubsection{Model Initialization}
We also studied how the performance of TI2V models is affected by the initialization methods, including starting from T2I or T2V. We run the same total number of steps to check the final performance on VBench-I2V. 
From Tab.~\ref{tab:ti2v_init}, we observe that initializing from T2V model can achieve a better camera motion score and slightly better dynamic degree, while comparable to initializing from T2I on all other dimensions.
\vspace{0.1cm}
\begin{table}[h!]
    \centering
    \small
    \resizebox{\linewidth}{!}{%
    \begin{tabular}{l|c|c|c|c|c|c|c|c|c|c|c}
        \Xhline{1pt}
        \multirow{2}{*}{\textbf{Initialization}} & \textbf{Subj} & \textbf{Bg} & \textbf{Temp} & \textbf{Mot} & \textbf{Dyn} & \textbf{Aesth} & \textbf{Img} & \textbf{I2V} & \textbf{I2V} & \textbf{Cam} & \textbf{Avg} \\
        & \textbf{Cons} & \textbf{Cons} & \textbf{Flick} & \textbf{Smooth} & \textbf{Deg} & \textbf{Qual} & \textbf{Qual} & \textbf{Subj} & \textbf{Bg} & \textbf{Mot} & \textbf{Scores} \\
        \Xhline{1pt}
        T2V & 94.1 & 98.2 & 96.5 & 97.7 & \textbf{37.1} & \textbf{57.9} & 65.5 & \textbf{96.6} & 96.9 & \textbf{38.0} & \textbf{77.9}  \\
        T2I & \textbf{94.5} & \textbf{98.7} & \textbf{96.9} & \textbf{97.9} & 36.5 & 57.4 & \textbf{66.1} & 96.6 & \textbf{97.3} & 29.8 & 77.2\\
        \Xhline{1pt}
    \end{tabular}
    }
    \caption{Results for different model initialization on VBench-I2V.}
    \label{tab:ti2v_init}
\end{table}

\subsection{Video Data Engine}
\label{sec:data}
\begin{figure*}[t]
    \centering
    \includegraphics[width=1\linewidth]{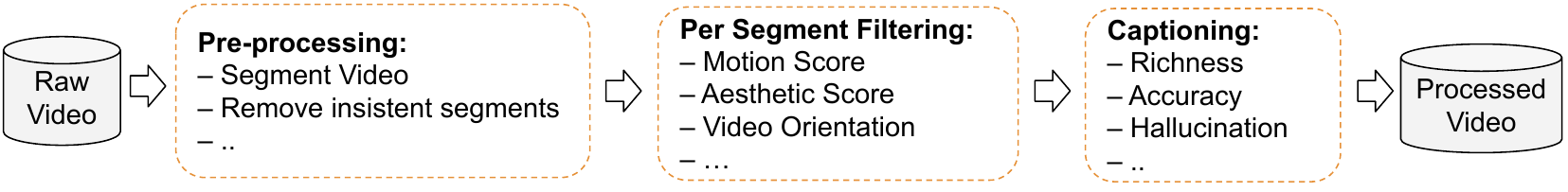}
    \caption{An overview of our video data engine, including video pre-processing, filtering, and video captioning.}
    \label{fig:data_overview}
\end{figure*}

Data quality is pivotal for video generation models. However, curating large-scale, high-quality datasets remains challenging due to issues like noisy captions, hallucinations, and limited diversity in video content and duration. To address these concerns, we propose a \textbf{Video Data Engine} (Fig.~\ref{fig:data_overview})—a comprehensive pipeline that improves dataset quality and reduces hallucinations, ultimately enhancing model performance. More details can be found in Sec.~\ref{sec:data_detailed} in the appendix.

Our approach focuses on three key questions: (1) How to preprocess raw videos for better consistency? (2) What is the effect of data filtering on model performance? (3) How can advanced video captioning reduce hallucinations and improve outcomes? We use Panda-70M~\citep{chen2024panda} as a working example and produce a curated subset, Panda-30M, via our pipeline.

\noindent\textbf{Video Pre-processing and Feature Extraction.}
We employ PySceneDetect\footnote{\href{https://github.com/Breakthrough/PySceneDetect}{https://github.com/Breakthrough/PySceneDetect}} to remove abrupt transitions and inconsistent segments, yielding more coherent clips. We then extract key features (e.g., motion and aesthetic scores) to guide subsequent filtering.
\vspace{0.2cm}

\noindent\textbf{Data Engine for Filtering}
Effective data filtering is crucial for improving dataset quality and reducing hallucinations. We develop an automated filtering infrastructure that supports efficient data selection, quality control, and continuous improvement throughout the model's development lifecycle. For instance, we can sample high-quality videos with predefined resolutions / motion scores for the fine-tuning stage. This filtering system allows us to systematically remove low-quality videos and focus on data that enhances model performance. From Panda-30M, we further apply filtering based on motion score and aesthetic score to obtain Panda-10M, named as a high-quality version of Panda-30M. The results are summarized in Tab.~\ref{tab:high-quality}: instead of pursuing data volume, higher-quality videos have the potential to achieve more promising results.  





\begin{table}[h]
    \small
    \centering
    \begin{tabular}{p{2cm}|c|c|c|c}
    \Xhline{1pt}
    \multirow{2}{*}{Data} & \multicolumn{1}{c|}{MSRVTT} & \multicolumn{3}{c}{VBench} \\
    \cline{2-5}
     & FVD $\downarrow$ & Quality $\uparrow$ & Semantic $\uparrow$ & Total $\uparrow$ \\
    \Xhline{1pt}
    Panda-30M            & 770.9 & 80.4 & \textbf{73.6} & 65.6 \\
    Panda-10M            & \textbf{759.2} & \textbf{80.8} & 73.4 & \textbf{66.2} \\
    \Xhline{1pt}
    \end{tabular}
    \vspace{0.1cm}
    \caption{Compare Panda-30M and Panda-10M (high-quality) using XL T2V model.}
    \label{tab:high-quality}
\end{table}

\vspace{0.2cm}
\noindent\textbf{Video Captioning Model}
High-quality video-text pairs are essential for training text-to-video models. Existing datasets often suffer from noisy or irrelevant captions, limited in describing temporal dynamics. We initially attempted a frame-based captioning approach followed by LLM summarization~\cite{blattmann2023stable}, but found that single-frame captions fail to represent motion, and LLM summarization can induce hallucinations. To improve caption quality while balancing cost, we employ LLaVA-Hound-7B~\cite{zhang2024llavahound}—a video LLM capable of producing more coherent and motion-aware descriptions.

\vspace{0.2cm}

\begin{figure}[!t]
    \centering
    \includegraphics[width=\linewidth]{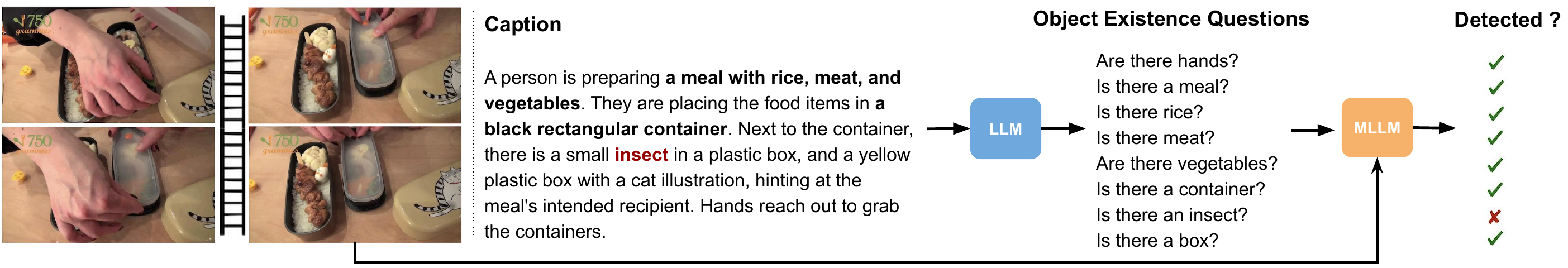}
    \caption{An overview of DSG-Video’s approach to detecting object hallucinations in captions: we use an LLM to generate questions and another MLLM to validate the presence of the object across frames. If the MLLM fails to detect the object in all frames, the object is classified as a hallucination.}
    \label{fig:dsg_video_overview}
    \vspace{-0.2cm}
\end{figure} 

\noindent\textbf{Caption Evaluation and Ablations}
To objectively assess caption accuracy, we introduce \textbf{DSG-Video}(Fig.~\ref{fig:dsg_video_overview}), a module inspired by DSG~\cite{Cho2024DSG}, that detects hallucinated objects by probing captions with LLM-generated questions and verifying object presence in sampled video frames using a multimodal LLM. This yields two metrics, $\text{DSG-Video}_{i}$ and $\text{DSG-Video}_{s}$~\footnote{More details are illustrated in Appendix \ref{sec:data_detailed}.}, reflecting hallucination at the object and sentence levels, respectively. We compare two captioning strategies—frame-based plus LLM summarization (FCapLLM) and direct video captioning (VCap)—on the Panda-30M dataset. As shown in Tab.~\ref{tab:caption-eval}, VCap reduces hallucinations and increases the diversity of described objects, leading to improved T2V model performance. These results demonstrate that richer, more accurate video descriptions can significantly enhance downstream generation quality.

\begin{table}[h]
\centering
\small
\begin{tabular}{c|c|cc|cc}
\Xhline{1pt}

\textbf{Caption} & \textbf{Total Object} & \textbf{DSG-$\text{Video}_{i} (\downarrow)$} & \textbf{DSG-$\text{Video}_{s} (\downarrow)$} & \textbf{MSRVTT FVD ($\downarrow$)} & \textbf{VBench ($\uparrow$)} \\
\hline 
FCapLLM & 1249 & 6.4 & 24.0 & 808.1 & 64.2 \\
\rowcolor{tbgray}  VCap & 1911 & \textbf{5.3} & \textbf{15.0} & \textbf{770.9} & \textbf{65.6} \\

\Xhline{1pt}
\end{tabular}
\caption{Compare different captions using XL T2V model. DSG-Video metrics are calculated from 100 random captions.}

\label{tab:caption-eval}
\end{table}

\section{Results}
\label{sec:exp}
Based on all of these studies, we scale our T2V and STIV model from 600M to 8.7B. We show the main results in \cref{tab:t2v_metrics} and \cref{tab:ti2v_main}, comparing our models with state-of-the-art open sourced and close sourced models, which demonstrates the effectiveness of our recipes. Specifically, we do finetuning on top of the pretrained video generation models (SFT), based on the 20,000 videos filtered from Panda-70M~\cite{chen2024panda} using the method mentioned in Section~\ref{sec:data}. Since we adopt MaskDiT technique in our pretraining stage, we try finetuning our model in an unmask manner (UnmaskSFT). We also finetuned our STIV model to become a temporal upsampler to interpolate the videos generated by our main T2V and STIV models to boost the motion smoothness (+ TUP). 

\vspace{0.2cm}
\label{sec:stive exp results}
\noindent\textbf{T2V Performance}
We first showcase the effectiveness of our T2V model as the foundation for STIV.
\cref{tab:t2v_metrics} presents a comparison of different T2V model variants on VBench, including the VBench-Quality, VBench-Semantic, and VBench-Total scores.
Our analysis reveals that scaling up model parameters in our T2V model improves semantic following capability. Specifically, as model size increase from XL to XXL and M, VBench-Semantic scores rise from 72.5 to 72.7 and then to 74.8. This explicit emergence (from XL, XXL to M), suggesting larger models are better at capturing semantic information. However, the impact on video quality, measured by VBench-Quality, remains modest, with only a slight increase from 80.7 to 81.2 and then to 82.1. This finding suggests that scaling has a greater effect on the model's semantic capabilities than on video quality. Furthermore, increasing the spatial resolution from 256 to 512 significantly boosts the VBench-Semantic score from 74.8 to 77.0. Detailed results can be found in ~\cref{tab:t2v_detailed}.
\vspace{0.2cm}

\noindent\textbf{The Influence of SFT}
Additionally, fine-tuning the model with high-quality SFT data markedly enhances the VBench-Quality score from 82.2 to 83.9. Finetuning our model without any masked tokens slightly increases the performance of model on the semantic score. Our best model achieves a VBench-Semantic score of 79.5, outperforming prominent closed source models such as KLING, PIKA, and Gen-3. With the temporal upsampler, our model can achieve the state-of-the-art quality score compared with all other models.

\begin{table}[t]
\centering
\small
\begin{tabular}{l|c|c|c}
\Xhline{1pt}
\textbf{Model} & \textbf{Quality} $\uparrow$ & \textbf{Semantic} $\uparrow$ & \textbf{Total} $\uparrow$ \\ 
\Xhline{1pt}
\multicolumn{4}{c}{\textit{Open Sourced Models}}\\
\hline
OpenSora V1.2~\citep{zheng2024open-sora} & 81.4 & 73.4 & 79.8 \\
AnimateDiff-V2~\citep{guo2023animatediff} & 82.9 & 69.8 & 80.3 \\
VideoCrafter-2.0~\citep{chen2024videocrafter2} & 82.2 & 73.4 & 80.4 \\
T2V-Turbo~\cite{li2024t2v-turbo} & 82.2 & 74.5 & 80.6 \\
CogVideoX-2B~\citep{yang2024cog-video-x} & 82.2 & 75.8 & 80.9 \\
Allegro~\citep{zhou2024allegro} & 83.1 & 73.0 & 81.1 \\
CogVideoX-5B~\citep{yang2024cog-video-x} & 82.8 & 77.0 & 81.6 \\
LaVie-2~\citep{wang2023lavie} & 83.2 & 75.7 & 81.8 \\
\hline
\multicolumn{4}{c}{\textit{Close Sourced Models}}\\
\hline
Gen-2~\citep{runway2023gen} & 82.5 & 73.0 & 80.6 \\
PIKA~\citep{pika2023pika} & 82.9 & 71.8 & 80.7 \\
EMU3~\citep{girdhar2023emu} & 84.1  & 68.4 & 81.0 \\
KLING~\citep{kuaishou2024kling} & 83.4 & 75.7 & 81.9 \\
Gen-3~\citep{runway2024gen} & 84.1 & 75.2 & 82.3 \\
\hline
\multicolumn{4}{c}{\textit{Ours}}\\
\hline
XL & 80.7 & 72.5 & 79.1 \\
XXL & 81.2 & 72.7 & 79.5 \\
M & 82.1 & 74.8 & 80.6 \\
M-512 & 82.2 & 77.0 & 81.2 \\
M-512 SFT & 83.9 & 78.3 & 82.8 \\
M-512 SFT + TUP & 84.2 & 78.5 & \textbf{83.1} \\
M-512 UnmaskSFT & 83.7 & \textbf{79.5} & 82.9 \\
M-512 UnmaskSFT + TUP & \textbf{84.4} & 77.2 & 83.0 \\
\Xhline{1pt}
\end{tabular}
\caption{Performance comparison of T2V variants with open-sourced and close-sourced models on VBench.}
\label{tab:t2v_metrics}
\end{table}
\begin{table}[h]
\centering

\begin{tabular}{l|c|c|c}
\Xhline{1pt}
\textbf{Model} & \textbf{Quality} $\uparrow$ & \textbf{I2V} $\uparrow$ & \textbf{Total} $\uparrow$ \\ 
\Xhline{1pt}
VideoCrafter-I2V~\citep{chen2023videocrafter1} & 81.3 & 89.0 & 85.1 \\
Consistent-I2V~\citep{ren2024consisti2v} & 78.9 & 94.8 & 86.8 \\
DynamicCrafter-256~\citep{xing2023dynamicrafter} & 80.2 & 96.6 & 88.4 \\
SEINE-512~\citep{chen2023seine} & 80.6 & 96.3 & 88.4 \\
I2VGen-XL~\citep{zhang2023i2vgen} & 81.2 & 95.8 & 88.5 \\
DynamicCrafter-512~\citep{xing2023dynamicrafter} & 81.6 & 96.6 & 89.1 \\ 
Animate-Anything~\cite{dai2023animateanything} & 81.2 & \textbf{98.3} & 89.8 \\
SVD~\citep{blattmann2023stable} & \textbf{82.8} & 96.9 & 89.9 \\
\hline
STIV-XL & 79.1 & 95.7 & 87.4 \\
STIV-M & 78.8 & 96.3 & 87.6 \\
STIV-M-512 & \underline{82.1} & \underline{98.0} & \textbf{90.1} \\
STIV-M-512-JIT & 81.9 & 97.6 & 89.8 \\
\Xhline{1pt}
\end{tabular}
\caption{Performance comparison of STIV-TI2V variants with open-sourced and close-sourced models on VBench-I2V.}
\label{tab:ti2v_main}
\end{table}

\vspace{0.2cm}

\noindent\textbf{TI2V Performance}
As shown in Tab.~\ref{tab:ti2v_main}, our model delivers competitive performance compared to state-of-the-art approaches. It also reveals that while scaling up improves the I2V score, it has minimal impact on quality. In contrast, increasing the resolution leads to noticeable improvements in both quality and I2V scores. We provide complete results for the decomposed dimensions in ~\cref{tab:ti2v_metrics}.


\section{Flexible Applications}

Here, we demonstrate how to extend our STIV to various applications, such as video prediction, frame interpolation, multi-view generation, and long video generation.
\vspace{0.2cm}

\noindent\textbf{Video Prediction}
We initialize from a STIV-XXL model to train a text-video-to-video model conditioned on the first four frames. As shown in Fig.~\ref{tab:tv2v}, the video-to-video model (STIV-V2V) shows significantly lower FVD scores compared to the text-to-video model (T2V) on MSRVTT~\citep{xu2016msr} test set and MovieGen Bench~\citep{polyak2024movie}. This result indicates that video-to-video models can achieve superior performance, which is promising for applications in autonomous driving and embodied AI where high fidelity and consistency in generated video frames are crucial.

\vspace{0.2cm}

\noindent\textbf{Frame Interpolation}
We propose STIV-TUP, a temporal upsampler initialized from an STIV-XL model, and continue train conditioned on consecutive frames sampled by stride of 2 with the text conditioning.  Fig.~\ref{tab:t_upsampler_performance} shows that our STIV can also be used to do decent frame interpolation conditioned on both text and image. We observe that using text conditions is slightly better in FID and FVD on the MSRVTT test set. We also cascade the temporal upsampler with our main model to explore whether it can boost the main performance. As shown in Tab.~\ref{tab:t2v_metrics} and Tab.~\ref{tab:multi_task}, using a temporal upsampler on the top of the main models can improve the quality performance while maintaining other scores.

\vspace{0.2cm}

\noindent\textbf{Multi-View Generation}
Multi-view generation is a specialized task focused on creating novel views from a given input image. This task places demands on view consistency and can greatly benefit from a well-pretrained video generation model. By adapting a video generation model for multi-view generation, we can evaluate whether the pretraining has effectively captured underlying 3D information, which would enhance multi-view generation. 
\vspace{0.2cm}



\begin{figure*}[t]
    \centering
    \fcolorbox{darkgreen}{lightcream}{
        \parbox{\textwidth}{
            \centering \textbf{\Large \textcolor{darkgreen}{Multi-View Generation}}
        }
    }
    \\[1em]
    \includegraphics[height=0.9\textheight]{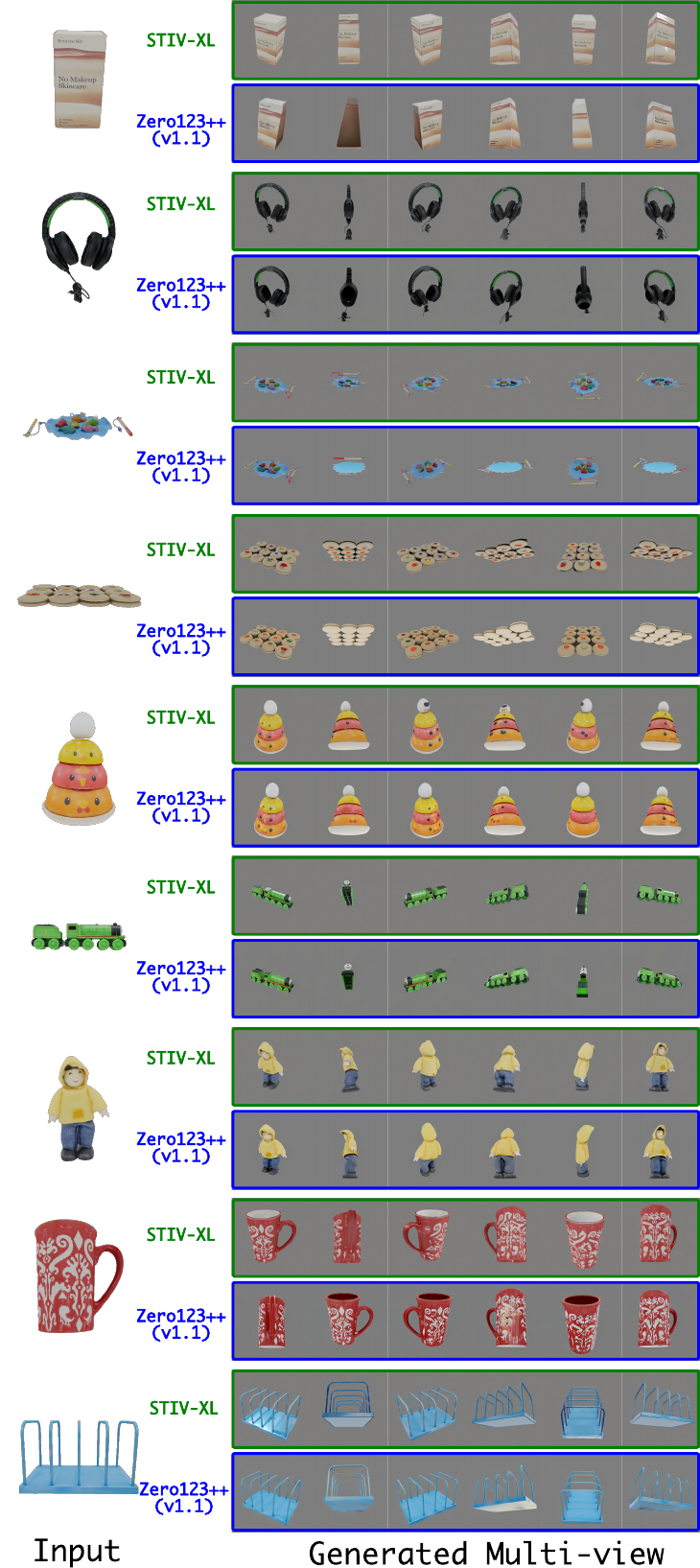}
    \caption{The visual comparison between our STIV-XL with Zero123++ \cite{shi2023zero123pp} on GSO \cite{downs2022google}.}
    \label{fig:multiview}
\end{figure*}

Here, we adopt the novel view camera definitions outlined in Zero123++ \cite{shi2023zero123pp}, which specifies six novel view cameras for each input image. The initial frame in our TI2V model is set as the given image, and the next six frames, representing novel views, are predicted as future frames within TI2V. For training, we began with our TI2V-XL checkpoint trained with a 256 resolution, fine-tuning it for 110k steps on Objaverse \cite{deitke2023objaverse}. For a fair comparison, we increased the image resolution to 320 during finetuning, aligning with the settings used in Zero123++. Our evaluation used objects from the Google Scanned Objects dataset \cite{downs2022google}, where we compared the output multi-view images against ground-truth renderings. As shown in Fig. \ref{tab:multiview}, despite only using temporal attention for cross-view consistency, our approach achieves comparable performance to Zero123++ which uses full attention to all the views. This outcome validates the effectiveness of our spatiotemporal attention in maintaining 3D consistency. A visual comparison between our approach and Zero123++ is shown in \cref{fig:multiview}.

\begin{figure*}[t!]
\resizebox{\linewidth}{!}{
\begin{minipage}{.29\textwidth}
\begin{subfigure}[b]{\linewidth}
\centering
\small
\resizebox{\linewidth}{!}{
\begin{tabular}{l|c|c}
\Xhline{1pt}
\multirow{2}{*}{\textbf{Model}} & \textbf{MSRVTT}  & \textbf{MovieGen} \\
& \textbf{FVD} $\downarrow$ & \textbf{FVD} $\downarrow$ \\
\Xhline{1pt}
T2V & 536.2 & 347.2 \\
STIV-V2V & \textbf{183.7} & \textbf{186.3} \\
\Xhline{1pt}
\end{tabular}
}
\caption{Comparison of T2V and V2V models.}
\label{tab:tv2v}
\end{subfigure}

\end{minipage}
\begin{minipage}{.31\textwidth}

\begin{subfigure}[b]{\linewidth}
\centering
\small
\resizebox{\linewidth}{!}{
\begin{tabular}{l|c|cc}
\Xhline{1pt}
\multirow{2}{*}{\textbf{Model}} & \multirow{2}{*}{\textbf{use text}}  & \multicolumn{2}{c}{\textbf{MSRVTT}} \\
& & \textbf{FID} $\downarrow$ & \textbf{FVD} $\downarrow$ \\
\Xhline{1pt}
STIV-TUP & No & 2.2 & 6.3 \\
STIV-TUP & Yes & \textbf{2.0} & \textbf{5.9} \\
\Xhline{1pt}
\end{tabular}
}
\caption{Performance of STIV-TUP.}
\label{tab:t_upsampler_performance}
\end{subfigure}

\end{minipage}
\begin{minipage}{.37\textwidth}

\begin{subfigure}[b]{\linewidth}
\centering
\small
\resizebox{\linewidth}{!}{
\begin{tabular}{l|ccc}
\Xhline{1pt}
\multirow{2}{*}{\textbf{Model}} & \multicolumn{3}{c}{\textbf{GSO} \cite{downs2022google}} \\
& \textbf{PSNR} $\uparrow$ & \textbf{SSIM} $\uparrow$ & \textbf{LPIPS} $\downarrow$ \\
\Xhline{1pt}
Zero123++ \cite{shi2023zero123pp} & 21.200 & 0.723 & \textbf{0.143} \\
STIV-TI2V-XL & \textbf{21.643} & \textbf{0.724} & 0.156 \\
\Xhline{1pt}
\end{tabular}
}
\caption{Multiview generation comparison.}
\label{tab:multiview}
\end{subfigure}
\end{minipage}
}
\caption{Demonstration of flexible applications of STIV framework.}
\end{figure*}

\vspace{0.2cm}

\begin{figure*}[htbp]
    \centering
    \fcolorbox{darkgreen}{lightcream}{
        \parbox{\textwidth}{
            \centering \textbf{\Large \textcolor{darkgreen}{Long Video Generation}}
        }
    }
    \\[1em]


    \centering
    \textbf{Prompt:} \textit{A drone camera circles around a beautiful historic church built on a rocky outcropping along the Amalfi Coast, the view showcases historic and magnificent architectural details and tiered pathways and patios...} \\[0.5em]
    \begin{minipage}{\textwidth}
        \centering
        \begin{minipage}[t]{0.18\textwidth}
            \centering
            \fcolorbox{blue}{white}{\includegraphics[width=\linewidth]{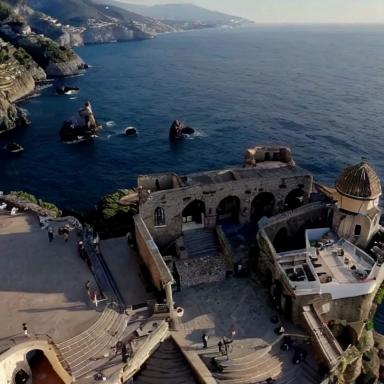}}
            \\[0.5em]
            \textbf{Reference Image}
        \end{minipage}
        \hfill
        \includegraphics[width=0.18\textwidth]{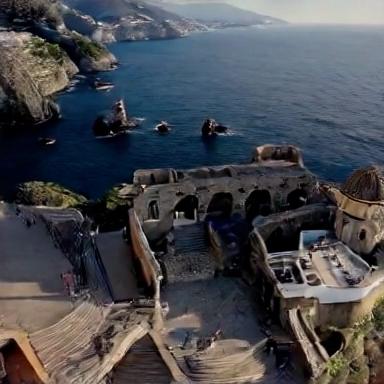}
        \includegraphics[width=0.18\textwidth]{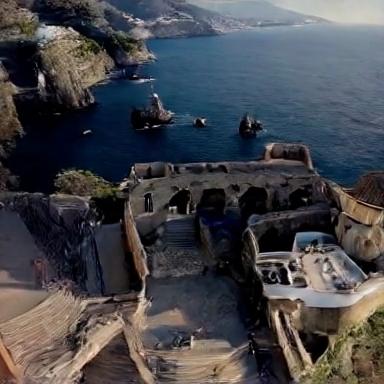}
        \includegraphics[width=0.18\textwidth]{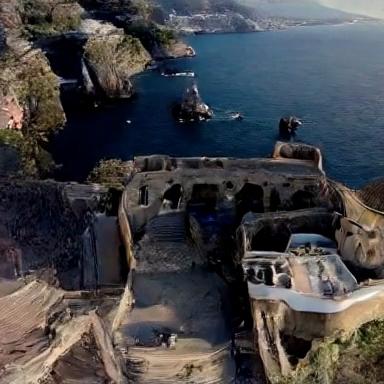}
        \includegraphics[width=0.18\textwidth]{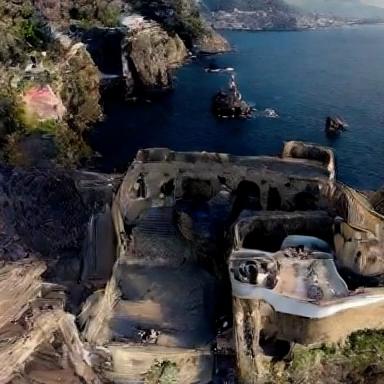}
    \end{minipage}
    \\[1em]

    \begin{minipage}{\textwidth}
        \hspace*{0.18\textwidth}
        \includegraphics[width=0.18\textwidth]{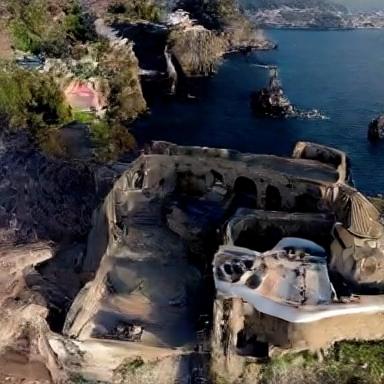}
        \includegraphics[width=0.18\textwidth]{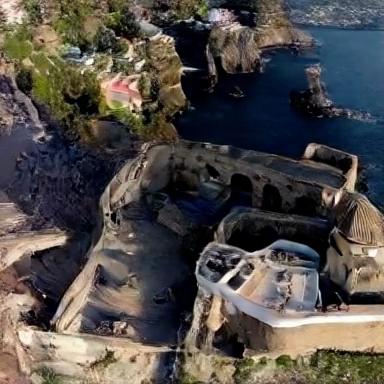}
        \includegraphics[width=0.18\textwidth]{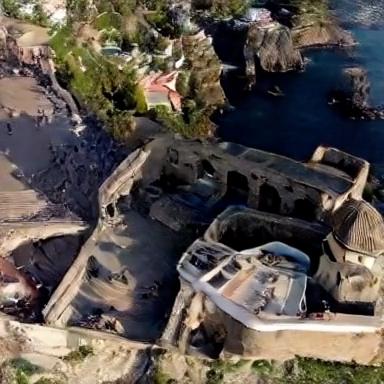}
        \includegraphics[width=0.18\textwidth]{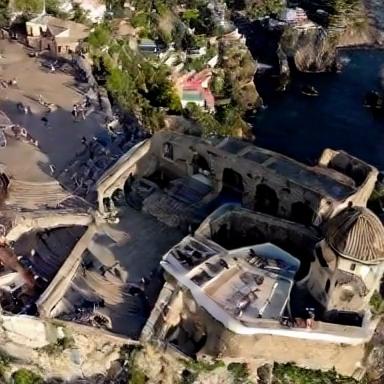}
    \end{minipage}
    \\[1.5em]

    \centering
    \textbf{Prompt:} \textit{A slow cinematic push in on an ostrich standing in a 1980s kitchen.} \\[0.5em]
    \begin{minipage}{\textwidth}
        \centering
        \includegraphics[width=0.18\textwidth]{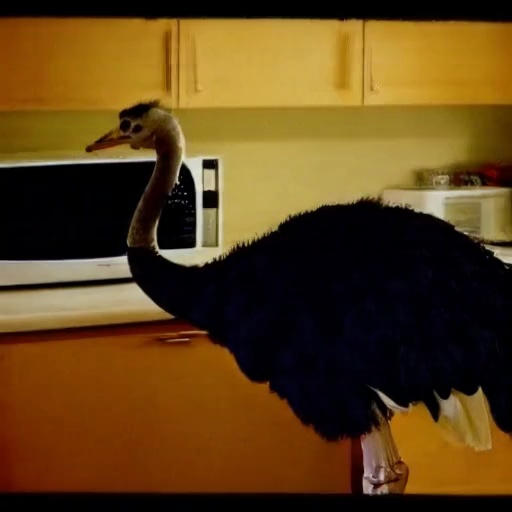}
        \includegraphics[width=0.18\textwidth]{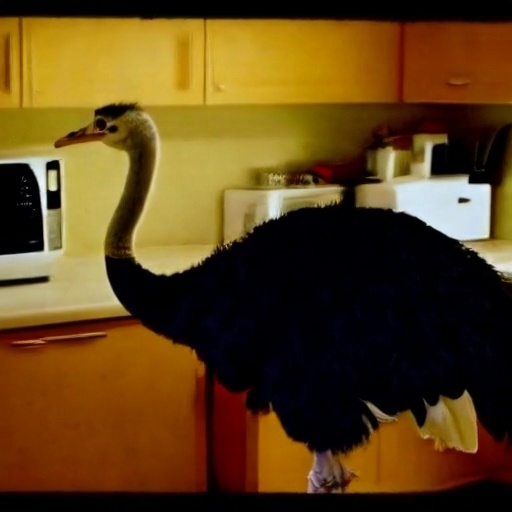}
        \includegraphics[width=0.18\textwidth]{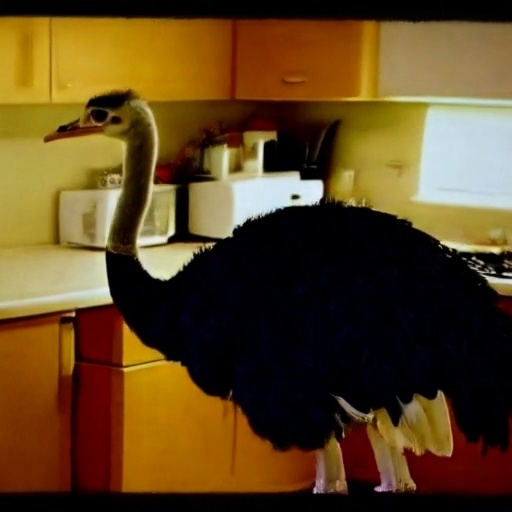}
        \includegraphics[width=0.18\textwidth]{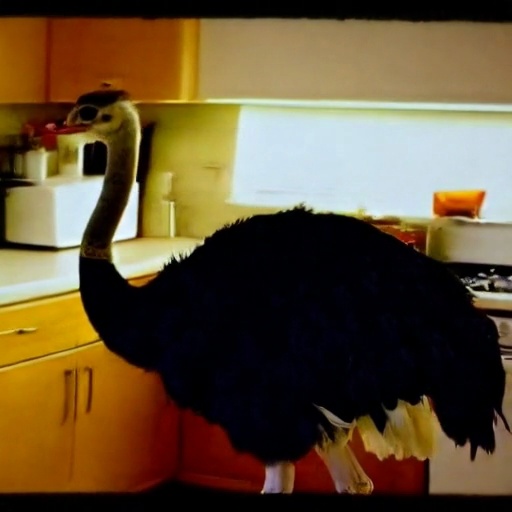}
        \includegraphics[width=0.18\textwidth]{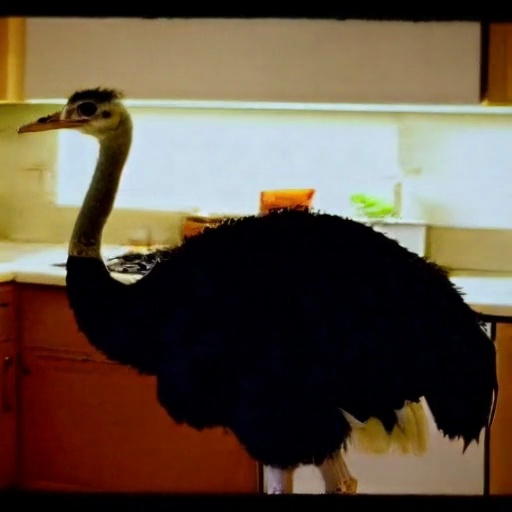}
    \end{minipage}
    \\[1em]

    \begin{minipage}{\textwidth}
        \hspace*{0.18\textwidth}
        \includegraphics[width=0.18\textwidth]{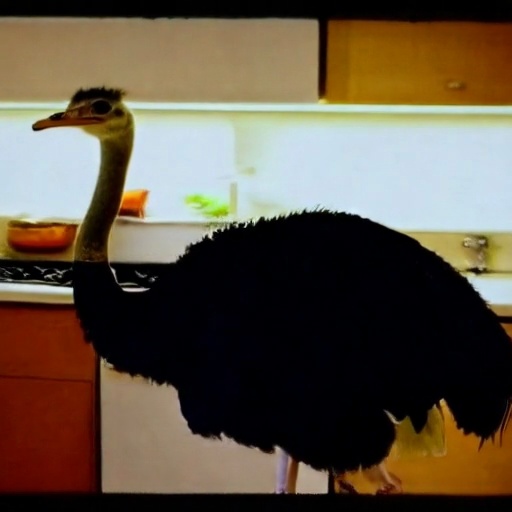}
        \includegraphics[width=0.18\textwidth]{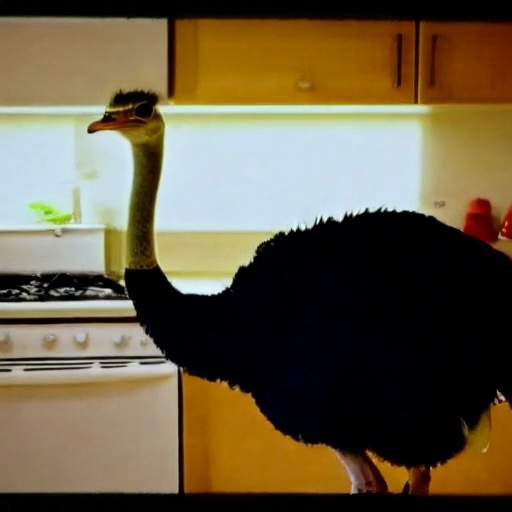}
        \includegraphics[width=0.18\textwidth]{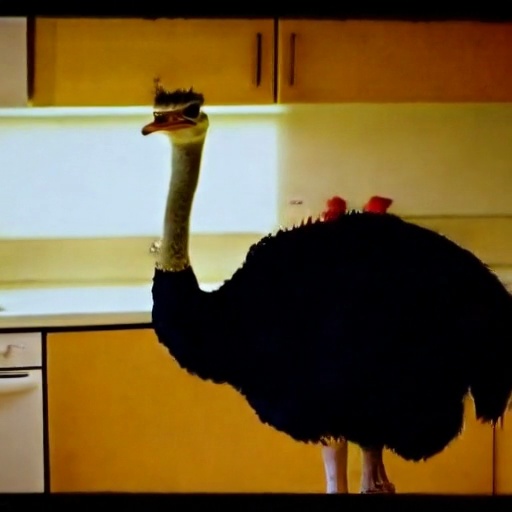}
        \includegraphics[width=0.18\textwidth]{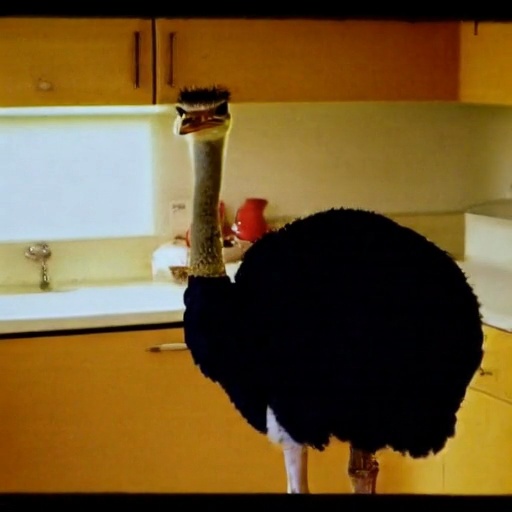}
    \end{minipage}
    \\[1em]

    \caption{Visualizations of long video generation framework.}
    \label{fig:long}
\end{figure*}

\noindent\textbf{Long Video Generation}
We develop an effective and efficient framework to generate long videos. Specifically, we propose a hierarchical framework, including training our STIV on two different modes: (1) key frame prediction by learning uniformly sampled video frames with stride of 20 with image conditioning dropout and (2) interpolated frame generation by learning consecutive video frames with first and last frame as image conditions. During sampling stage, we change the image and micro conditions, and first use the first mode to generate key frames and then generate the interpolated frames use the second mode, leading to a long video.
It is natural to reuse the STIV model to autoregressively generate the videos conditioned on previous generated one. However, in practice, we found this rollout approach can be hurt by error propagation in the previous video, and lacks some global consistency between frames. Therefore, we propose a simple yet effective baseline, purely based on our STIV framework. 
As mentioned in the main text, we design a hierarchical framework, train our STIV on two different modes: (1) key frame prediction by learning uniformly sampled video frames with a stride of 20 with image conditioning dropout, and (2) interpolated frame generation by learning consecutive video frames with the first and last frame as image conditions. During the sampling stage, we change the image and micro conditions, and first use the first mode to generate key frames and then generate the interpolated frames using the second mode, leading to a long video. We showcase one long T2V and one TI2V example in Figure~\ref{fig:long}; we achieve \((20-1)\times20=380\) frames in total. We uniformly sampled 8 frames out of the 380 frames.
Noted that this is only an early exploration of long video generation, and we do not have many long enough videos in our training distribution, so we leave it as future work to further explore the architecture to boost long video synthesis.

\clearpage
\section{Related Work}
\label{sec:relatedwork}

\noindent\textbf{Text-To-Video Generation}
In recent years, diffusion-based methods have emerged as the dominant approach in text-to-video generation, both for close-source models ~\citep{openai2024sora,pika2023pika,polyak2024movie} and open-source models~\citep{yang2024cogvideox, zheng2024open-sora}. \citep{he2022latent, chen2023videocrafter1, chen2024videocrafter2} leverages latent diffusion models (LDMs)~\citep{rombach2022high} to enhance training efficiency. VideoLDM~\citep{blattmann2023align} integrates temporal convolution and attention mechanisms into the LDM U-Net for video generation. Recently, there has been a shift from UNet to diffusion transformer-based architectures~\citep{polyak2024movie, yang2024cogvideox, zhou2024allegro, gao2024lumina}. CogVideoX ~\citep{yang2024cog-video-x} adopts the framework from SD3~\citep{esser2024scaling} to incorporate self-attention on the entire 3D video sequence with text conditions. Lumina-T2X~\citep{lu2024unified} employs zero-init attention to transform noise into different modalities. In contrast to previous models, our focus is to scale our diffusion transformer-based video generation model with spatial, temporal, and cross attention to over 8B parameters using various techniques. This model achieves good performance on VBench and serves as a strong baseline for the development of our text-image-to-video model: STIV.

\vspace{0.2cm}

\noindent\textbf{Text-Image-To-Video Generation} 
Controlling video content solely through text poses significant challenges in achieving satisfactory alignment between the video and the input text, as well as fine-grained control over the video generation process. To address this issue, recent approaches have integrated both the first frame and text to enhance control over video generation~\cite{zhang2023i2vgen,girdhar2023emu,xing2023dynamicrafter,chen2023videocrafter1,ren2024consisti2v}, mostly based on U-Net architecture. I2VGen-XL~\citep{zhang2023i2vgen} builds upon the SDXL and employs a cascading technique to generate high-resolution video. DynamiCrafter \citep{xing2023dynamicrafter} and VideoCrafter~\cite{chen2023videocrafter1} use cross-attention to incorporate image condition. ConsistentI2V~\citep{ren2024consisti2v} employs a similar frame replacement strategy, but it also requires spatial temporal attention over the initial frame and special noise initialization to enhance consistency.
Animate Anything \citep{dai2023fine} also employs the frame replacement technique, but it requires the use of motion strength loss to enhance the motion. However, their Dynamic Degree on VBench-I2V is relatively low, at 2.7\%.
We apply frame replacement on the DiT architecture, along with our proposed image condition dropout method, and JIT-CFG can generative high quality I2V videos while effectively addresses the motion staleness issue.
\section{Conclusion}
\label{sec:conclusion}
In conclusion, we conduct a comprehensive study on how to build a good video generation model, and present a scalable and flexible approach for integrating text and image conditioning within a unified video generation framework. Our model not only demonstrates good performance on public benchmarks, but also shows versatility in downstream applications, supporting controllable video generation, video prediction, frame interpolation, long video generation, and multi-view generation, which collectively highlight its potential as a foundation for the broad research community.
\section{Acknowledgement}
We thank Yifan Jiang,Alex Schwing, Monica Zuendorf, Saeed Khorram, Pengsheng Guo and Zhe Gan for their regular discussions on model design and training recipes. We also acknowledge the indirect influence of these discussions on our approach. Furthermore, we thank Meng Cao for data and Ruoming Pang for training infrastructure support. We also acknowledge the leadership guidance from Yang Zhao throughout the project. Finally, we thank the Axlearn team \footnote{\href{https://github.com/apple/axlearn}{https://github.com/apple/axlearn}} at Apple for providing training infrastructure, which greatly facilitated our experiments.
\newpage
{
    \small
    \bibliographystyle{ieeenat_fullname}
}

\clearpage
\setcounter{page}{1}
\maketitlesupplementary

\appendix

\section{Joint Image-Text Classifier-free Guidance}
We introduce a novel framework, Joint Image-Text Classifier-Free Guidance (JIT-CFG), in Section \ref{sec:image_cond}, which facilitates the seamless integration of text and image conditions to enhance the modeling performance. 
This is accomplished through a modified velocity estimate, expressed as:
\[
    \begin{split}
\hat{\bm{F}}_\theta(\bm{x}_t,\bm{c}_T,   & \bm{c}_I, t)= {\bm{F}}_\theta(\bm{x}_t, \varnothing, \varnothing, t) \\
& \quad + \omega \cdot \left( \bm{F}_\theta(\bm{x}_t, \bm{c}_T, \bm{c}_I, t)  - {\bm{F}}_\theta(\bm{x}_t, \varnothing, \varnothing, t) \right)
\end{split}
\]
The approach employs text and image condition dropout, which is also critical for unifying T2V and TI2V tasks.
\vspace{0.2cm}

\noindent\textbf{Probability mass shift} Our model learns $P(\bm{x}|\bm{c}_T, \bm{c}_I)$, the probability distribution of generating video $\bm{x}$ given the text prompt $\bm{c}_T$ and image condition $\bm{c}_I$. Here, we demonstrate how JIT-CFG shifts the probability mass toward regions of higher likelihood, conditioned on $\bm{c}_T$ and $\bm{c}_I$. First, consider a score-matching model with JIT-CFG 
\[
    \begin{split}
\hat{\bm{s}}_\theta(\bm{x}_t,\bm{c}_T,   & \bm{c}_I, t)= {\bm{s}}_\theta(\bm{x}_t,  t) \\
& \quad + \omega \cdot \left( \bm{s}_\theta(\bm{x}_t, \bm{c}_T, \bm{c}_I, t)  - {\bm{s}}_\theta(\bm{x}_t, t) \right)
\end{split}
\]
Using the definition of score and \textit{Bayes' Rule}, we derive
\[
    \begin{split}
&\hat{\bm{s}}_\theta(\bm{x}_t,\bm{c}_T,   \bm{c}_I, t) \\=&
\nabla logP_t(\bm{x}_t)
  + \omega \cdot \left( \nabla logP_t(\bm{x}_t, \bm{c}_T, \bm{c}_I)  - \nabla logP_t(\bm{x}_t) \right)\\=&
  \nabla logP_t(\bm{x}_t)
  + \omega \cdot \nabla logP(\bm{c}_T, \bm{c}_I|\bm{x}_t) \\=&
  \nabla log \left(P_t(\bm{x}_t)P_t^{\omega}(\bm{c}_T, \bm{c}_I|\bm{x}_t) \right) \\=&
   \nabla log \left(P_t^{1-\omega}(\bm{x}_t)P_t^{\omega}(\bm{x}_t|\bm{c}_T, \bm{c}_I) \right)
  ,
\end{split}
\]
where $\omega$ determines the influence of the text and image conditions during sampling from the tempered distribution. For a flow-matching model employing linear interpolants, the velocity and score are related as \cite{ma2024sit}:

\[
 \begin{split}
\hat{\bm{s}}_\theta(\bm{x}_t,\bm{c}_T,   \bm{c}_I, t) = \frac{t}{1-t}\hat{\bm{F}}_\theta(\bm{x}_t,\bm{c}_T,   & \bm{c}_I, t)-\frac{1}{1-t}\bm{x}_t,
\end{split}
\]
It implies $\frac{\partial\hat{\bm{s}}}{\partial\hat{\bm{F}}}=\frac{t}{1-t}>0$,  meaning that the JIT-CFG-guided velocity $\hat{\bm{F}}$ shifts the probability mass in alignment with the modified score $\hat{\bm{s}}$ by adjusting the tempered distribution.
\vspace{0.2cm}

\noindent\textbf{CFG-Renormalization}
\label{supp:renorm}
Empirically, we observed that the magnitude of the modified velocity, $||\hat{\bm{F}}_\theta(\bm{x}_t,\bm{c}_T,  \bm{c}_I, t)||$ tends to be very large during the early stages of integration in inference (i.e. when $t$ is small).
This behavior sometimes leads to overshooting beyond the learned latent distribution, resulting in artifacts in the generated output.
We identified this issue as primarily due to the significant difference between the conditional velocity, $\bm{F}_\theta(\bm{x}_t, \bm{c}_T, \bm{c}_I, t)$, and the unconditional velocity, ${\bm{F}}_\theta(\bm{x}_t, \varnothing, \varnothing, t)$ when $t$ is small.

To mitigate this, we propose a simple yet effective renormalization method that re-scales the magnitude of the modified velocity to $||\bm{F}_\theta(\bm{x}_t, \bm{c}_T, \bm{c}_I, t) ||$ while preserving its direction. Formally, this is defined as:

\[
\tilde{\bm{F}}_\theta(\bm{x}_t,\bm{c}_T,  \bm{c}_I, t)=||\bm{F}_\theta(\bm{x}_t, \bm{c}_T, \bm{c}_I, t) ||\frac{\hat{\bm{F}}_\theta(\bm{x}_t,\bm{c}_T,  \bm{c}_I, t)}{||\hat{\bm{F}}_\theta(\bm{x}_t,\bm{c}_T,  \bm{c}_I, t)||} 
\]
As shown in Table \ref{tab:t2i_metrics}, this technique significantly improves performance across various T2I evaluation benchmarks.

\section{Implementation Details for T2V and STIV}
Given that we use spatial-temporal attention, we first pretrain the T2I model using only an image dataset. Subsequently, we load the EMA weights from the T2I model, excluding the temporal attention.
In our work, we use the per-frame VAE, which is the same one used in the T2I model. On top of that, we use a temporal patch of size 2 in the DiT part for video models. We modify the T2I cubify weights by inflating the 3D convolution weight in the temporal dimension.
For video training data, we select one frame from every three frames and add independent and identically distributed Gaussian noise to each frame.
Following standard practice, we randomly replace text prompt with empty string 10\% during training. In our JIT-CFG setting, we also independently randomly drop image condition 8\% during training. For both T2V and STIV models, the JIT-CFG scale is set to 7.5. The training schedule follows the progressive training recipe described in section \ref{sec:model_scaling}.

\section{Implementation Details of Text Encoders} We used our internal CLIP text encoder to encode text into embeddings. Concretely, a text is first tokenized via a T5 tokenizer. The tokenized text is mapped into embeddings via an embedding lookup table and further encoded via 32 layers of transformer with casual attention. Each transformer layer contains 20 attention heads. Each attention head has 64 hidden dimensions. The output text embedding has a dimension of 1280.

\section{Ablation Study on Text-to-Image Generation}
\label{appendix:t2i_ablation}

\noindent\textbf{Baseline Setup}
For our base model, we employed the PixArt-$\alpha$ architecture \cite{chen2023pixart}, which builds on the DiT \cite{peebles2023scalable} model with added cross-attention layers to integrate image tokens with text embeddings.
As pre-trained components, we used the open-source sd-vae-ft-ema model\footnote{https://huggingface.co/stabilityai/sd-vae-ft-ema} and OpenAI CLIP L14 model\footnote{https://huggingface.co/openai/clip-vit-large-patch14}, both of which are widely adopted in the community. We conduct our experiments using the XL model configuration with a $256^2$ image size.
The full baseline model, which includes the VAE and CLIP text encoder, has approximately 1.06 billion parameters.
For noise generation and denoising, we used a diffusion-based approach with Stable Diffusion’s default noise schedule.
The training was conducted with a batch size of 4,096 over 400k steps, which corresponds to approximately 1.4 epochs on our internal text-to-image dataset.

Table \ref{tab:t2i_metrics} summarizes the results of our ablation study, focusing on the following aspects:
\vspace{0.2cm}

\noindent\textbf{Stabilized Training}
Leveraging recent advancements in LLM and diffusion model architectures, we integrated QK-Norm~\cite{henry2020qk-norm} to manage the activation scale within attention layers. Additionally, we applied sandwich-norm~\cite{gong2022sandwich-norm} to both the inputs and outputs of the attention layer and the feedforward layer.
Projected conditions, including timestep embeddings, pooled CLIP text embeddings, and micro condition embeddings, were normalized before being input to AdaLN.
These normalization techniques enhanced training stability, allowing us to increase the learning rate from \(1 \times 10^{-4}\) to \(2 \times 10^{-4}\), and also resulting in quality improvements.
\vspace{0.2cm}

\noindent\textbf{Noising and Denoising Process Formulation} We explored optimized noising/denoising formulations by replacing the diffusion process with a flow-based linear interpolant\cite{ma2024sit}. Additionally, we applied renormalization at each inference step to counteract potential side effects from high classifier-free guidance (CFG) values. Here, the norm of the prediction with CFG was linearly scaled to match the conditional prediction norm, as explained in Sec.~\ref{supp:renorm}.
\vspace{0.2cm}

\noindent\textbf{Training Cost Optimization} To reduce training costs, we evaluated three strategies: (1) switching from the AdamW optimizer to Adafactor, (2) applying MaskDiT training with a 50\% masking ratio, and (3) using a shared AdaLN module across layers instead of unique instances per layer. These changes reduced per-device HBM usage from approximately 28GB to 11GB, allowing us to train on v5e TPUs instead of the more costly v5p TPUs. Notably, as shown in Table \ref{tab:t2i_metrics}, masked training may adversely affect metrics such as FID and HPS. However, we found additional unmask finetuning for a short duration (e.g. 50k steps) can fix the artifacts causing these score drops. However, this additional training phase was not included in the final configuration, as further training on video generation can address this issue as well.
\vspace{0.2cm}

\noindent\textbf{Enhanced Pre-trained Models and Conditioning} We evaluated improvements from advanced pre-trained models and additional conditioning techniques. Specifically, we upgraded from the OpenAI CLIP L14 to an internally trained CLIP-bigG model \cite{lai2025veclip} and from a 4-channel to an 8-channel VAE. We also introduced 2D RoPE to support masked training and added micro-conditions, inspired by SDXL \cite{podell2023sdxl}, to mitigate cropping artifacts in elongated objects. Finally, synthetic captions generated via \cite{lai2024revisit} were included in our training data, resulting in notable performance gains.
\vspace{0.2cm}

\begin{figure}
    \centering
    \includegraphics[width=0.8\linewidth]{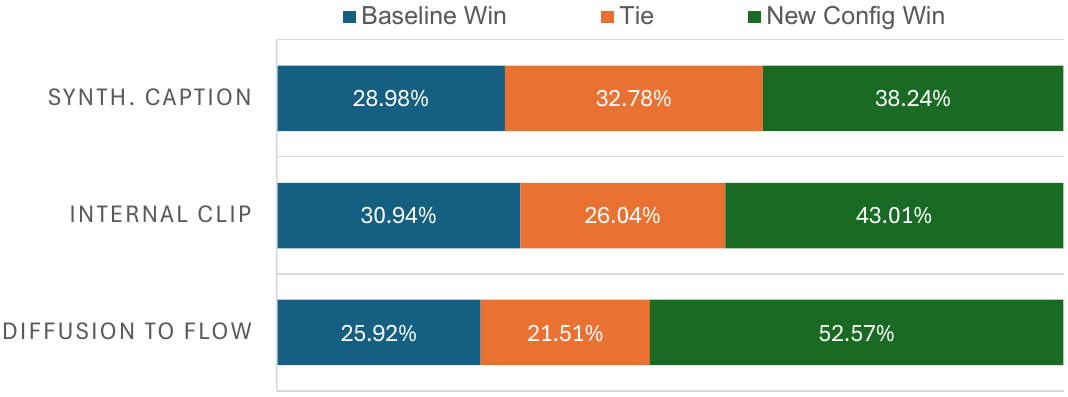}
    \caption{Human evaluation results on significant changes in T2I ablation study Tab.~\ref{tab:t2i_metrics}.}
    \label{fig:t2i_human_eval}
\end{figure}

    




\noindent\textbf{Human Evaluation of Model Changes} To validate improvements observed in automated metrics, we conducted human evaluations for key modifications, including the addition of synthetic captions, upgrade of CLIP model, and transition from diffusion to flow matching based objective. Human raters are asked to asses image fidelity, text-image alignment, and visual appeal, and give 5 level preference ratings for image pairs. Each pair is sent to 5 raters for rating and the image pair will be considered tie of combined voting is neutral. Results from Figure \ref{fig:t2i_human_eval} demonstrate clear alignment between automated metrics and human judgments. This justifies our usage of automatic evaluation as daily development metrics to maintain generation quality and prevent regressions leading to significant quality losses.

\section{Detailed Results for Imaging Dropout}
\begin{figure*}[htbp]
    \centering
    \fcolorbox{darkgreen}{lightcream}{
        \parbox{\textwidth}{
            \centering \textbf{\Large \textcolor{darkgreen}{STIV-M-512 V.S. STIV-M-512-JIT}}
        }
    }
    \\[1em]

    \centering
    \textbf{Prompt:} \textit{A red panda taking a bite of a pizza.} \\[0.5em]
    \begin{minipage}{\textwidth}
        \centering
        \begin{minipage}[t]{0.18\textwidth}
            \centering
            \fcolorbox{blue}{white}{\includegraphics[width=\linewidth]{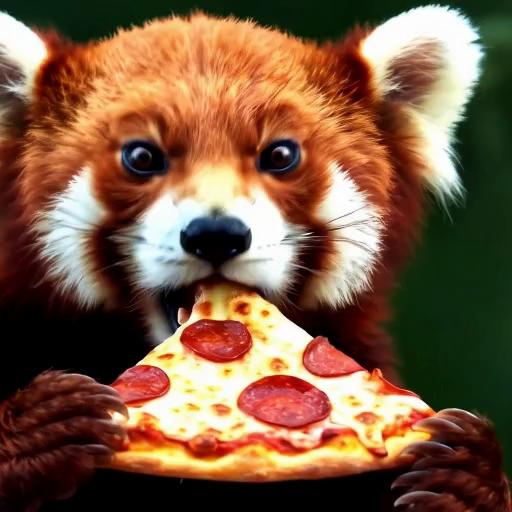}}
            \\[0.5em]
            \textbf{Reference Image}
        \end{minipage}
        \hfill
        \includegraphics[width=0.18\textwidth]{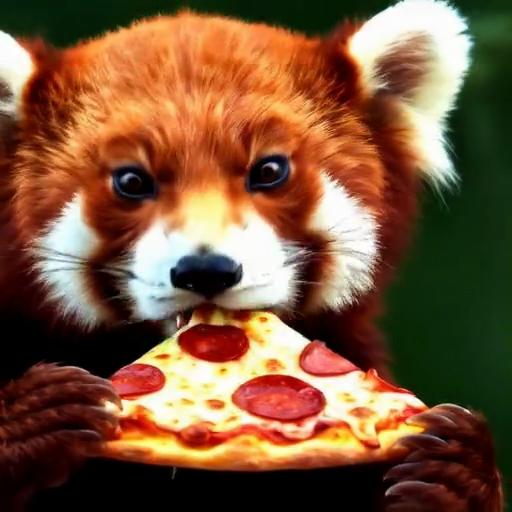}
        \includegraphics[width=0.18\textwidth]{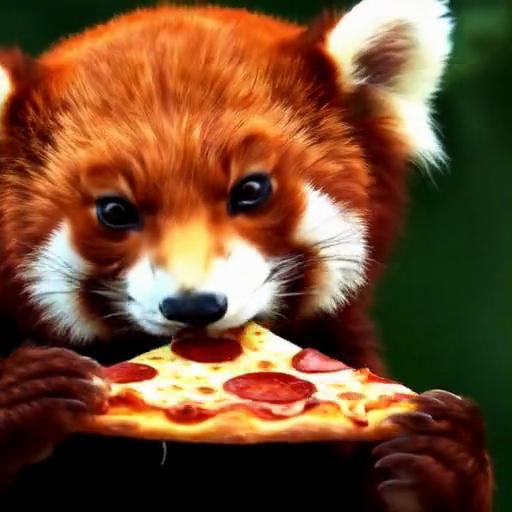}
        \includegraphics[width=0.18\textwidth]{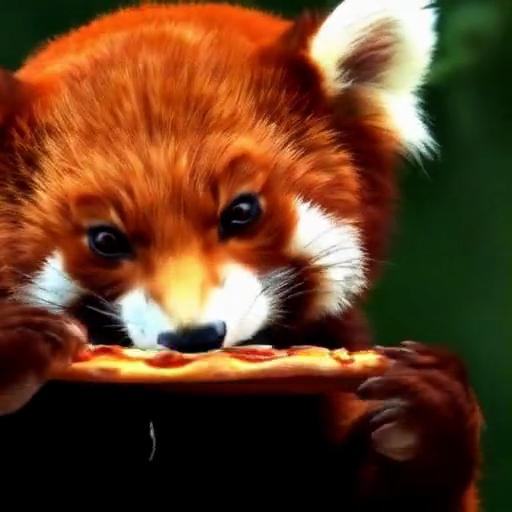}
        \includegraphics[width=0.18\textwidth]{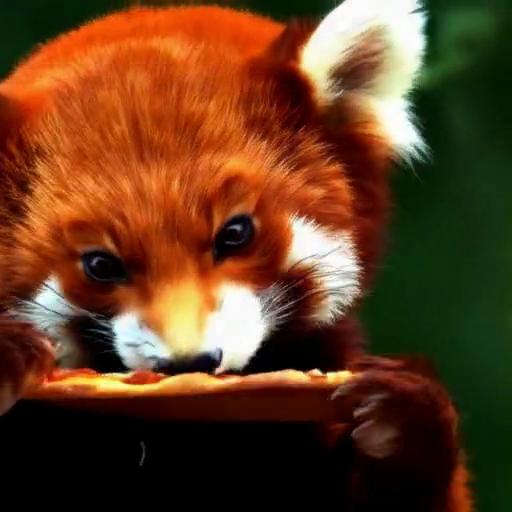}
    \end{minipage}
    \\[1em]

    \centering
    \begin{minipage}{\textwidth}
        \centering
        \begin{minipage}[t]{0.18\textwidth}
            \centering
            \fcolorbox{blue}{white}{\includegraphics[width=\linewidth]{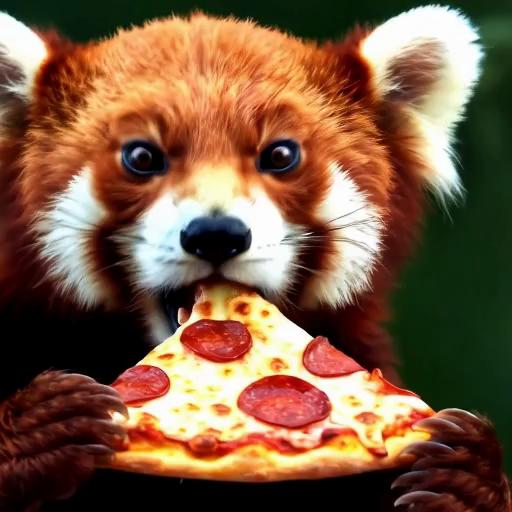}}
            \\[0.5em]
            \textbf{Reference Image}
        \end{minipage}
        \hfill
        \includegraphics[width=0.18\textwidth]{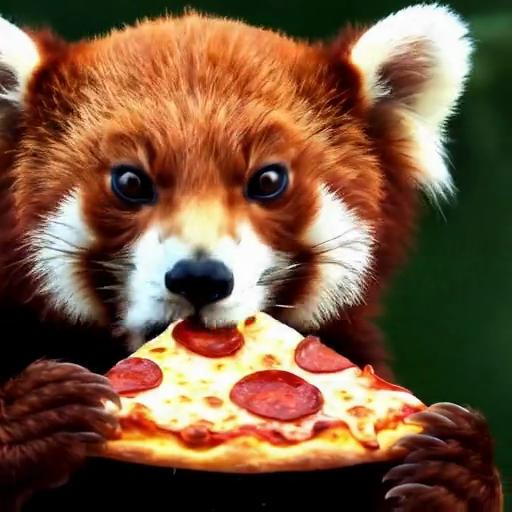}
        \includegraphics[width=0.18\textwidth]{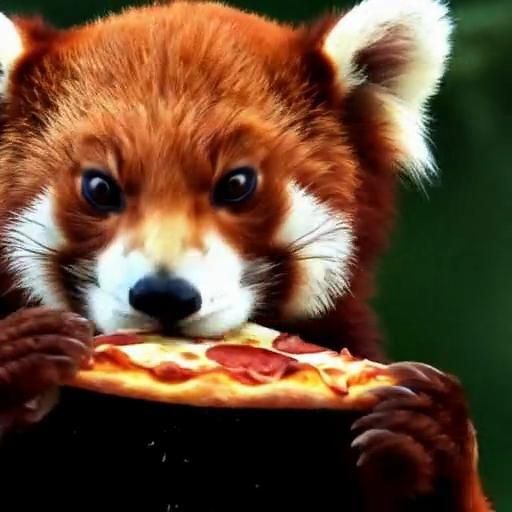}
        \includegraphics[width=0.18\textwidth]{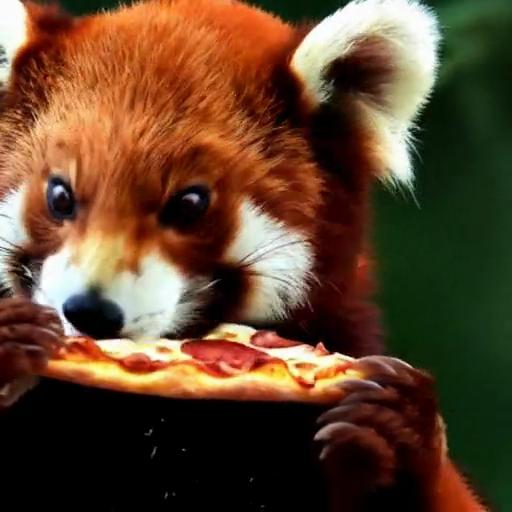}
        \includegraphics[width=0.18\textwidth]{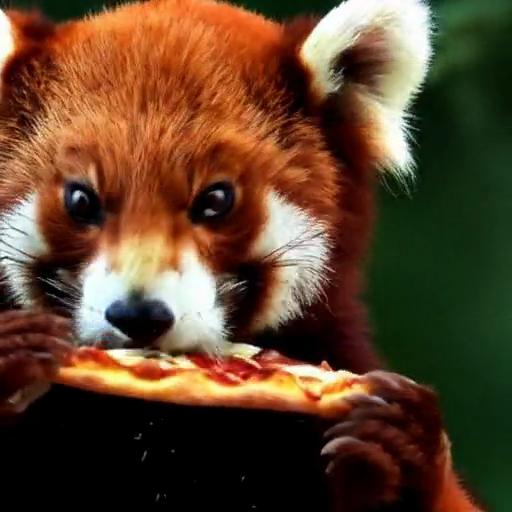}
    \end{minipage}
    \\[1em]

    \centering
    \textbf{Prompt:} \textit{A rocket blasting off from the launch pad, accelerating rapidly into the sky.} \\[0.5em]
    \begin{minipage}{\textwidth}
        \centering
        \begin{minipage}[t]{0.18\textwidth}
            \centering
            \fcolorbox{blue}{white}{\includegraphics[width=\linewidth]{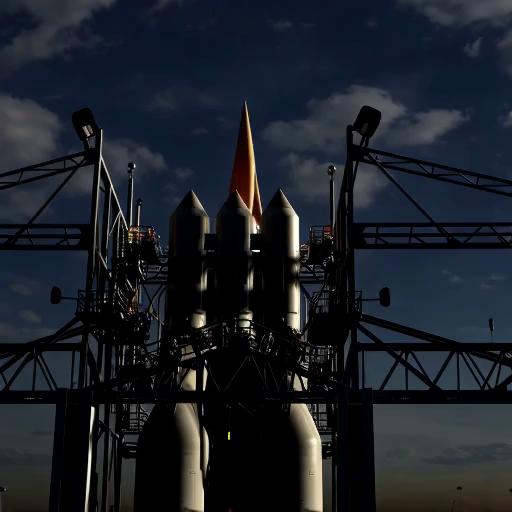}}
            \\[0.5em]
            \textbf{Reference Image}
        \end{minipage}
        \hfill
        \includegraphics[width=0.18\textwidth]{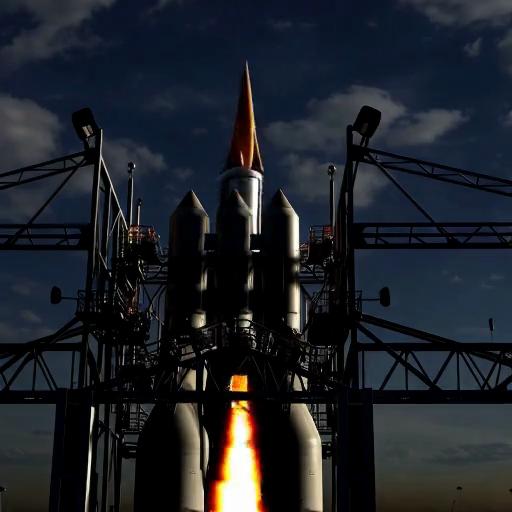}
        \includegraphics[width=0.18\textwidth]{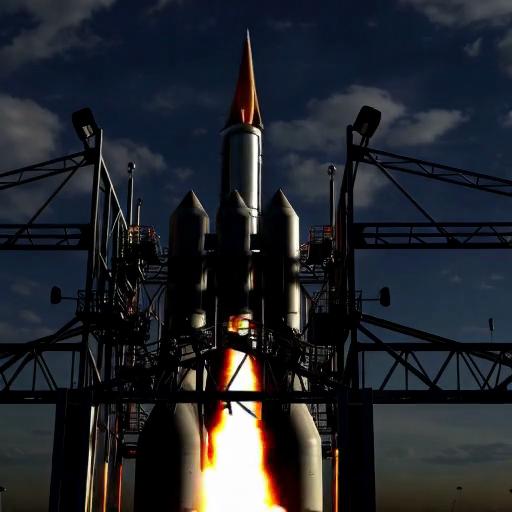}
        \includegraphics[width=0.18\textwidth]{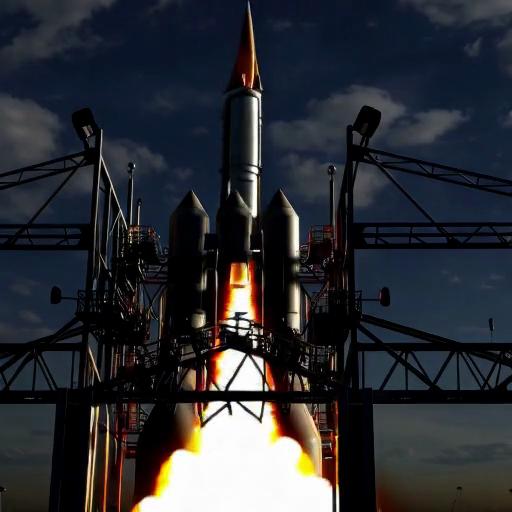}
        \includegraphics[width=0.18\textwidth]{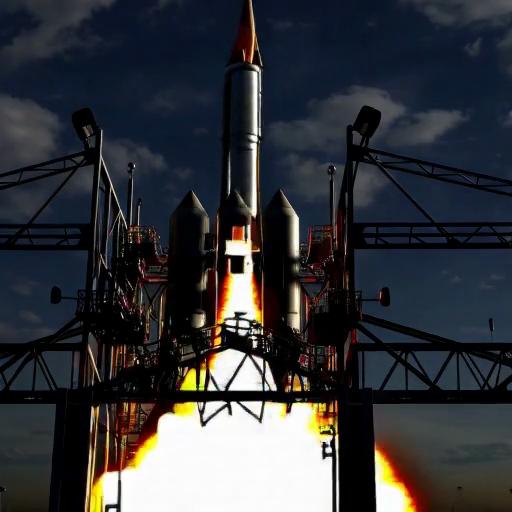}
    \end{minipage}
    \\[1em]

    \centering
    \begin{minipage}{\textwidth}
        \centering
        \begin{minipage}[t]{0.18\textwidth}
            \centering
            \fcolorbox{blue}{white}{\includegraphics[width=\linewidth]{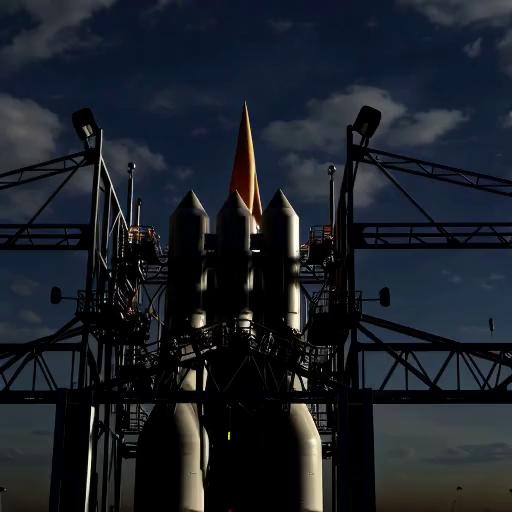}}
            \\[0.5em]
            \textbf{Reference Image}
        \end{minipage}
        \hfill
        \includegraphics[width=0.18\textwidth]{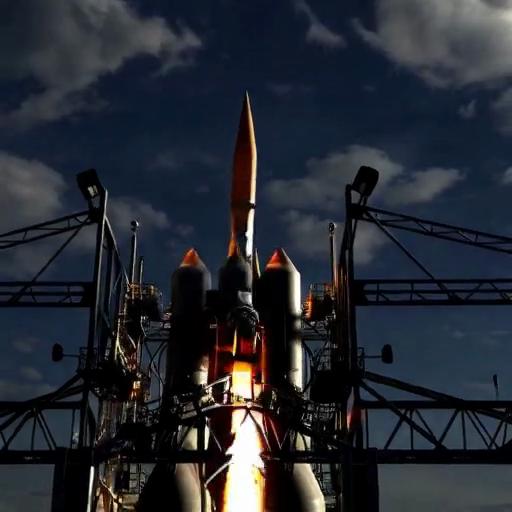}
        \includegraphics[width=0.18\textwidth]{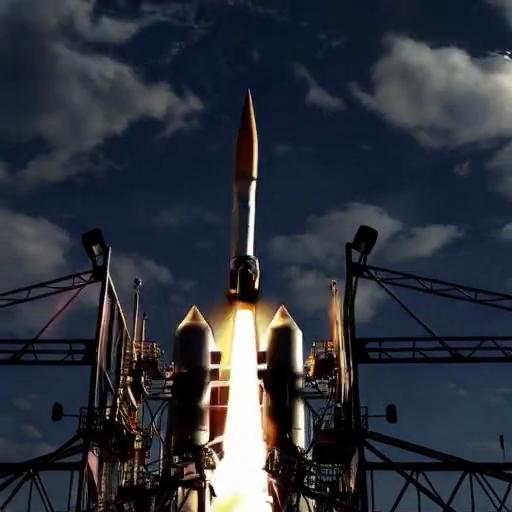}
        \includegraphics[width=0.18\textwidth]{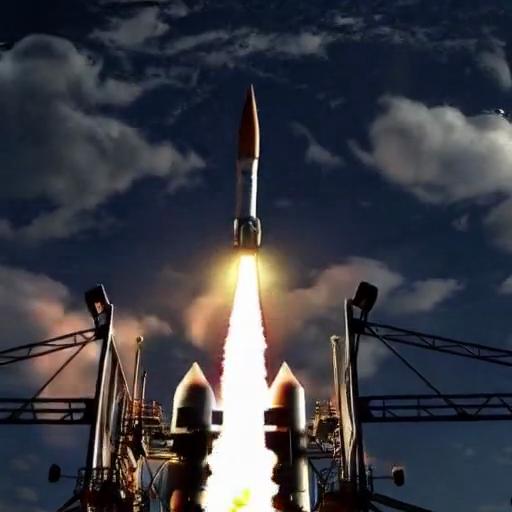}
        \includegraphics[width=0.18\textwidth]{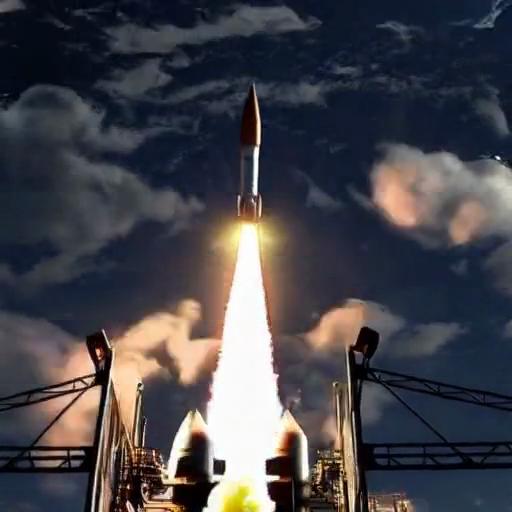}
    \end{minipage}
    \\[1em]
\caption{Visualization of STIV-M-512 V.S. STIV-M-512-JIT. (Given the same prompt, the figures in the top row are generated by STIV-M-512, while the figures in the bottom row are generated by STIV-M-512-JIT.)}
\label{fig:dpnodp1}
\end{figure*}

\begin{figure*}[htbp]
    \centering
    \centering
    \fcolorbox{darkgreen}{lightcream}{
        \parbox{\textwidth}{
            \centering \textbf{\Large \textcolor{darkgreen}{STIV-M-512 V.S. STIV-M-512-JIT}}
        }
    }
    \\[1em]

    \centering
    \textbf{Prompt:} \textit{A sports car accelerating rapidly on an open highway, the engine roaring.} \\[0.5em]
    \begin{minipage}{\textwidth}
        \centering
        \begin{minipage}[t]{0.18\textwidth}
            \centering
            \fcolorbox{blue}{white}{\includegraphics[width=\linewidth]{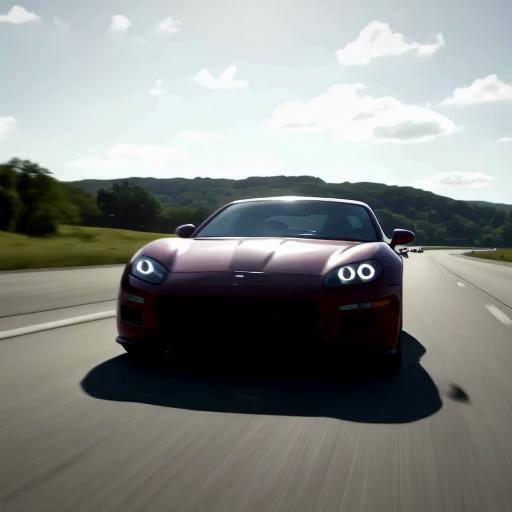}}
            \\[0.5em]
            \textbf{Reference Image}
        \end{minipage}
        \hfill
        \includegraphics[width=0.18\textwidth]{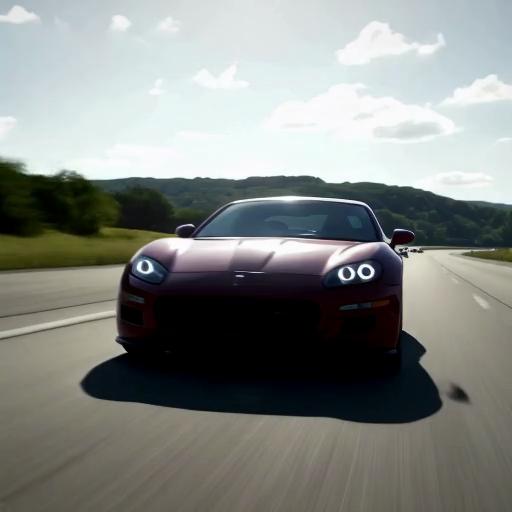}
        \includegraphics[width=0.18\textwidth]{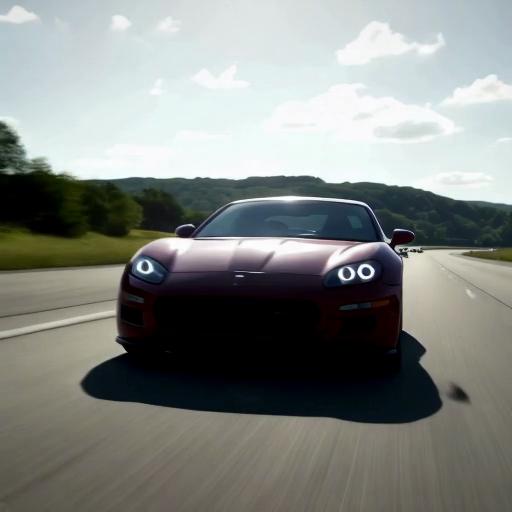}
        \includegraphics[width=0.18\textwidth]{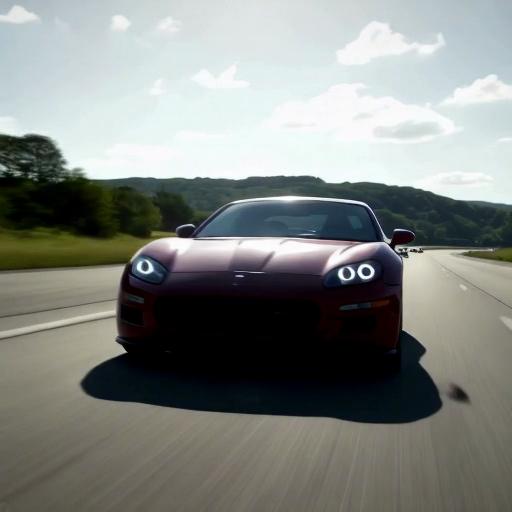}
        \includegraphics[width=0.18\textwidth]{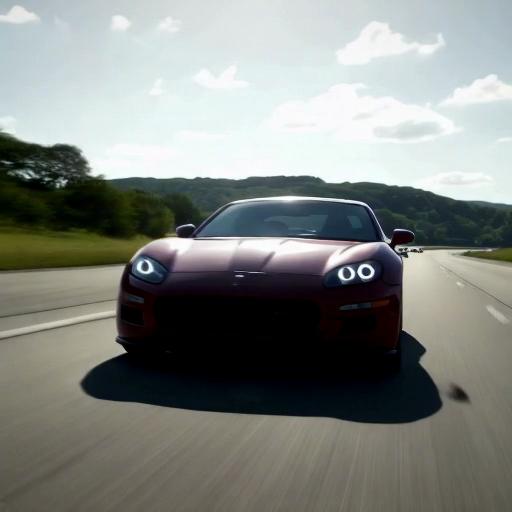}
    \end{minipage}
    \\[1em]

    \centering
    \begin{minipage}{\textwidth}
        \centering
        \begin{minipage}[t]{0.18\textwidth}
            \centering
            \fcolorbox{blue}{white}{\includegraphics[width=\linewidth]{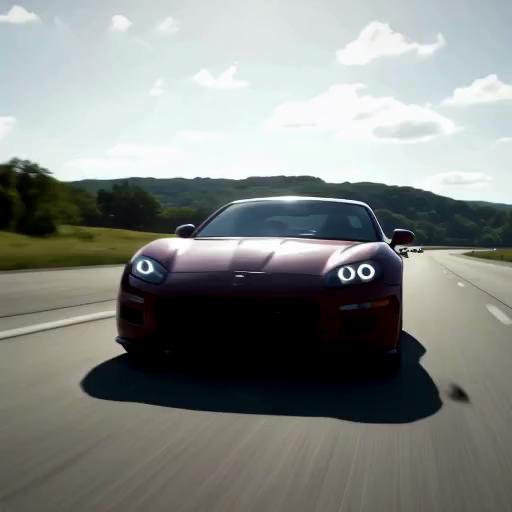}}
            \\[0.5em]
            \textbf{Reference Image}
        \end{minipage}
        \hfill
        \includegraphics[width=0.18\textwidth]{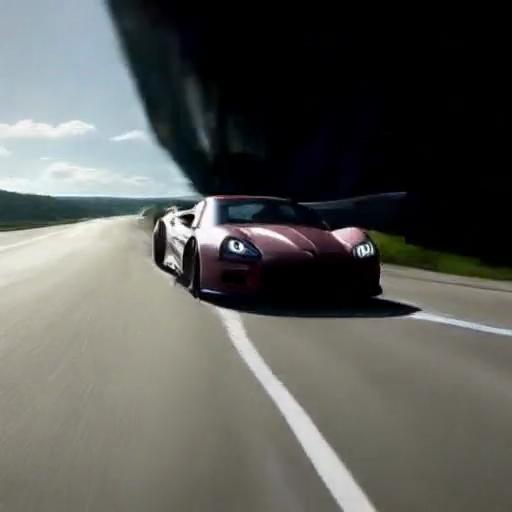}
        \includegraphics[width=0.18\textwidth]{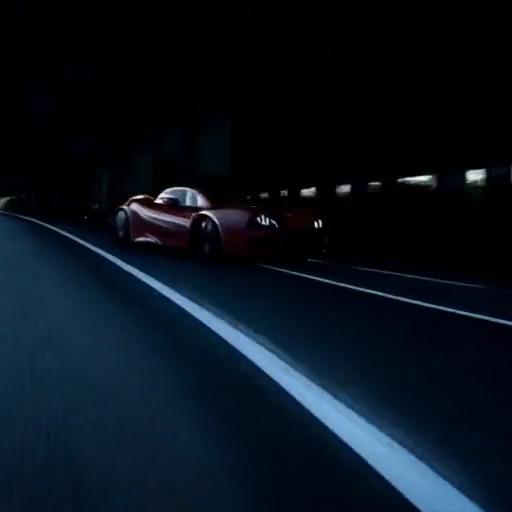}
        \includegraphics[width=0.18\textwidth]{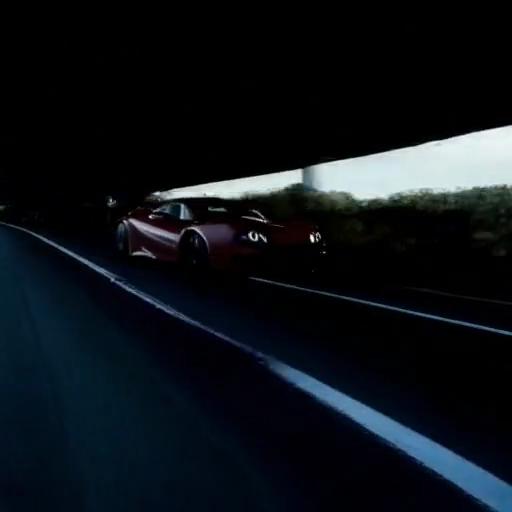}
        \includegraphics[width=0.18\textwidth]{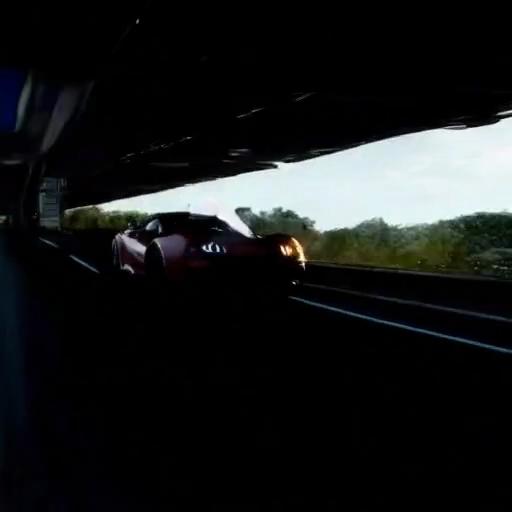}
    \end{minipage}
    \\[1em]

    \centering
    \textbf{Prompt:} \textit{A glass of iced coffee condensing water on the outside, with droplets forming and sliding down the glass in slow motion.} \\[0.5em]
    \begin{minipage}{\textwidth}
        \centering
        \begin{minipage}[t]{0.18\textwidth}
            \centering
            \fcolorbox{blue}{white}{\includegraphics[width=\linewidth]{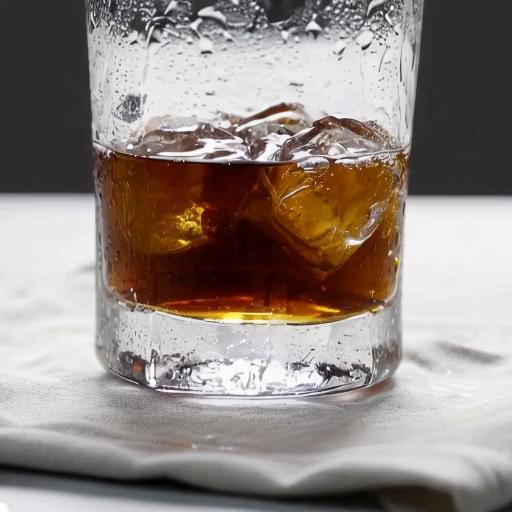}}
            \\[0.5em]
            \textbf{Reference Image}
        \end{minipage}
        \hfill
        \includegraphics[width=0.18\textwidth]{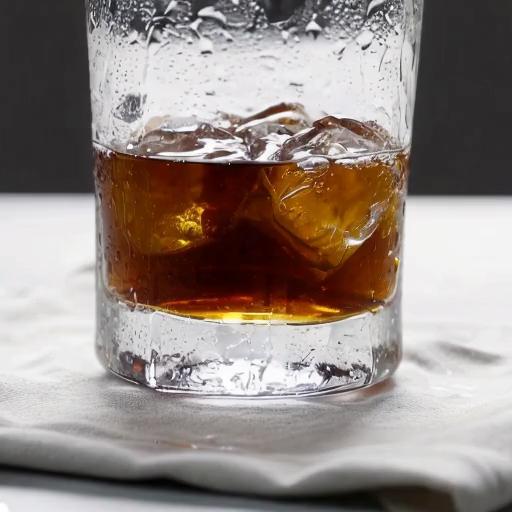}
        \includegraphics[width=0.18\textwidth]{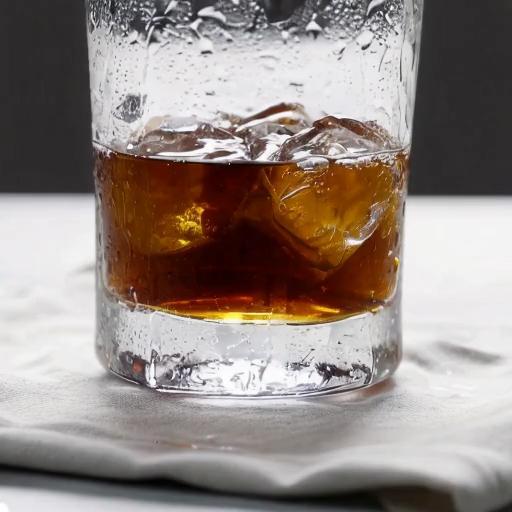}
        \includegraphics[width=0.18\textwidth]{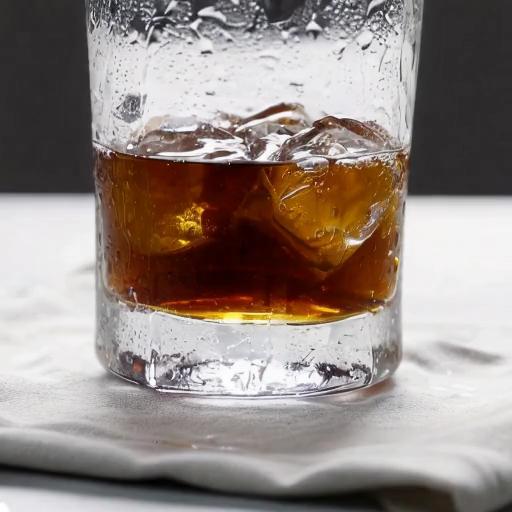}
        \includegraphics[width=0.18\textwidth]{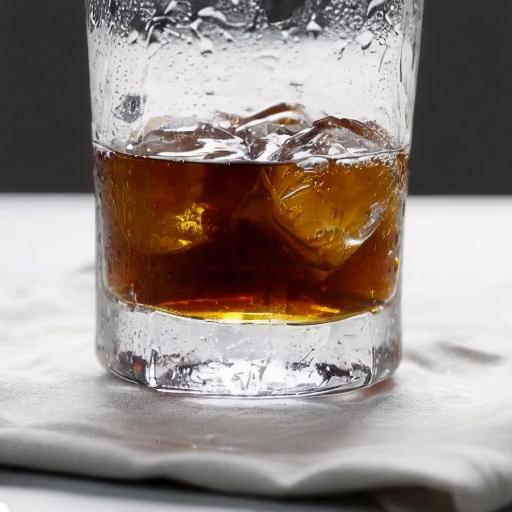}
        
    \end{minipage}
    \\[1em]

    \centering
    \begin{minipage}{\textwidth}
        \centering
        \begin{minipage}[t]{0.18\textwidth}
            \centering
            \fcolorbox{blue}{white}{\includegraphics[width=\linewidth]{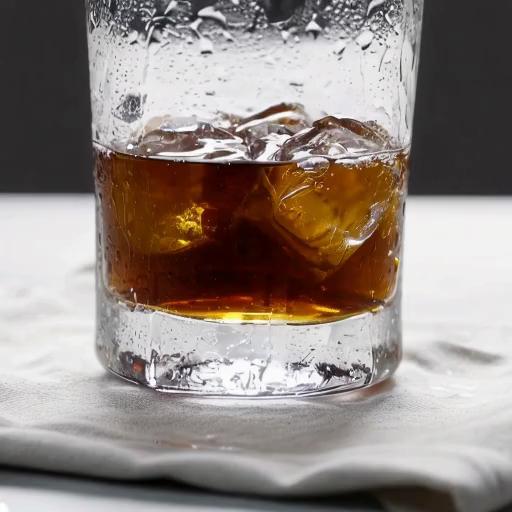}}
            \\[0.5em]
            \textbf{Reference Image}
        \end{minipage}
        \hfill
        \includegraphics[width=0.18\textwidth]{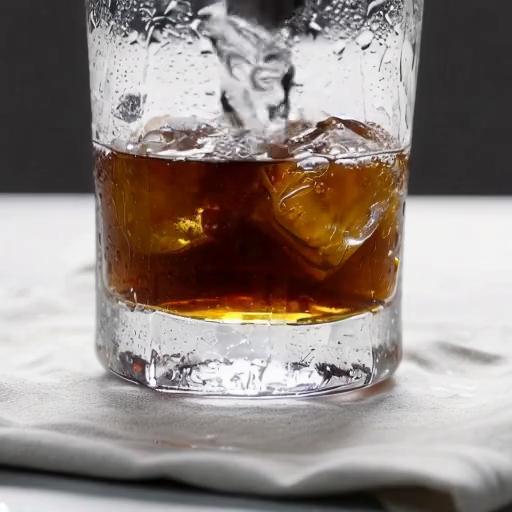}
        \includegraphics[width=0.18\textwidth]{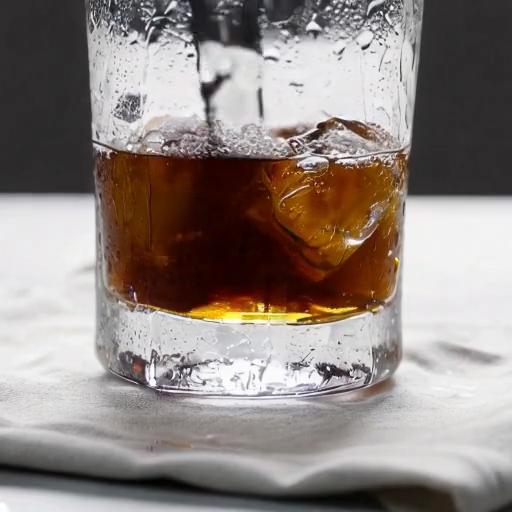}
        \includegraphics[width=0.18\textwidth]{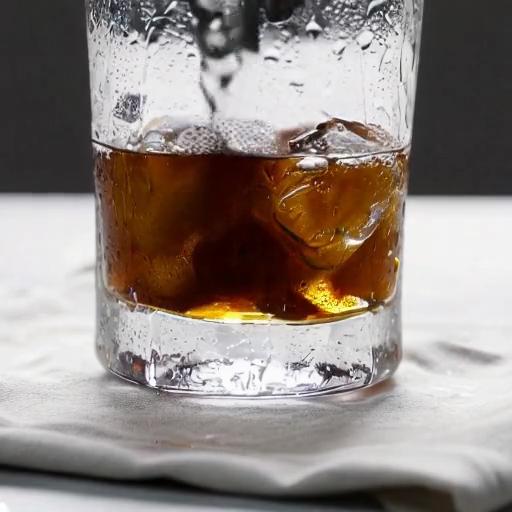}
        \includegraphics[width=0.18\textwidth]{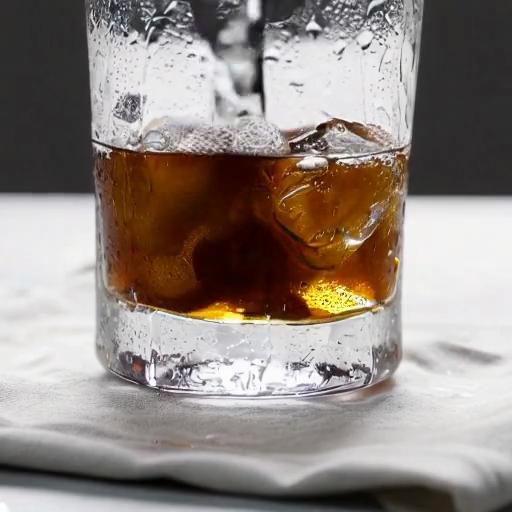}
        
    \end{minipage}
    \\[1em]
\caption{Visualization of STIV-M-512 V.S. STIV-M-512-JIT. (Given the same prompt, the figures in the top row are generated by STIV-M-512, while the figures in the bottom row are generated by STIV-M-512-JIT.)}
\label{fig:dpnodp2}
\end{figure*}

\begin{figure*}[htbp]
    \centering
    \centering
    \fcolorbox{darkgreen}{lightcream}{
        \parbox{\textwidth}{
            \centering \textbf{\Large \textcolor{darkgreen}{STIV-M-512 V.S. STIV-M-512-JIT}}
        }
    }
    \\[1em]

    \centering

    \centering
    \textbf{Prompt:} \textit{Cars and pedestrians move through a bustling downtown street lined with skyscrapers, their lights reflecting off the windows of the towering buildings as day turns to dusk.} \\[0.5em]
    \begin{minipage}{\textwidth}
        \centering
        \begin{minipage}[t]{0.18\textwidth}
            \centering
            \fcolorbox{blue}{white}{\includegraphics[width=\linewidth]{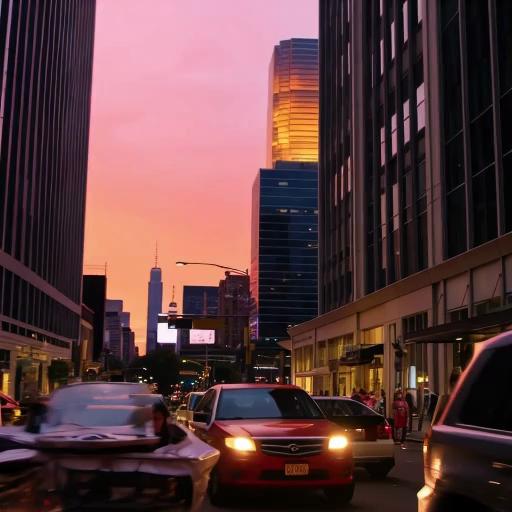}}
            \\[0.5em]
            \textbf{Reference Image}
        \end{minipage}
        \hfill
        \includegraphics[width=0.18\textwidth]{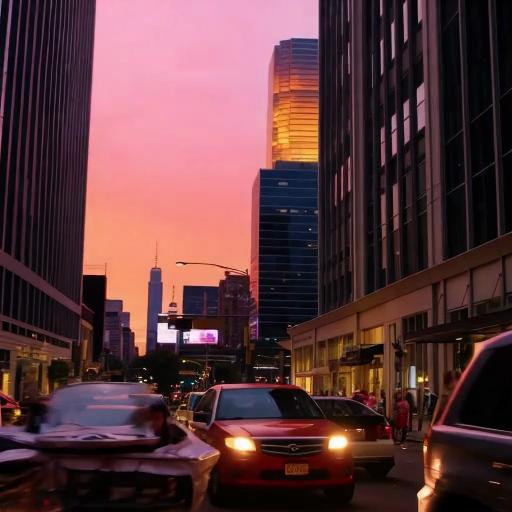}
        \includegraphics[width=0.18\textwidth]{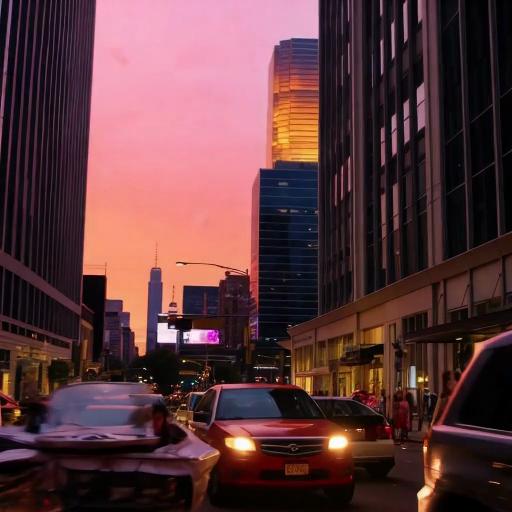}
        \includegraphics[width=0.18\textwidth]{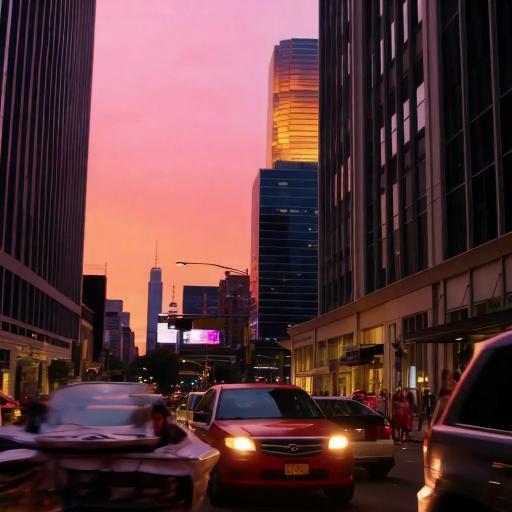}
        \includegraphics[width=0.18\textwidth]{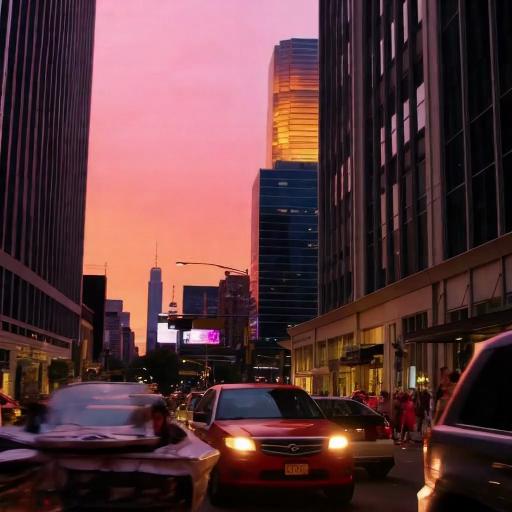}
    \end{minipage}
    \\[1em]

    \centering
    \begin{minipage}{\textwidth}
        \centering
        \begin{minipage}[t]{0.18\textwidth}
            \centering
            \fcolorbox{blue}{white}{\includegraphics[width=\linewidth]{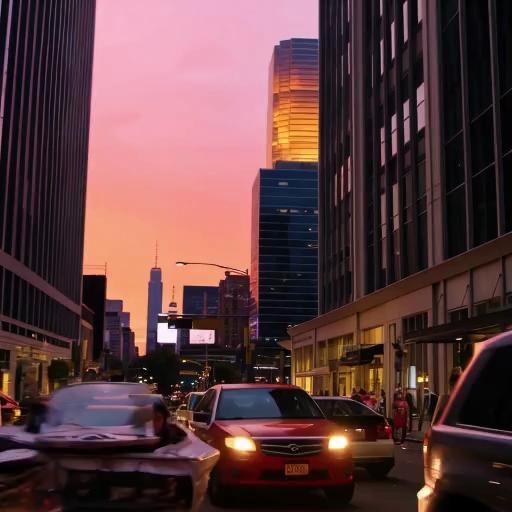}}
            \\[0.5em]
            \textbf{Reference Image}
        \end{minipage}
        \hfill
        \includegraphics[width=0.18\textwidth]{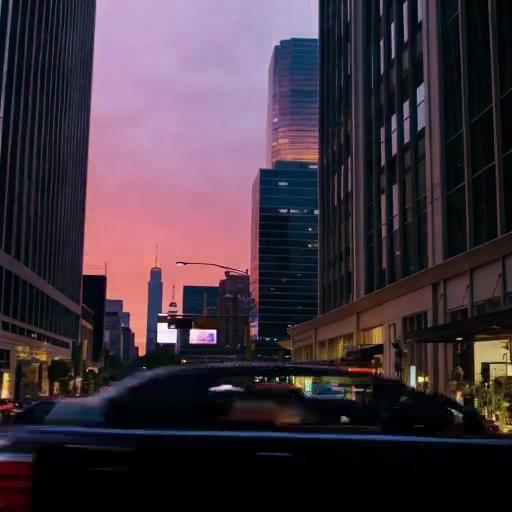}
        \includegraphics[width=0.18\textwidth]{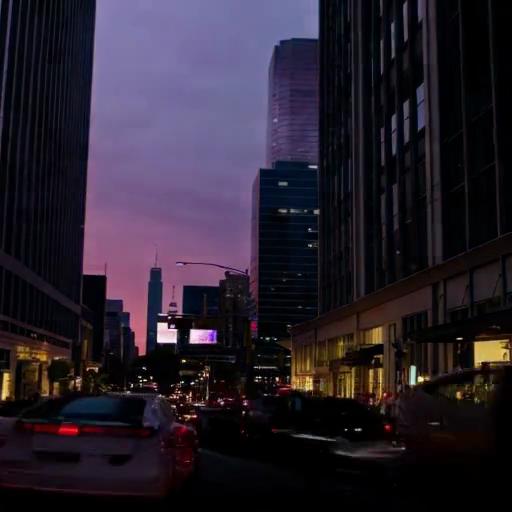}
        \includegraphics[width=0.18\textwidth]{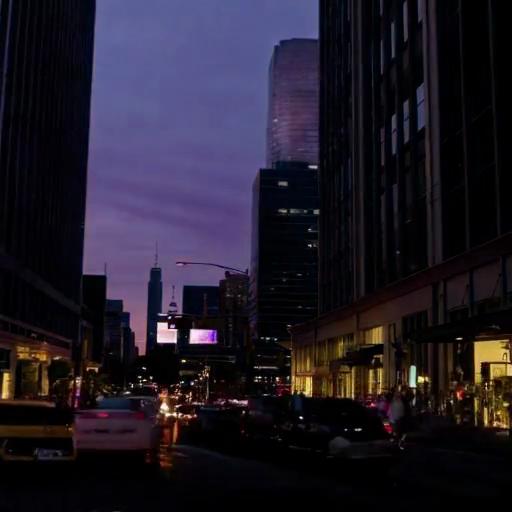}
        \includegraphics[width=0.18\textwidth]{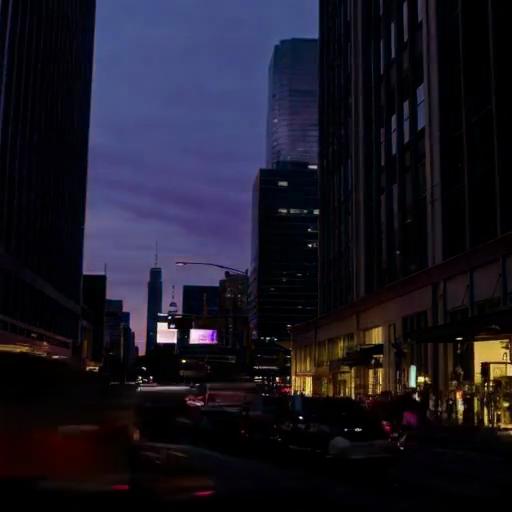}
    \end{minipage}
    \\[1em]
    \centering
    \centering
    \textbf{Prompt:} \textit{Robots move efficiently through a futuristic laboratory, adjusting holographic displays and conducting experiments, while scientists observe and interact with the high-tech equipment.} \\[0.5em]
    \begin{minipage}{\textwidth}
        \centering
        \begin{minipage}[t]{0.18\textwidth}
            \centering
            \fcolorbox{blue}{white}{\includegraphics[width=\linewidth]{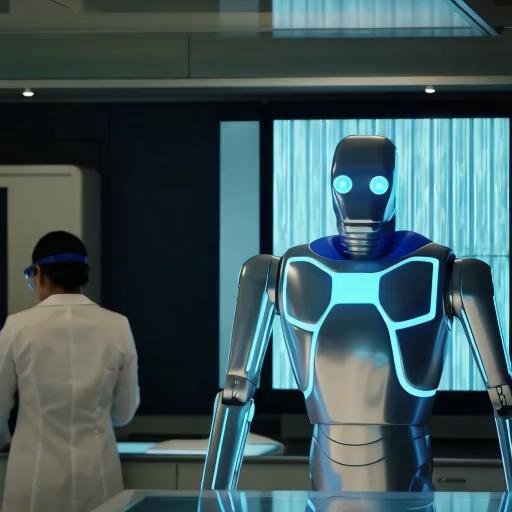}}
            \\[0.5em]
            \textbf{Reference Image}
        \end{minipage}
        \hfill
        \includegraphics[width=0.18\textwidth]{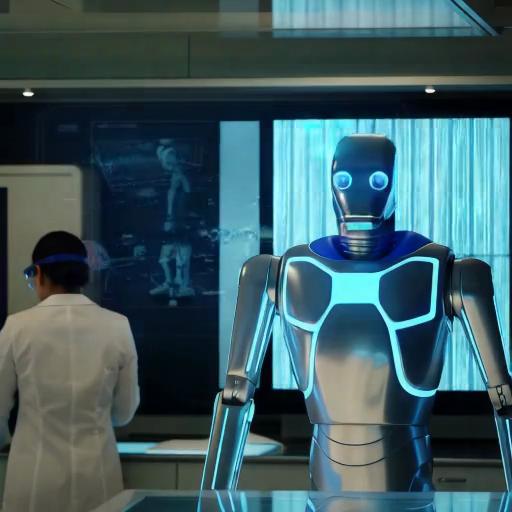}
        \includegraphics[width=0.18\textwidth]{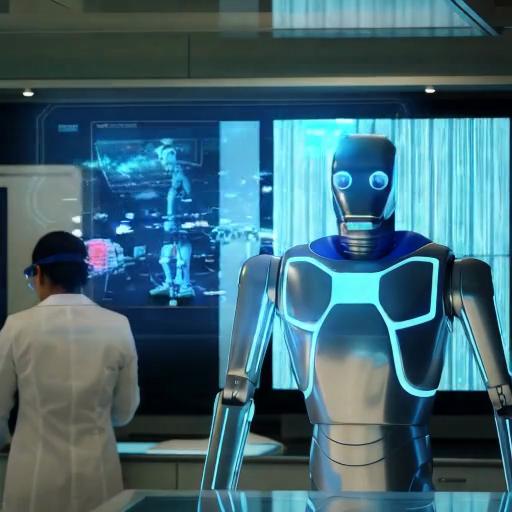}
        \includegraphics[width=0.18\textwidth]{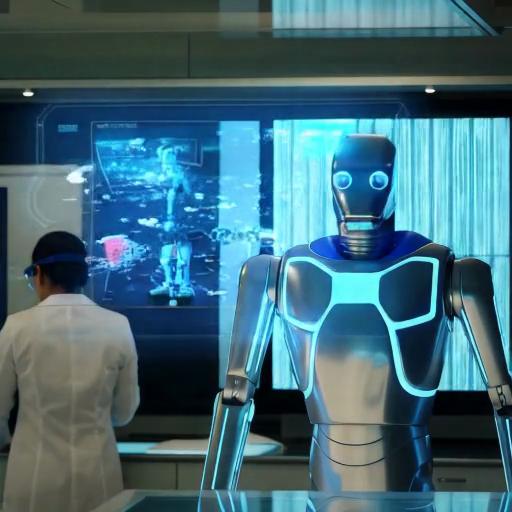}
        \includegraphics[width=0.18\textwidth]{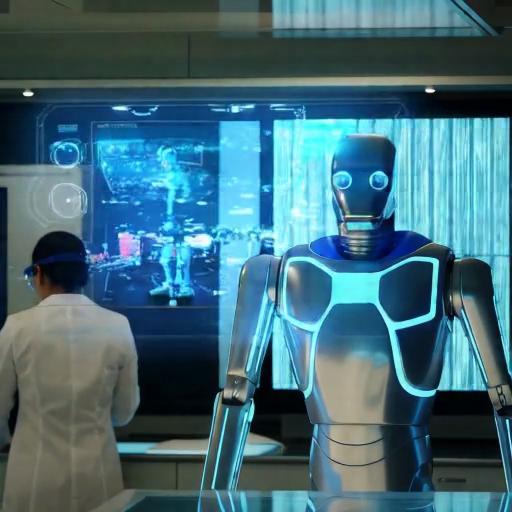}

    \end{minipage}
    \\[1em]

    \centering
    \begin{minipage}{\textwidth}
        \centering
        \begin{minipage}[t]{0.18\textwidth}
            \centering
            \fcolorbox{blue}{white}{\includegraphics[width=\linewidth]{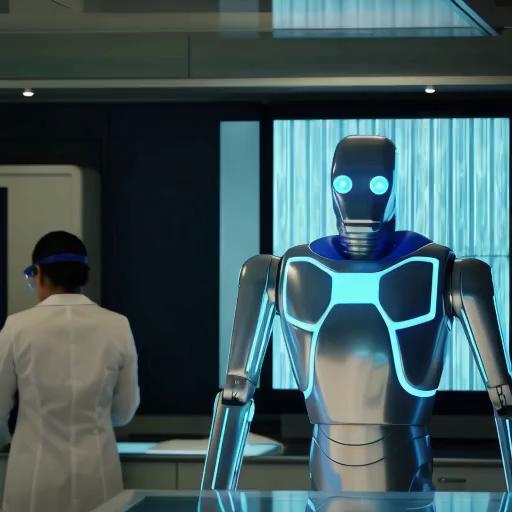}}
            \\[0.5em]
            \textbf{Reference Image}
        \end{minipage}
        \hfill
        \includegraphics[width=0.18\textwidth]{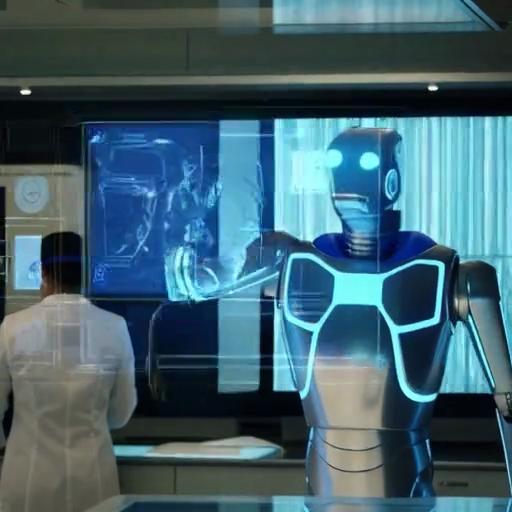}
        \includegraphics[width=0.18\textwidth]{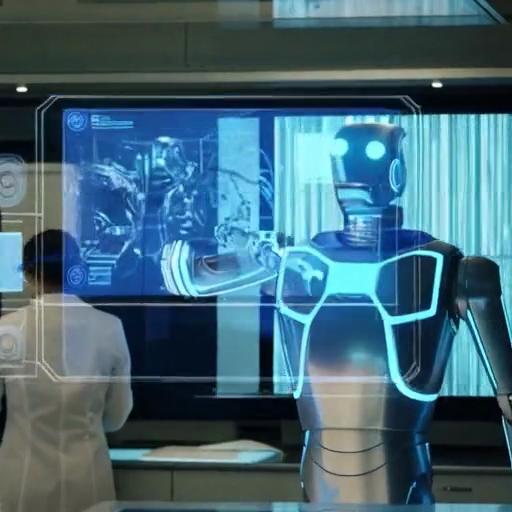}
        \includegraphics[width=0.18\textwidth]{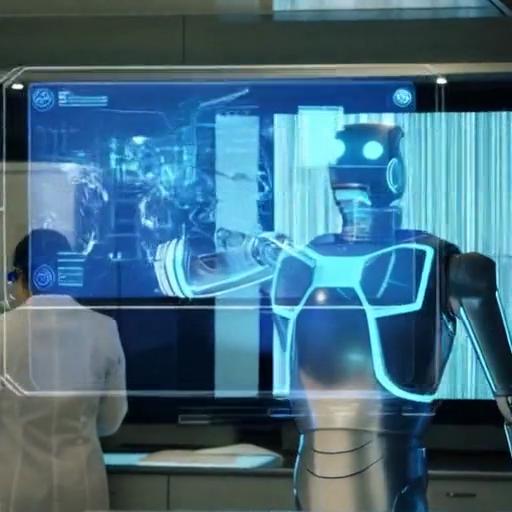}
        \includegraphics[width=0.18\textwidth]{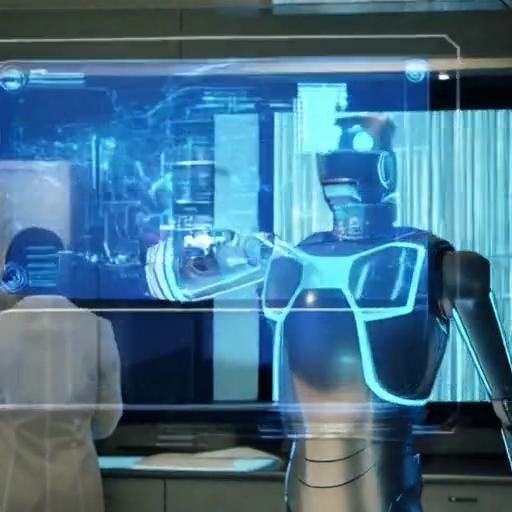}
    \end{minipage}
    \\[1em]
    
\caption{Visualization of STIV-M-512 V.S. STIV-M-512-JIT. (Given the same prompt, the figures in the top row are generated by STIV-M-512, while the figures in the bottom row are generated by STIV-M-512-JIT.)}
\label{fig:dpnodp3}
\end{figure*}

As mentioned in Section~\ref{sec:greater_motion}, after adding imaging dropout. We observe this phenomenon happens when we scale our model to 8B with >= 512 resolutions, probably due to the model being more easily overfitting to follow the first frame with a larger model capacity, and it becomes worse under the higher resolution. Specifically, we showcase some examples to see the different between generated videos without image dropout and videos with image dropout (STIV-M-512). We generate the videos conditioned on the first frame and text prompt borrowed from MovieGenBench~\citep{polyak2024movie}
As shown in Fig.~\ref{fig:dpnodp1} to ~\ref{fig:dpnodp3}, using image condition dropout in general achieves better performance than the baseline in terms of motion quality.

\clearpage

\section{Detailed Results for T2V and STIV}
\subsection{Details of VBench and VBench-I2V Evaluation Metrics}
\label{sec:detailed_vbench}
We follow the same as the evaluation protocol provided by VBench~\cite {huang2024vbench}.
\subsubsection{Video Quality}
\textit{Video Quality} is divided into two aspects: \textit{Temporal Quality} and \textit{Image Quality}. Temporal Quality evaluates cross-frame consistency, including (1) \textit{Subject Consistency}, ensuring that subjects maintain a consistent appearance across frames; (2) \textit{Background Consistency}, assessing stability in the background using feature similarity; (3) \textit{Temporal Flickering}, measuring smooth transitions in both static and dynamic areas; (4) \textit{Motion Smoothness}, evaluating the fluidity and realism of motion; and (5) \textit{Dynamic Degree}, analyzing the presence of large-scale dynamics or motions. Image Quality focuses on individual images and evaluates (1) \textit{Aesthetic Quality}, considering artistic appeal and visual richness, and (2) \textit{Imaging Quality}, measuring clarity, noise, and other distortions.

\subsubsection{Video-Condition Consistency}
\textit{Video-Condition Consistency} ensures alignment with the input prompt and is categorized into \textit{Semantics} and \textit{Style}, each with finer-grained dimensions.
\vspace{0.2cm}

\noindent\textbf{Semantics}
(1) \textit{Object Class}: Measures the success of generating specific objects described in the text prompt.  
(2) \textit{Multiple Objects}: Evaluates the ability to compose multiple objects from different classes in a single frame.  
(3) \textit{Human Action}: Assesses whether the generated video accurately captures actions described in the prompt.  
(4) \textit{Color}: Ensures synthesized object colors align with the text description.  
(5) \textit{Spatial Relationship}: Checks whether spatial relationships between objects align with the prompt.  
(6) \textit{Scene}: Evaluates consistency between generated scenes and the intended description (e.g., “ocean” versus “river”).
\vspace{0.2cm}

\noindent\textbf{Style}
(1) \textit{Appearance Style}: Measures consistency of styles mentioned in the prompt, such as “oil painting” or “cyberpunk”
(2) \textit{Temporal Style}: Assesses temporal continuity of styles across frames, ensuring smooth transitions.
\vspace{0.1cm}

\noindent\textbf{Overall Consistency}
We further evaluate \textit{Overall Consistency} using metrics that combine semantic and style alignment, reflecting both the accuracy and coherence of generated videos.

VBench-I2V builds upon the VBench with three new Video-Image Alignment metrics: Subject Consistency, Background Consistency, and Camera Motion Control. These additional metrics provide a more comprehensive evaluation by focusing on how well the generated video aligns with the input image and specified prompt instructions.
Specifically, Subject Consistency evaluates the alignment between the subject in the input image and the generated video, ensuring coherence in character or object representation. Background Consistency assesses the continuity of the background scene between the input image and the video, highlighting the model's ability to maintain a consistent environment. Camera Motion Control, under Video-Text Alignment, examines the adherence to camera control directions as described in the prompt, which is crucial for generating realistic video sequences that respond to specified dynamic instructions.

\subsection{Detailed Results on VBench and VBench-I2V}
We showcase the detailed version of the performance shown in Tab.~\ref{tab:t2v_detailed} and Tab.~\ref{tab:ti2v_metrics}.

\begin{table*}[!htp]\centering
\scriptsize
\begin{tabular}{l|c|c|c|c|c|c|c|c|c|c}
\Xhline{1pt}
\multirow{2}{*}{\textbf{Model}} & \textbf{Subject} & \textbf{Back.} & \textbf{Temporal} & \textbf{Motion} & \textbf{Dynamic} & \textbf{Aesthetic} & \textbf{Imaging} & \textbf{Object} & \textbf{Multiple} & \textbf{Human} \\
& \textbf{Cons.} & \textbf{Cons.} & \textbf{Flickering} & \textbf{Smooth.} & \textbf{Degree} & \textbf{Quality} & \textbf{Quality} & \textbf{Class} & \textbf{Objects} & \textbf{Action} \\
\Xhline{1pt}
CogVideoX-5B~\citep{yang2024cog-video-x} & 96.2 & 96.5 & 98.7 & 96.9 & 80.0 & 62.0 & 62.9 & 85.2 & 62.1 & 99.4 \\
CogVideoX-2B~\citep{yang2024cog-video-x} & 96.8 & 96.6 & 98.9 & 97.7 & 59.9 & 60.8 & 61.7 & 83.4 & 62.6 & 98.0 \\
Allegro~\citep{zhou2024allegro} & 96.3 & 96.7 & 99.0 & 98.8 & 55.0 & 63.7 & 63.6 & 87.5 & 59.9 & 91.4 \\
AnimateDiff-V2~\citep{guo2023animatediff} & 95.3 & 97.7 & 98.8 & 97.8 & 40.8 & 67.2 & 70.1 & 90.9 & 36.9 & 92.6 \\
OpenSora V1.2~\citep{zheng2024open-sora} & 96.8 & 97.6 & \textbf{99.5} & 98.5 & 42.4 & 56.9 & 63.3 & 82.2 & 51.8 & 91.2 \\
T2V-Turbo~\citep{li2024t2v-turbo} & 96.3 & 97.0 & 97.5 & 97.3 & 49.2 & 63.0 & \textbf{72.5} & 94.0 & 54.7 & 95.2 \\
VideoCrafter-2.0~\citep{chen2024videocrafter2} & 96.9 & 98.2 & 98.4 & 97.7 & 42.5 & 63.1 & 67.2 & 92.6 & 40.7 & 95.0 \\
LaVie-2~\citep{wang2023lavie} & 97.9 & 98.5 & 98.8 & 98.4 & 31.1 & \textbf{67.6} & 70.4 & \textbf{97.5} & 64.9 & 96.4 \\
LaVIE~\citep{wang2023lavie} & 91.4 & 97.5 & 98.3 & 96.4 & 49.7 & 54.9 & 61.9 & 91.8 & 33.3 & 96.8 \\
ModelScope~\citep{wang2023modelscope} & 89.9 & 95.3 & 98.3 & 95.8 & 66.4 & 52.1 & 58.6 & 82.2 & 39.0 & 92.4 \\
VideoCrafter~\citep{chen2023videocrafter1} & 86.2 & 92.9 & 97.6 & 91.8 & \textbf{89.7} & 44.4 & 57.2 & 87.3 & 25.9 & 93.0 \\
CogVideo~\citep{hong2022cogvideo} & 92.2 & 95.4 & 97.6 & 96.5 & 42.2 & 38.2 & 41.0 & 73.4 & 18.1 & 78.2 \\
\hline
PIKA~\citep{pika2023pika} & 96.9 & 97.4 & \textbf{99.7} & \textbf{99.5} & 47.5 & 62.4 & 61.9 & 88.7 & 43.1 & 86.2 \\
Gen-3~\citep{runway2024gen} & 97.1 & 96.6 & 98.6 & 99.2 & 60.1 & 63.3 & 66.8 & 87.8 & 53.6 & 96.4 \\
Gen-2~\citep{runway2023gen} & 97.6 & 97.6 & 99.6 & \textbf{99.6} & 18.9 & 67.0 & 67.4 & 90.9 & 55.5 & 89.2 \\
KLING~\citep{kuaishou2024kling} & \textbf{98.3} & 97.6 & 99.3 & 99.4 & 46.9 & 61.2 & 65.6 & 87.2 & 68.1 & 93.4 \\
EMU3~\citep{girdhar2023emu} & 95.3 & 97.7 & 98.6 & 98.9 & 79.3 & 59.6 & 62.6 & 86.2 & 44.6 & 77.7 \\
\hline
XL & 96.0 & 98.5 & 98.4 & 96.5 & 62.5 & 56.3 & 59.3 & 91.5 & 41.3 & 98.0 \\
XXL & 97.5 & \textbf{98.9} & 99.1 & 98.2 & 48.6 & 56.2 & 59.7 & 91.1 & 49.1 & \textbf{99.0} \\
M-256 & 96.0 & 98.5 & 98.6 & 97.2 & 68.1 & 57.0 & 60.8 & 88.8 & 62.1 & 98.0 \\
M-512 & 95.9 & 96.9 & 98.8 & 98.0 & 59.7 & 60.6 & 62.5 & 85.9 & \textbf{72.4} & 96.0 \\
M-512-SFT & 96.7 & 97.4 & 98.7 & 98.3 & 70.8 & 61.7 & 63.9 & 88.1 & 67.7 & 97.0 \\
M-512-SFT+TUP & 94.8 & 95.9 & 98.7 & 99.2 & 70.8 & 63.7 & 65.0 & 88.9 & 70.3 & 95.0 \\
M-512-UnMSFT & 94.3 & 96.9 & 98.8 & 96.7 & 77.8 & 61.4 & 68.6 & 90.0 & 72.3 & 97.0 \\
M-512-UnMSFT+TUP & 95.2 & 95.8 & 98.8 & 99.2 & 70.8 & 63.6 & 65.9 & 90.0 & 69.8 & 94.0 \\
\hline
\hline
\multirow{2}{*}{\textbf{Model}} & \textbf{Color} & \textbf{Spatial} & \textbf{Scene} & \textbf{App.} & \textbf{Temp.} & \textbf{Overall} & \textbf{Quality} & \textbf{Semantic} & \textbf{Total} & \textbf{Averaged} \\
& & \textbf{Rel.} & & \textbf{Style} & \textbf{Style} & \textbf{Cons.} & \textbf{Score} & \textbf{Score} & \textbf{Score} & \textbf{Scores} \\
\Xhline{1pt}
CogVideoX-5B~\citep{yang2024cog-video-x} & 82.8 & 66.4 & 53.2 & 24.9 & 25.4 & 27.6 & 82.8 & 77.0 & 81.6 & 70.0 \\
CogVideoX-2B~\citep{yang2024cog-video-x} & 79.4 & 69.9 & 51.1 & 24.8 & 24.4 & 26.7 & 82.2 & 75.8 & 80.9 & 68.3 \\
Allegro~\citep{zhou2024allegro} & 82.8 & 67.2 & 46.7 & 20.5 & 24.4 & 26.4 & 83.1 & 73.0 & 81.1 & 67.5 \\
AnimateDiff-V2~\citep{guo2023animatediff} & 87.5 & 34.6 & 50.2 & 22.4 & 26.0 & 27.0 & 82.9 & 69.8 & 80.3 & 64.7 \\
OpenSora V1.2~\citep{zheng2024open-sora} & 90.1 & 68.6 & 42.4 & 24.0 & 24.5 & 26.9 & 81.4 & 73.4 & 79.8 & 66.0 \\
T2V-Turbo~\citep{li2024t2v-turbo} & 89.9 & 38.7 & 55.6 & 24.4 & 25.5 & 28.2 & 82.6 & 74.8 & 81.0 & 67.4 \\
VideoCrafter-2.0~\citep{chen2024videocrafter2} & 92.9 & 35.9 & 55.3 & \textbf{25.1} & 25.8 & 28.2 & 82.2 & 73.4 & 80.4 & 66.0 \\
LaVie-2~\citep{wang2023lavie} & 91.7 & 38.7 & 49.6 & \textbf{25.1} & 25.2 & 27.4 & 83.2 & 75.8 & 81.8 & 67.6 \\
LaVIE~\citep{wang2023lavie} & 86.4 & 34.1 & 52.7 & 23.6 & 25.9 & 26.4 & 78.8 & 70.3 & 77.1 & 63.8 \\
ModelScope~\citep{wang2023modelscope} & 81.7 & 33.7 & 39.3 & 23.4 & 25.4 & 25.7 & 78.1 & 66.5 & 75.8 & 62.4 \\
VideoCrafter~\citep{chen2023videocrafter1} & 78.8 & 36.7 & 43.4 & 21.6 & 25.4 & 25.2 & 81.6 & 72.2 & 79.7 & 62.3 \\
CogVideo~\citep{hong2022cogvideo} & 79.6 & 18.2 & 28.2 & 22.0 & 7.8 & 7.7 & 72.1 & 46.8 & 67.0 & 52.3 \\
\hline
PIKA~\citep{pika2023pika} & 90.6 & 61.0 & 49.8 & 22.3 & 24.2 & 25.9 & 82.9 & 71.8 & 80.7 & 66.1 \\
Gen-3~\citep{runway2024gen} & 80.9 & 65.1 & 54.6 & 24.3 & 24.7 & 26.7 & 84.1 & 75.2 & 82.3 & 68.5 \\
Gen-2~\citep{runway2023gen} & 89.5 & 66.9 & 48.9 & 19.3 & 24.1 & 26.2 & 82.5 & 73.0 & 80.6 & 66.1 \\
KLING~\citep{kuaishou2024kling} & 89.9 & \textbf{73.0} & 50.9 & 19.6 & 24.2 & 26.4 & 83.4 & 75.7 & 81.9 & 68.8 \\
EMU3~\citep{girdhar2023emu} & 88.3 & 68.7 & 37.1 & 20.9 & 23.3 & 24.8 & 84.1 & 68.4 & 81.0 & 66.7 \\
\hline
XL & 86.4 & 42.4 & 54.4 & 22.4 & 26.3 & 27.8 & 80.7 & 72.5 & 79.1 & 66.1 \\
XXL & 90.8 & 45.1 & 45.5 & 22.1 & 26.1 & 27.4 & 81.2 & 72.7 & 79.5 & 65.9 \\
M-256 & 83.6 & 44.5 & 54.7 & 22.5 & 26.6 & 28.4 & 82.7 & 74.8 & 80.6 & 67.9 \\
M-512 & 91.2 & 51.0 & 53.6 & 23.9 & 25.8 & 27.8 & 82.2 & 77.0 & 81.2 & 68.8 \\
M-512-SFT & 93.7 & 58.0 & 52.8 & 24.6 & 26.2 & 28.5 & 83.9 & 78.3 & 82.8 & 70.3 \\
M-512-SFT+TUP & \textbf{94.7} & 50.6 & \textbf{57.3} & 24.5 & \textbf{26.7} & 28.6 & 84.2 & 78.5 & \textbf{83.1} & 70.3 \\
M-512-UnMSFT & 92.0 & 59.8 & 53.1 & 24.8 & \textbf{26.7} & \textbf{28.8} & 83.7 & \textbf{79.5} & 82.9 & \textbf{71.2} \\
M-512-UnMSFT+TUP & 87.7 & 46.9 & 57.1 & 24.5 & 26.6 & 28.5 & \textbf{84.4} & 77.2 & 83.0 & 69.7 \\
\Xhline{1pt}
\end{tabular}
\caption{Detailed Evaluation Results for Text-To-Video Generation Models.}
\label{tab:t2v_detailed}
\end{table*}

\clearpage

\begin{table*}[!htp]\centering
\small
\begin{tabular}{l|c|c|c|c|c|c}
\Xhline{1pt}
\multirow{2}{*}{\textbf{Model}} & \textbf{Subject} & \textbf{Background} & \textbf{Temporal} & \textbf{Motion} & \textbf{Dynamic} & \textbf{Aesthetic} \\
& \textbf{Consistency} & \textbf{Consistency} & \textbf{Flickering} & \textbf{Smoothness} & \textbf{Degree} & \textbf{Quality} \\
\Xhline{1pt}
DynamicCrafter-256~\citep{xing2023dynamicrafter} &94.7 &98.3 &98.1 &97.8 &40.6 &58.7 \\
DynamicCrafter-512~\citep{xing2023dynamicrafter} &93.8 &96.6 &95.6 &96.8 &69.7 &60.9 \\
Animate-Anything~\citep{dai2023animateanything} &98.9 &98.2 &98.1 &98.6 &2.7 &67.1 \\
SVD~\citep{blattmann2023stable} &95.5 &96.6 &98.1 &98.1 &52.4 &60.2 \\
SEINE-512~\citep{chen2023seine} &95.3 &97.1 &97.3 &97.1 &27.1 &64.6 \\
VideoCrafter-I2V~\citep{chen2024videocrafter2} &97.9 &98.8 &98.2 &98.0 &22.6 &60.8 \\
Consistent-I2V~\citep{ren2024consisti2v} &95.3 &98.3 &97.6 &97.4 &18.6 &59.0 \\
I2VGen-XL~\citep{zhang2023i2vgen} &94.2 &97.1 &98.3 &26.1 &26.1 &64.8 \\
\hline
STIV-M &95.4 &98.9 &97.2 &98.1 &32.1 &59.0 \\
STIV-M-512 &99.5 &99.3 &99.5 &99.6 &10.2 &62.5 \\
STIV-M-512-JIT &98.1 &98.6 &98.7 &99.1 &24.0 &65.4 \\
\hline
\hline
\multirow{2}{*}{\textbf{Model}} & \textbf{Imaging} & \textbf{I2V} & \textbf{I2V} & \textbf{Camera} & \textbf{I2V} & \textbf{Final} \\
& \textbf{Quality} & \textbf{Subject} & \textbf{Background} & \textbf{Motion} & \textbf{Quality} & \textbf{Score} \\
\Xhline{1pt}
DynamicCrafter-256~\citep{xing2023dynamicrafter} &62.3 &97.1 &97.6 &20.9 &80.2 &88.4 \\
DynamicCrafter-512~\citep{xing2023dynamicrafter} &68.6 &97.2 &97.4 &32.0 &81.6 &89.1 \\
Animate-Anything~\citep{dai2023animateanything} &72.1 &98.8 &98.6 &13.1 &81.2 &89.8 \\
SVD~\citep{blattmann2023stable} &69.8 &98.8 &98.6 &62.3 &82.8 &89.9 \\
SEINE-512~\citep{chen2023seine} &71.4 &97.2 &96.9 &21.0 &80.6 &88.4 \\
VideoCrafter-I2V~\citep{chen2024videocrafter2} &71.7 &91.2 &91.3 &33.6 &81.3 &85.1 \\
Consistent-I2V~\citep{ren2024consisti2v} &66.9 &95.8 &96.0 &33.9 &78.9 &86.8 \\
I2VGen-XL~\citep{zhang2023i2vgen} &69.1 &96.5 &96.8 &18.5 &81.2 &88.5 \\
\hline
STIV-M &66.1 &97.0 &97.4 &22.7 &78.8 &87.6 \\
STIV-M-512 &71.5 &99.2 &97.3 &13.2 &82.1 &90.1 \\
STIV-M-512-JIT &71.0 &98.8 &97.5 &15.1 &81.9 &89.8 \\
\Xhline{1pt}
\end{tabular}
\caption{Detailed Evaluation Results for Text-Image-To-Video Generation Models.}
\label{tab:ti2v_metrics}
\end{table*}

\section{Details of Model Initialization Ablations}
\label{sec:appendix_model_initialisation}

To facilitate a fair comparison for different initialization methods we estimate the FLOPs associated with spatial-temporal computation in the transformer for various model training steps (Tables ~\ref{tab:t2v_high_res_init_flops} and ~\ref{tab:t2v_high_frame_count_init_flops}). When controlling for FLOPs we take into account, the compute used to pretrain the intermediate models, the reduction in an effective number of tokens due to masking in the relevant attention blocks, the increased parameter count when temporal attention is included, and the increased number of tokens passed to the model during high resolution training. For both the high resolution and higher frame count experiments we attebyto keep the compute budget across model initialization ablations similar. Tables ~\ref{tab:t2v_highres_init_vbench_quality_breakdown} and ~\ref{tab:t2v_high_frame_count_init_vbench_quality_breakdown} show the VBench quality metrics for high resolution and high frame count XL sized models respectively.

\begin{table*}[h!]
    \centering
    \begin{tabular}{r|l|c|c|c|c|c}
        \Xhline{1pt}
        Init. Method & Models & Stage 1 & Stage 2 & Stage 3 & Stage 4 & Total \\
        \Xhline{1pt}
        Scratch             & T2V-512                             & 5.93 &                         &                         &                         & 5.93 \\
        T2V-256             & T2I-256, T2V-256, T2V-512           & 1.11 & 2.05 & 2.84 &                         & 6.00 \\
        T2I-512             & T2I-256, T2I-512, T2V-512           & 1.11 & 8.43 & 4.02 &                         & 5.97 \\
        Both                & T2I-256, T2V-256, T2I-512, T2V-512  & 1.11 & 2.05 & 8.43 & 1.98 & 5.98 \\
        \Xhline{1pt}
    \end{tabular}
    \caption{A breakdown of FLOPs for training high resolution T2V models. Unit $10^{21}$.}
    \label{tab:t2v_high_res_init_flops}
\end{table*}
\begin{table*}[h!]
    \centering
    \begin{tabular}{l|l|c|c|c|c}
        \Xhline{1pt}
        Init. Method & Models & Stage 1 & Stage 2 & Stage 3 & Total \\
        \Xhline{1pt}
        T2I                        & T2I-256, T2V-256-40                           & 1.11 & 2.05 &                         & 3.16 \\
        T2V (int.)           & T2I-256, T2V-256-20, T2V-256-40            & 1.11 & 1.02 & 1.02 & 3.16 \\
        T2V (ext.)           & T2I-256, T2V-256-20, T2V-256-40            & 1.11 & 1.02 & 1.02 & 3.16 \\
        T2V 2x (int.) & T2I-256, T2V-256-20 2x stride, T2V-256-40 & 1.11 & 1.02 & 1.02 & 3.16 \\
        \Xhline{1pt}
    \end{tabular}
    \caption{A breakdown of FLOPs for training high frame count T2V models. Unit: $10^{21}$.}
    \label{tab:t2v_high_frame_count_init_flops}
\end{table*}
\begin{table*}[h]
\small
\centering
\begin{tabular}{r|c|c|c|c|c|c|c|c}
\Xhline{1pt}

\textbf{Initial} & \textbf{Subject} & \textbf{Background} & \textbf{Temporal} & \textbf{Motion} & \textbf{Dynamic} & \textbf{Aesthetic} & \textbf{Imaging} & \textbf{Object} \\
\textbf{Method} & \textbf{Cons.} & \textbf{Cons.} & \textbf{Flickering} & \textbf{Smoothness} & \textbf{Degree} & \textbf{Quality} & \textbf{Quality} & \textbf{Class} \\

\Xhline{1pt}
Scratch & \textbf{93.1} & 97.1 & 97.9 & 97.3 & \textbf{61.4} & 58.6 & 58.6 & 87.0 \\
T2V-256 & 91.9 & 97.1 & 98.0 & \textbf{97.5} & 58.6 & 59.4 & 59.7 & \textbf{91.2} \\
T2I-512 & 92.3 & 97.2 & 98.2 & 97.0 & 52.2 & 60.0 & 59.3 & 88.8 \\
Both    & 92.4 & \textbf{97.3} & \textbf{98.3} & 97.4 & 53.9 & \textbf{60.7} & \textbf{60.6} & 88.2 \\
\hline
\hline

\textbf{Initial} & \textbf{Multiple} & \textbf{Human} & \textbf{Color} & \textbf{Spatial} & \textbf{Scene} & \textbf{App.} & \textbf{Temp.} & \textbf{Overall} \\
\textbf{Method} & \textbf{Objects} & \textbf{Action} & & \textbf{Relationship} & & \textbf{Style} & \textbf{Style} & \textbf{Cons.} \\

\Xhline{1pt}
Scratch & 29.7 & 95.4 & 88.3 & 33.8 & 46.9 & 21.6 & 25.8 & 26.4 \\
T2V-256 & 45.7 & 95.8 & \textbf{89.0} & 36.3 & 50.0 & 21.9 & 25.8 & 27.3 \\
T2I-512 & 47.4 & \textbf{96.4} & 87.9 & \textbf{37.0} & 49.1 & 22.5 & 26.2 & 27.8 \\
Both    & \textbf{49.7} & 96.0 & 88.1 & 36.7 & \textbf{52.3} & \textbf{22.8} & \textbf{26.3} & \textbf{28.0} \\
\bottomrule
\end{tabular}
\caption{Detailed VBench metrics of different model initialization methods for higher resolution T2V model training.}
\label{tab:t2v_highres_init_vbench_quality_breakdown}
\end{table*}

\begin{table*}[h]
\small
\centering
\begin{tabular}{r|c|c|c|c|c|c|c|c}
\Xhline{1pt}
\textbf{Initial} & \textbf{Subject} & \textbf{Background} & \textbf{Temporal} & \textbf{Motion} & \textbf{Dynamic} & \textbf{Aesthetic} & \textbf{Imaging} & \textbf{Object} \\
\textbf{Method} & \textbf{Cons.} & \textbf{Cons.} & \textbf{Flickering} & \textbf{Smoothness} & \textbf{Degree} & \textbf{Quality} & \textbf{Quality} & \textbf{Class} \\
\Xhline{1pt}
T2I                         & \textbf{93.2} & \textbf{98.1} & \textbf{98.7} & 95.2 & 57.8 & 54.2 & 58.2 & 84.6 \\
T2V (int.)            & 91.7 & 97.7 & 97.7 & 96.8 & 64.7 & \textbf{54.7} & 59.2 & \textbf{86.9} \\
T2V (ext.)            & 91.3 & 97.5 & 97.8 & 96.9 & 58.6 & 54.6 & \textbf{60.0} & 86.1 \\
T2V 2x (int.)  & 91.0 & 97.3 & 97.2 & \textbf{97.0} & \textbf{70.3} & 54.1 & 59.4 & 85.8 \\
\hline
\hline
\textbf{Initial} & \textbf{Multiple} & \textbf{Human} & \textbf{Color} & \textbf{Spatial} & \textbf{Scene} & \textbf{App.} & \textbf{Temp.} & \textbf{Overall} \\
\textbf{Method} & \textbf{Objects} & \textbf{Action} & & \textbf{Relationship} & & \textbf{Style} & \textbf{Style} & \textbf{Cons.} \\
\Xhline{1pt}
T2I                         & \textbf{30.8} & 92.2 & 85.0 & \textbf{29.9} & 45.2 & 21.1 & 25.0 & 26.0 \\
T2V (int.)            & 25.5 & \textbf{95.4} & 85.3 & 28.6 & 41.4 & \textbf{21.2} & 25.3 & 26.6 \\
T2V (ext.)            & 28.5 & 95.2 & 84.2 & 25.9 & 36.8 & 20.9 & 25.6 & \textbf{26.8} \\
T2V 2x (int.)  & 29.3 & 94.0 & \textbf{87.7} & 28.6 & \textbf{44.2} & 20.9 & \textbf{25.7} & 26.7 \\
\bottomrule
\end{tabular}
\caption{Detailed VBench metrics of different model initialization methods for higher frame count T2V model training.}
\label{tab:t2v_high_frame_count_init_vbench_quality_breakdown}
\end{table*}


\section{Study of Class-to-Video on UCF-101}
\label{sec:ucf-101}

UCF-101 is an action recognition dataset, which contains 101 classes over 9.5K training videos. Here we train STIV from scratch and perform label-to-video (L2V) generation with 16 frames and 128$^\text{2}$ resolution. We follow TATS~\cite{ge2022long} to adopt the Inception Score (IS)~\cite{salimans2016is-score} and FVD for the evaluation\footnote{Following our baselines (\url{https://github.com/songweige/TATS/issues/13}), we apply C3D~\cite{tran2015c3d} pre-trained on UCF-101 for the IS logits. For FVD, we adopt I3D~\cite{carreira2017i3d} pre-trained on Kinetics-400~\cite{kay2017kinetics-400} to calculate the video embeddings.}. 

Tab.~\ref{table:ucf-101} shows that our L2V-XL achieves significant improvements, leading to +12\% IS and -22\% FVD over MAGVIT. This also highlights the effectiveness of our model design for convention video generation. From the ablation study over different modulations, only without spatial mask makes a lower FVD but degrades IS, while all other settings hurt the performance.

\begin{table*}[htbp]
\centering 
    \begin{tabular}{l|c|c|c}
        \Xhline{1pt}
        \multirow{2}{*}{Method} & \multirow{2}{*}{IS $\uparrow$} & \multirow{2}{*}{FVD $\downarrow$} & VBench- \\
        ~ & ~ & ~ & Quality $\uparrow$ \\
        \Xhline{1pt}
        CogVideo~\cite{hong2022cogvideo} & 50.5 & 626 & - \\
        TATS~\cite{ge2022long} & 79.3 & 332 & - \\
        MMVG~\cite{fu2023tvc} & 73.7 & 328 & - \\
        VideoFusion~\cite{luo2023video-fusion} & 80.0 & 173 & - \\
        MAGVIT~\cite{yu2023magvit} & 83.6 & 159 & - \\
        \Xhline{1pt}
        XL-128 & \underline{93.4} & 124 & 69.9 \\
        - Spatial Mask & 88.5 & \textbf{102} & \textbf{70.6} \\
        + Temporal Mask & \textbf{94.9} & 167 & 68.1 \\
        + Temporal ScaleShiftGate & 78.9 & 141 & 69.1 \\
        + Causal TemporalAttention & 86.9 & \underline{106} & \underline{70.3} \\
        \Xhline{1pt}
    \end{tabular}
    \caption{Performance of Class-to-Video Generation on UCF-101.}
    \label{table:ucf-101}
\end{table*}

\section{Details of Video Data Engine}
\label{sec:data_detailed}




\noindent\textbf{Details of Video Pre-processing and Feature Extraction}
To ensure high-quality input data, we first address the issue of inconsistent motions and unwanted transitions like cuts and fades in raw videos. Using PySceneDetect\footnote{\href{https://github.com/Breakthrough/PySceneDetect}{https://github.com/Breakthrough/PySceneDetect}}, we analyze video frames to identify and segment scenes with abrupt transitions or gradual fades. This process isolates and removes inconsistent segments, resulting in video clips that maintain visual consistency, reducing artifacts and improving the overall quality. After that, we extract several initial features for future filtering, including motion score, aesthetic score, text area, frame height, frame width, clarity score, temporal consistency, and video orientation, \textit{et al}.

\vspace{0.2cm}

\noindent\textbf{Details of Video Captioning and Categorization}
Video-text pairs play a crucial role in training text-to-video generation models. However, many video datasets lack well-aligned, high-quality captions and often include noisy or irrelevant content. Therefore, we’ve incorporated an additional video captioning module in our pipeline to generate comprehensive textual descriptions.

\begin{figure}[htbp]
\centering
\begin{minipage}{0.54\linewidth}
    We mainly explore two directions: (1) sample a few frames, apply an image captioner, and then use an LLM to summarize the resulting captions~\cite{blattmann2023stable}; (2) apply a video LLM to generate captions.
    We initially explored the first direction but found two major limitations. Firstly, the image captioner can only capture visual details in a single frame, resulting in a lack of descriptions of video motions.
    Secondly, the LLM may hallucinate when prompted to generate a dense caption based on multiple frame captions.
    Recent works~\cite{zhang2024llavahound, zhang2024video, chen2024sharegpt4video, zhang2024llavanext-video} use GPT-4V or GPT-4o to curate a fine-tuning dataset and train their video LLMs. To balance quality and cost in large-scale captioning, we select LLaVA-Hound-7B~\cite{zhang2024llavahound} as our video captioner.
    Then, we use an LLM to categorize the generated captions and obtain the distribution of videos.
\end{minipage}
\hfill
\begin{minipage}{0.42\linewidth}
    \centering
    \vspace{0.2cm}
    \includegraphics[width=\linewidth]{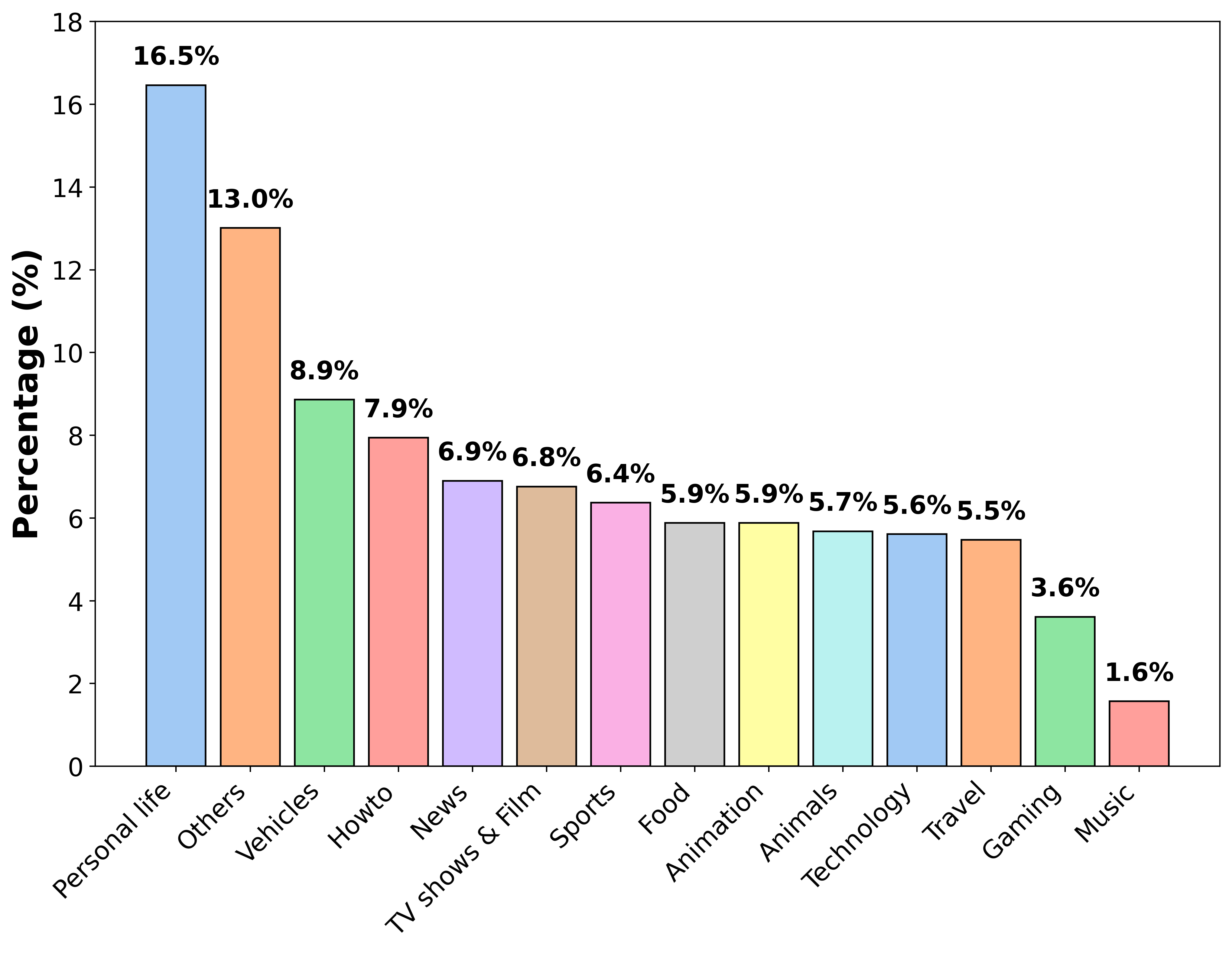}
    \vspace{-0.1cm}
    \caption{Category distribution of our curated Panda-30M dataset.}
    \label{fig:category_distribution}
\end{minipage}
\end{figure}

\noindent\textbf{Details of DSG-Video: Hallucination Evaluation}
To compare various captioning techniques, we develop an evaluation module that assesses both caption richness and accuracy.
We quantify caption richness by measuring the diversity of unique objects referenced across captions. And we identify hallucinations to assess accuracy.
Inspired by DSG~\cite{Cho2024DSG} for fine-grained text-to-image evaluation, we introduce \textbf{DSG-Video} to validate the presence of objects referenced in captions within the video content (Fig.~\ref{fig:dsg_video_overview}).
First, we use an LLM to automatically generate questions that probe key details of captions, such as object identity, actions, and context.
For instance, given a caption mentioning "a cat sitting on a couch", the LLM will generate questions like "Is there a cat in the video?" and "Is the cat on a couch?".
Second, we utilize a multimodal LLM to answer these object verification questions by evaluating the presence of each referenced object in $N$ uniformly sampled frames from the video.
For each question generated (e.g., "Is there a cat in this frame?"), the multimodal LLM examines each sampled frame and provides a response.
If, for a given question, all frame-based responses suggest that the entity in question is absent, we classify the object as hallucinated.
This approach ensures a thorough and frame-by-frame validation for each object in the video.
Based on this, we have two metrics: one that measures the fraction of hallucinated object instances (referred to as $\text{DSG-Video}_{i}$), and the other that calculates the fraction of sentences containing hallucinated objects (referred to as $\text{DSG-Video}_{s}$). We use these two metrics to assess caption quality.

$$\text{DSG-Video}_{i} = \frac{ \vert\{ \text{hallucinated objects}\}\vert }{\vert\{\text{all objects mentioned}\}\vert },$$
$$\text{DSG-Video}_{s} = \frac{ \vert\{ \text{sentences with hallucinated object}\}\vert }{\vert\{\text{  all sentences}\}\vert }.$$
\vspace{0.2cm}

\section{Details for Flexible Applications}
\label{sec:more_apps}
\subsection{Video Prediction}
We start from a STIV-XXL model to train a text-video-to-video model. Specifically, we replace the first four frames with ground truth frame latents to train additional 400K steps. The results are shown in the main text.

\subsection{Frame Interpolation}
We start from a STIV-XL model to train a text-video-to-video model. Specifically, we use a stride of two to sample the frames with ground truth frame latents to train additional 400K steps. The results are shown in the main text.

\subsection{Multi-view Generation}
Our model employs temporal attention to model cross-frame correspondence. However, this approach may be less effective than full-attention mechanisms, as temporal attention focuses solely on the same spatial position across all frames. To investigate this limitation, we conducted experiments on multiview generation, which involves predicting novel views of a specific object. For evaluation, we adopted the six-view scheme from Zero123++, consisting of elevation angles of \(30^\circ, -20^\circ, 30^\circ, -20^\circ, 30^\circ, -20^\circ\) and azimuth angles of \(30^\circ, 90^\circ, 150^\circ, 210^\circ, 270^\circ, 330^\circ\). This scheme was chosen because the large camera changes between adjacent views simulate a challenging scenario where the camera rotates around an object with significant motion, providing a robust test for the temporal attention mechanism's ability to capture large inter-frame motions.

For implementation, we redefined the input frames in our text-image-to-video model as the given views of the object, with an empty string as the text conditioning. The negative input views were set to zero latent vectors during classifier-free guidance training. For the generation, we used a guidance scale of 3.5. A visual comparison between our approach and Zero123++ is shown in Fig.~\ref{fig:multiview}.

\section{More Examples}
\label{sec:more_examples}
We show more examples at the end of the Appendix using the text prompts and image as first frame condition borrowed from MovieGenBench~\cite{polyak2024movie} and Sora~\cite{openai2024sora}.

\begin{figure*}[htbp]
    \centering
    \begin{minipage}{\textwidth}
        \centering
        \setlength{\fboxrule}{1.5pt}
        \fcolorbox{darkgreen}{lightcream}{
            \parbox{\textwidth}{
                \centering \textbf{\Large \textcolor{darkgreen}{Text-to-Video}}
            }
        } \\[1em]

    \textbf{Prompt:} \textit{A pirate ship sailing through a storm with enormous waves crashing against the sides, its crew fighting against the wind as lightning illuminates the scene.} \\[0.5em]
    \includegraphics[width=0.22\textwidth]{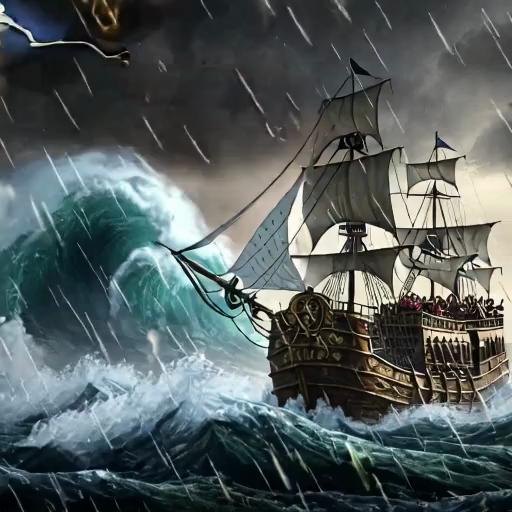}
    \includegraphics[width=0.22\textwidth]{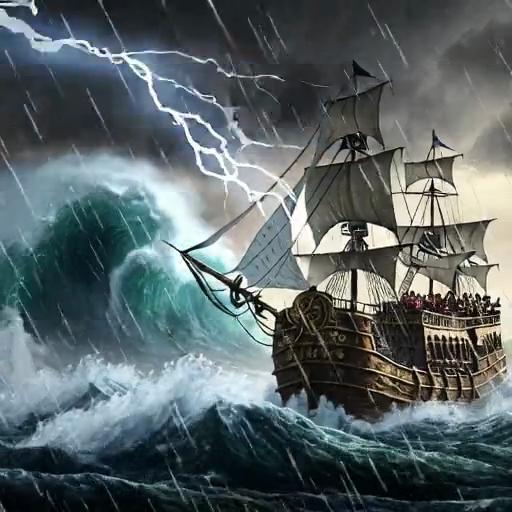}
    \includegraphics[width=0.22\textwidth]{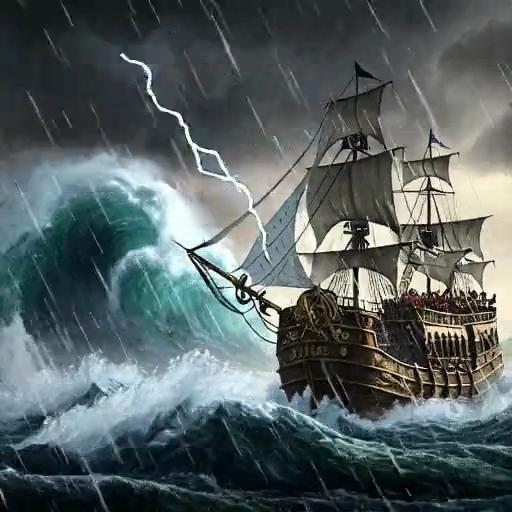}
    \includegraphics[width=0.22\textwidth]{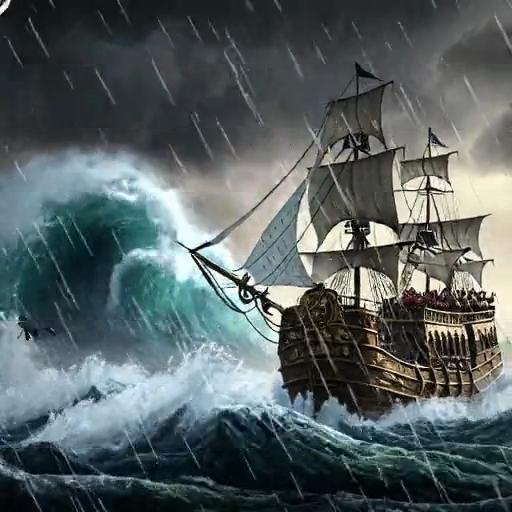}
    \\[1.5em]
    
    \textbf{Prompt:} \textit{A samurai on horseback charging across a field of cherry blossoms, slicing petals in mid-air as they fall, leaving a trail of pink in their path.} \\[0.5em]
    \includegraphics[width=0.22\textwidth]{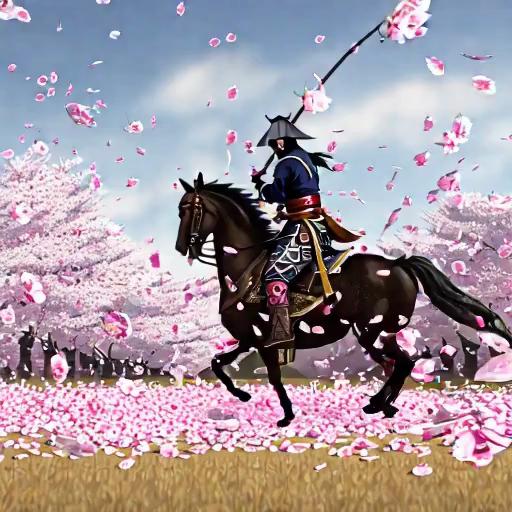}
    \includegraphics[width=0.22\textwidth]{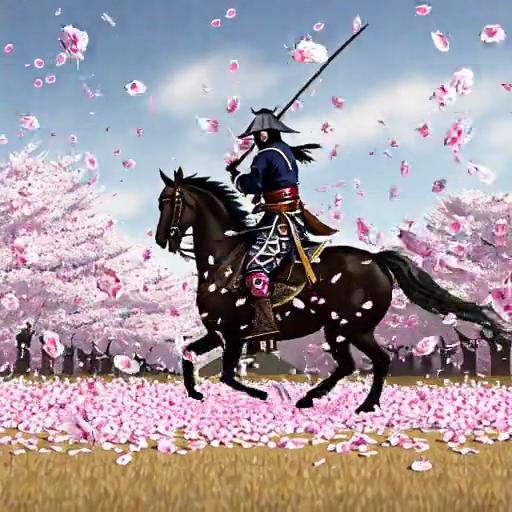}
    \includegraphics[width=0.22\textwidth]{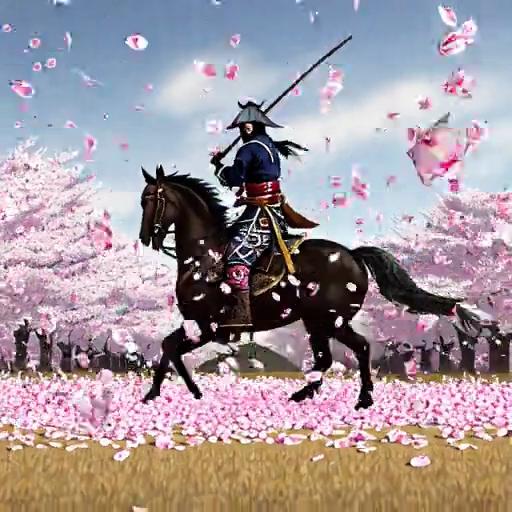}
    \includegraphics[width=0.22\textwidth]{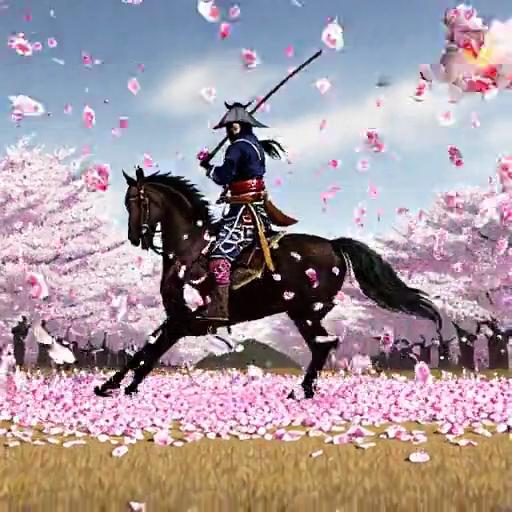}
    \\[1.5em]

        
        
        \textbf{Prompt:} \textit{Two pigs are eating a hotpot.} \\[0.5em]
        \includegraphics[width=0.22\textwidth]{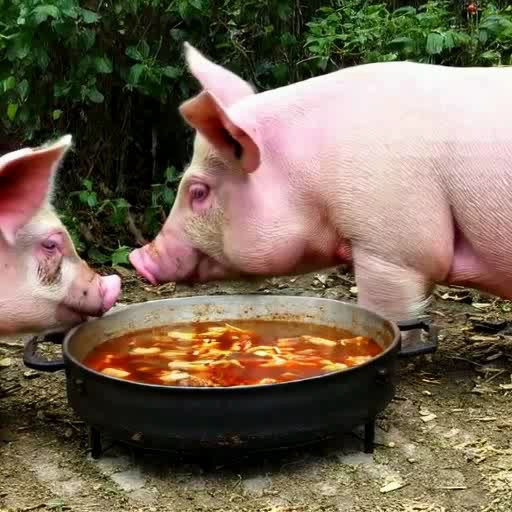}
        \includegraphics[width=0.22\textwidth]{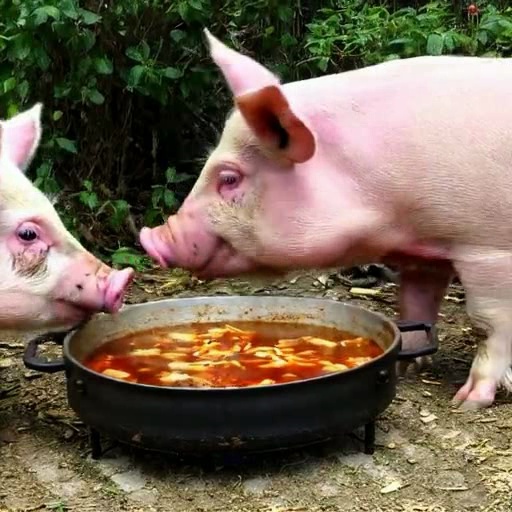}
        \includegraphics[width=0.22\textwidth]{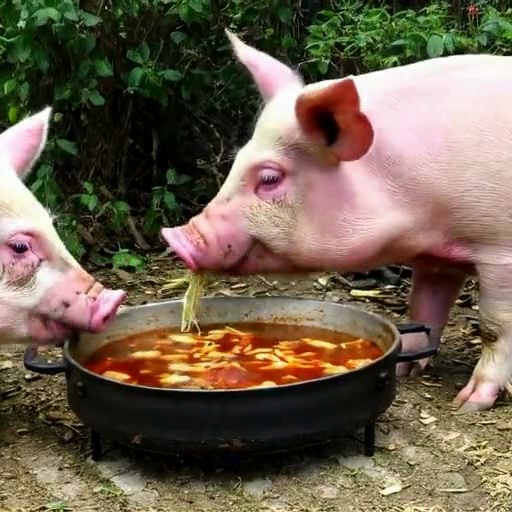}
        \includegraphics[width=0.22\textwidth]{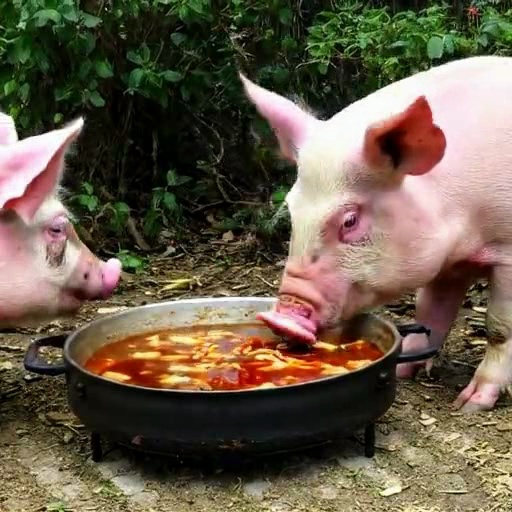}
        \\[1em]
        
        \textbf{Prompt:} \textit{Giant Pandas are eating hot noodles in a Chinese restaurant.} \\[0.5em]
        \includegraphics[width=0.22\textwidth]{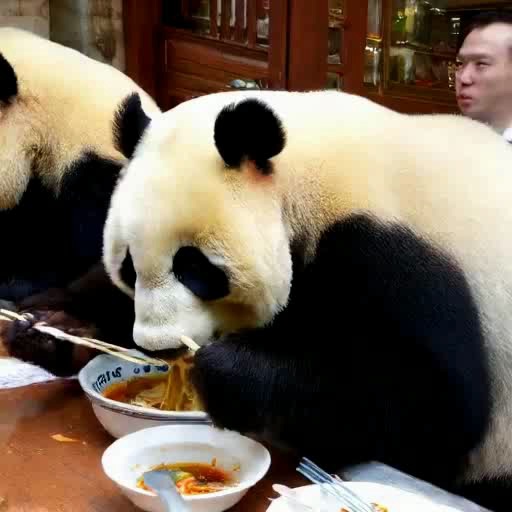}
        \includegraphics[width=0.22\textwidth]{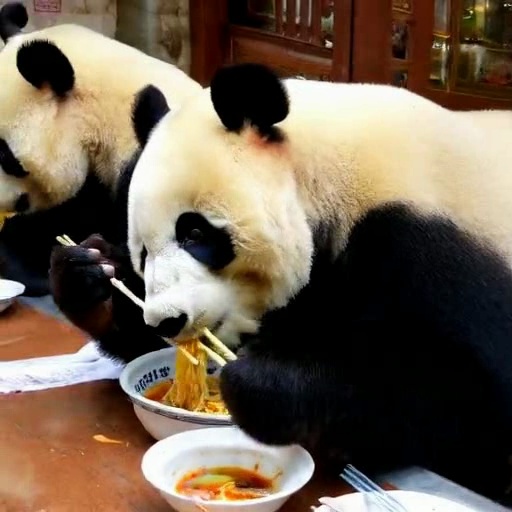}
        \includegraphics[width=0.22\textwidth]{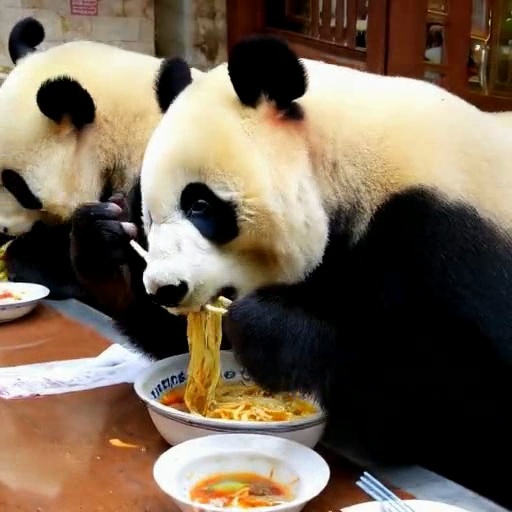}
        \includegraphics[width=0.22\textwidth]{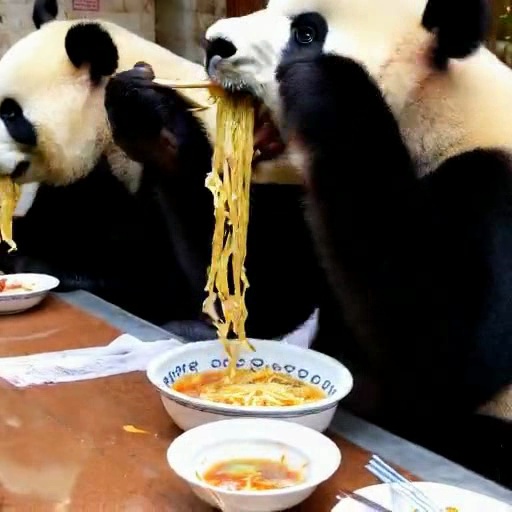}
        \\[1em]

    \end{minipage}
\end{figure*}

\begin{figure*}[htbp]
    \centering
    \begin{minipage}{\textwidth}
        \centering
        \setlength{\fboxrule}{1.5pt}
        \fcolorbox{darkgreen}{lightcream}{
            \parbox{\textwidth}{
                \centering \textbf{\Large \textcolor{darkgreen}{Text-to-Video}}
            }
        } \\[1em]

        \textbf{Prompt:} \textit{A zoom-in on a clock face, focusing on the intricate movement of the hands and the ticking mechanism inside.} \\[0.5em]
        \includegraphics[width=0.22\textwidth]{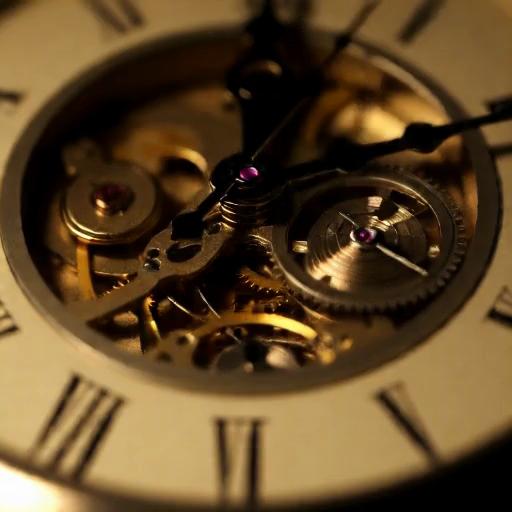}
        \includegraphics[width=0.22\textwidth]{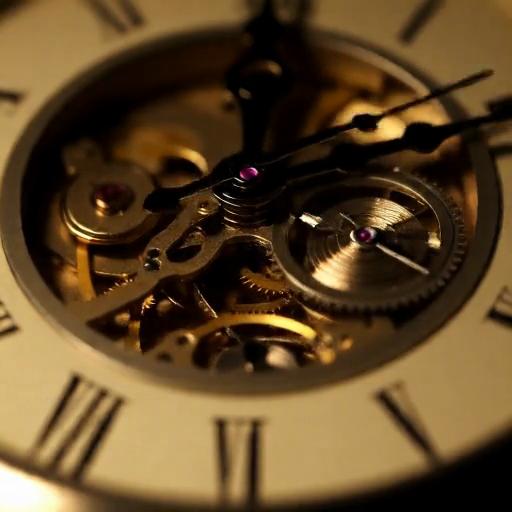}
        \includegraphics[width=0.22\textwidth]{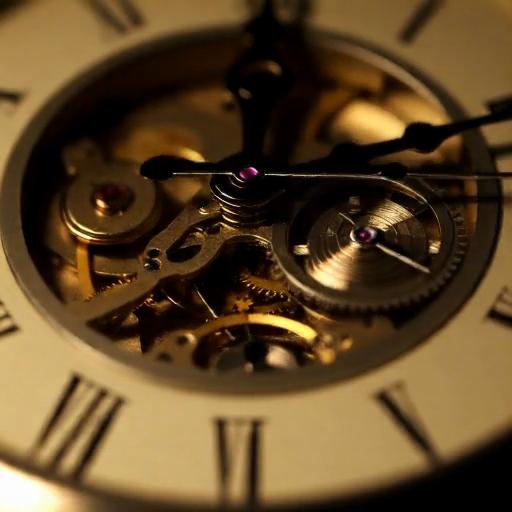}
        \includegraphics[width=0.22\textwidth]{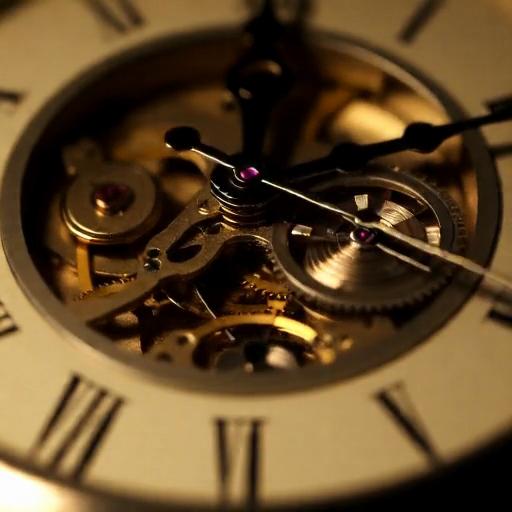}
        \\[1em]

        \textbf{Prompt:} \textit{Robots move efficiently through a futuristic laboratory, adjusting holographic displays and conducting experiments, while scientists observe and interact with the high-tech equipment.} \\[0.5em]
        \includegraphics[width=0.22\textwidth]{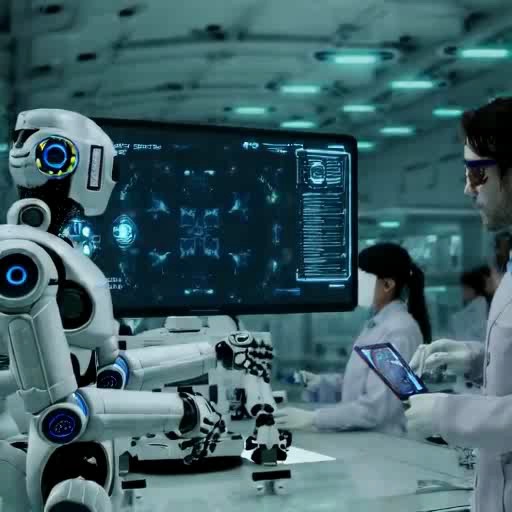}
        \includegraphics[width=0.22\textwidth]{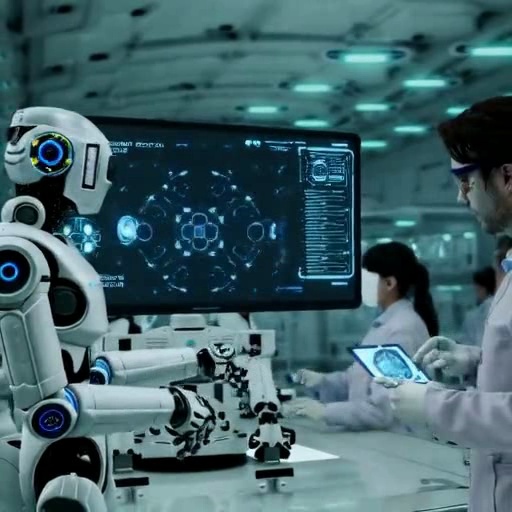}
        \includegraphics[width=0.22\textwidth]{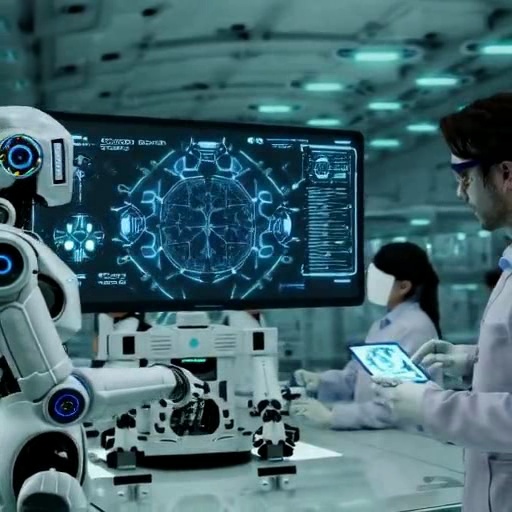}
        \includegraphics[width=0.22\textwidth]{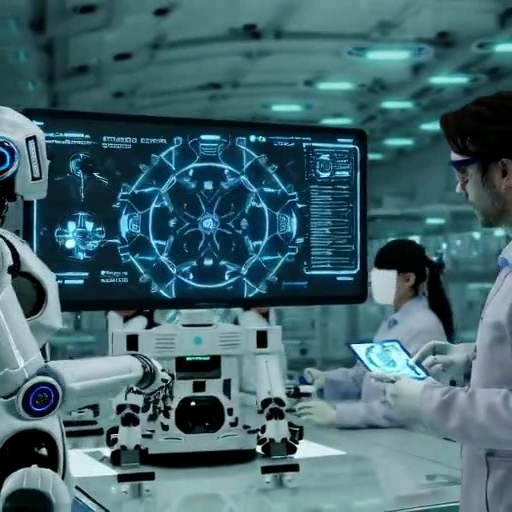}
        \\[1em]

    \textbf{Prompt:} \textit{A robotic arm wielding a glowing sword, battling a shadowy figure in a high-tech dojo, each strike creating sparks that light up the space.} \\[0.5em]
    \includegraphics[width=0.22\textwidth]{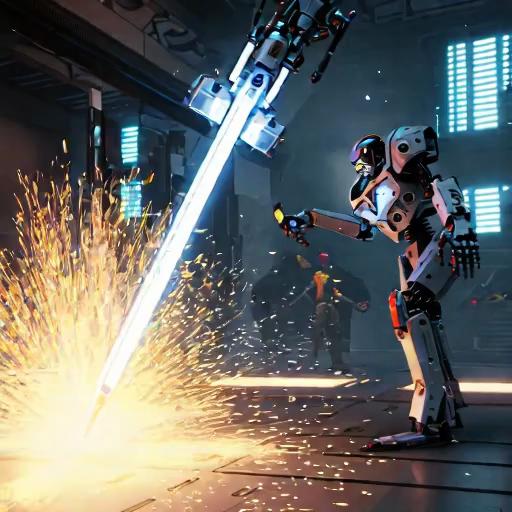}
    \includegraphics[width=0.22\textwidth]{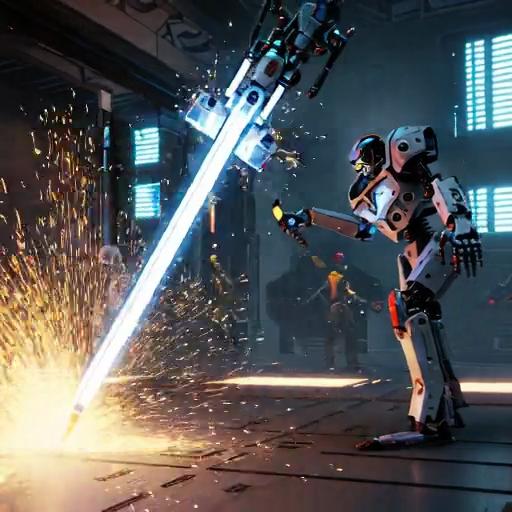}
    \includegraphics[width=0.22\textwidth]{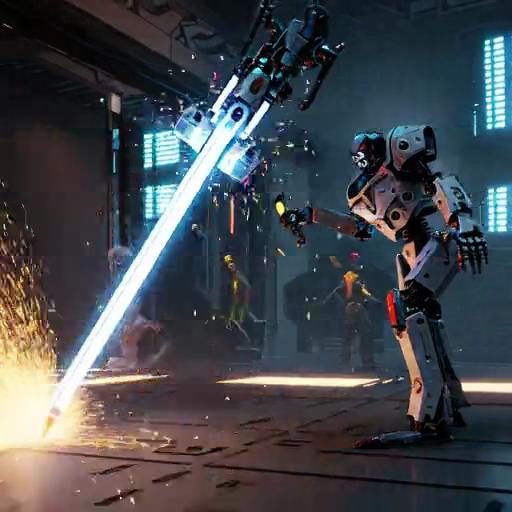}
    \includegraphics[width=0.22\textwidth]{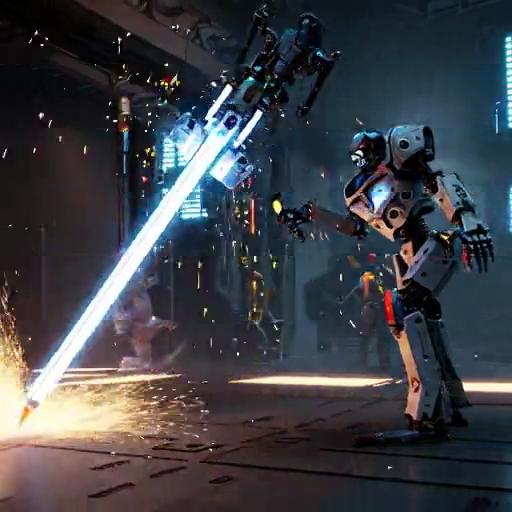}
    \\[1.5em]

    \textbf{Prompt:} \textit{A city skyline reflected in the water, but the reflection shows an alternate world with flying cars, towering robots, and futuristic architecture.} \\[0.5em]
    \includegraphics[width=0.22\textwidth]{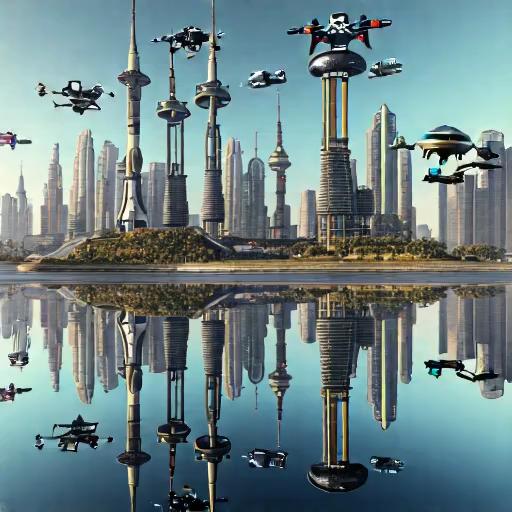}
    \includegraphics[width=0.22\textwidth]{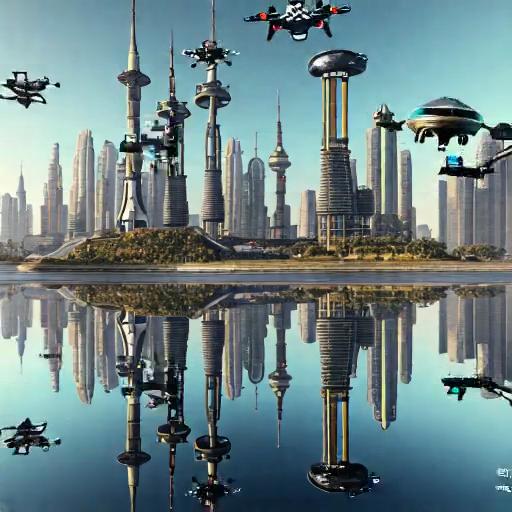}
    \includegraphics[width=0.22\textwidth]{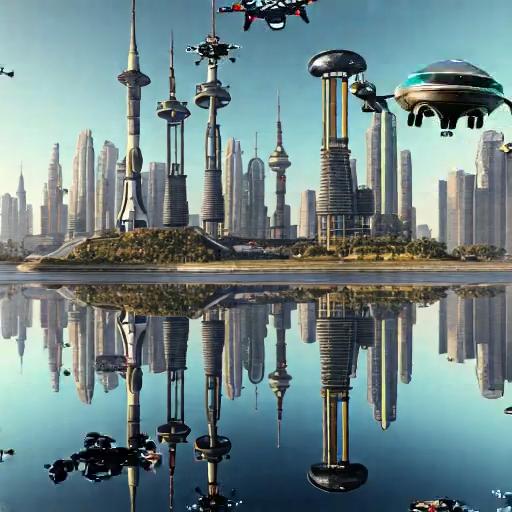}
    \includegraphics[width=0.22\textwidth]{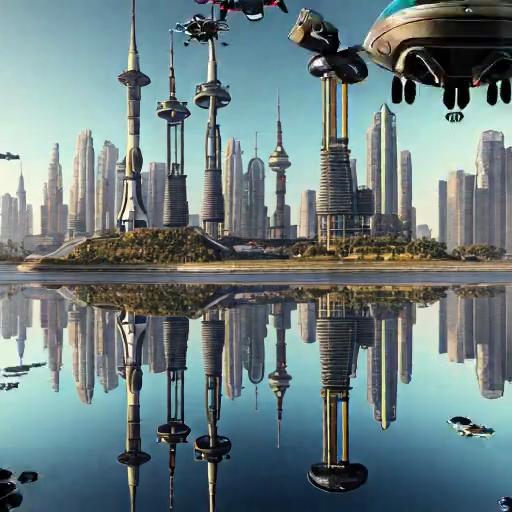}
    \\[1.5em]
    
    \end{minipage}
\end{figure*}

\begin{figure*}[htbp]
    \centering
    \fcolorbox{darkgreen}{lightcream}{
        \parbox{\textwidth}{
            \centering \textbf{\Large \textcolor{darkgreen}{Text-to-Video}}
        }
    }
    \\[1em]

        \textbf{Prompt:} \textit{A dog dressed as a chef, expertly flipping pancakes in a kitchen.} \\[0.5em]
        \includegraphics[width=0.22\textwidth]{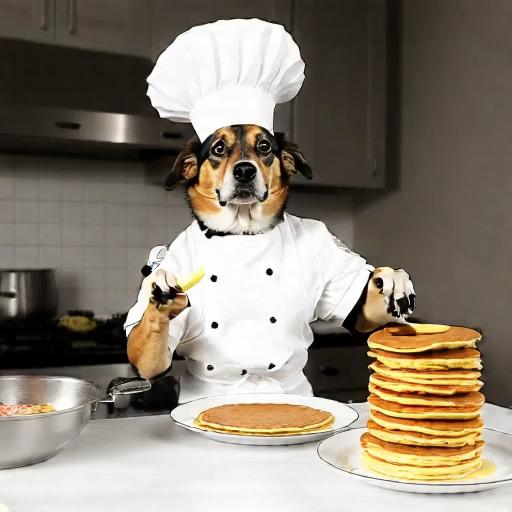}
        \includegraphics[width=0.22\textwidth]{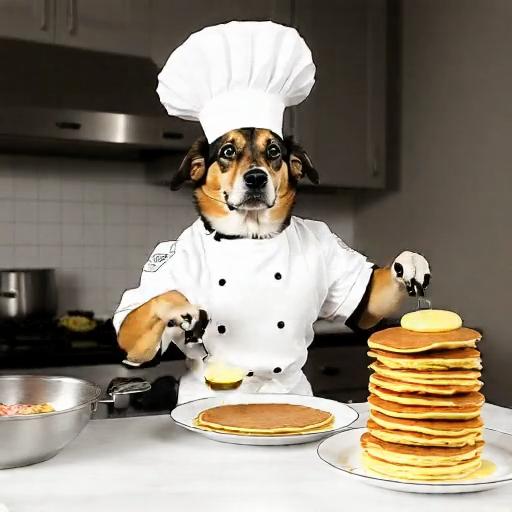}
        \includegraphics[width=0.22\textwidth]{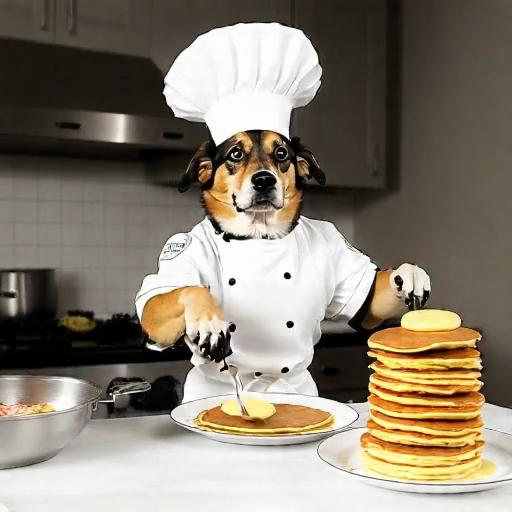}
        \includegraphics[width=0.22\textwidth]{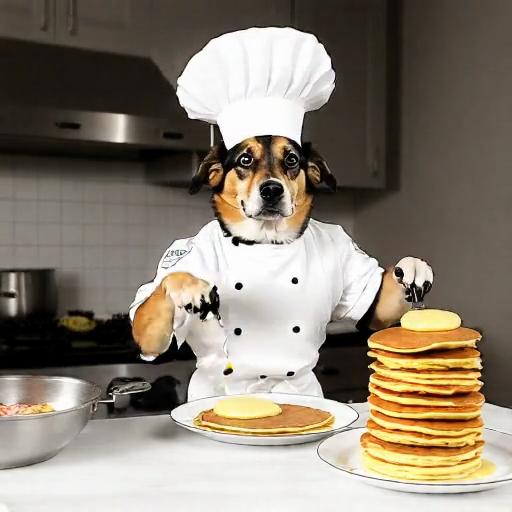}
        \\[1em]
        
    \textbf{Prompt:} \textit{A motocross bike accelerating out of a tight turn on a dirt track.} \\[0.5em]
        \includegraphics[width=0.22\textwidth]{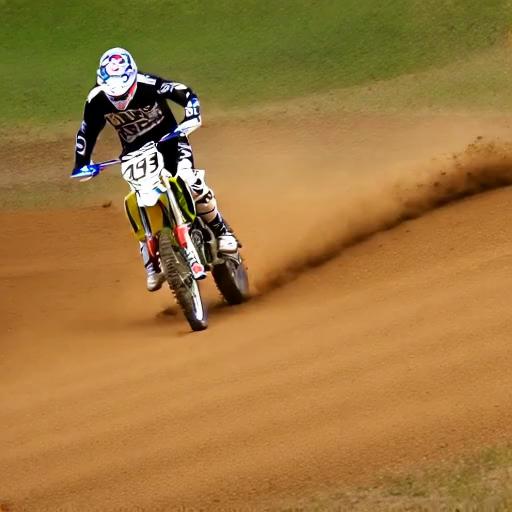}
        \includegraphics[width=0.22\textwidth]{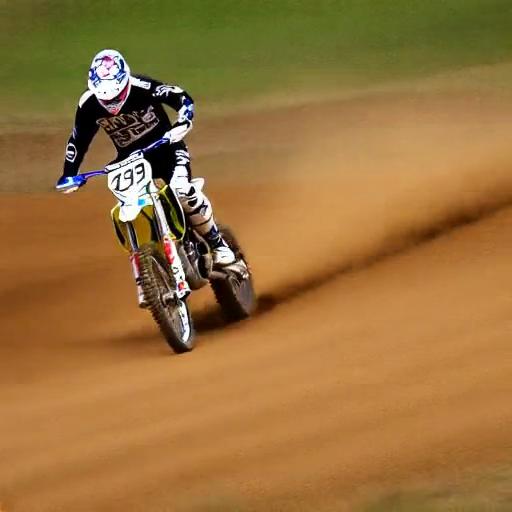}
        \includegraphics[width=0.22\textwidth]{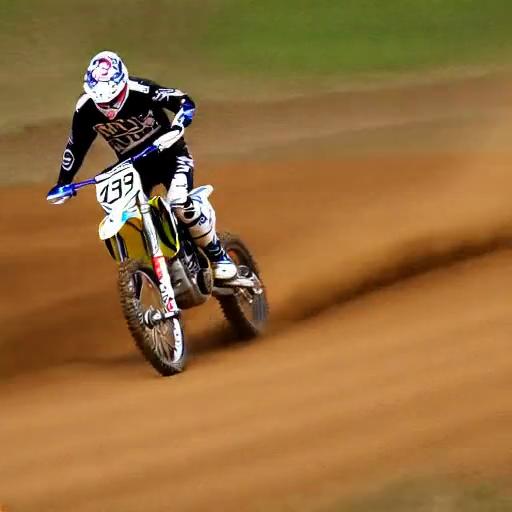}
        \includegraphics[width=0.22\textwidth]{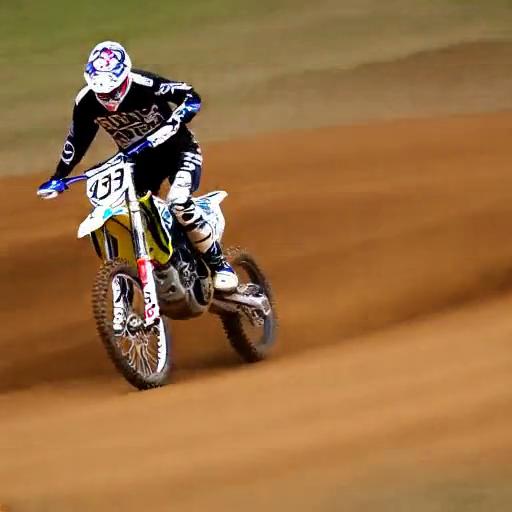}
        \\[1em]

    \textbf{Prompt:} \textit{A snowboarder performing a dramatic backflip over a frozen lake, landing gracefully and leaving a trail of sparkling ice dust in the air.} \\[0.5em]
    \includegraphics[width=0.22\textwidth]{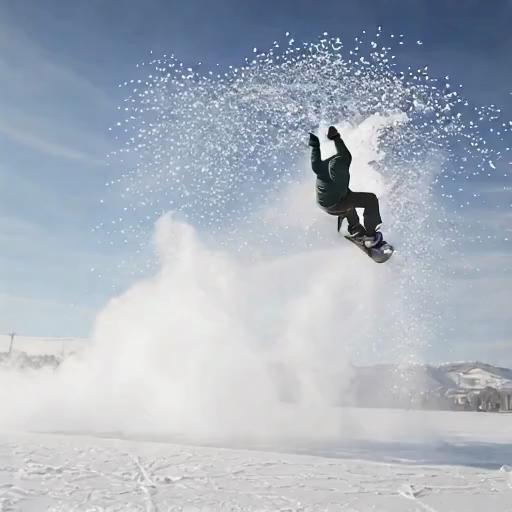}
    \includegraphics[width=0.22\textwidth]{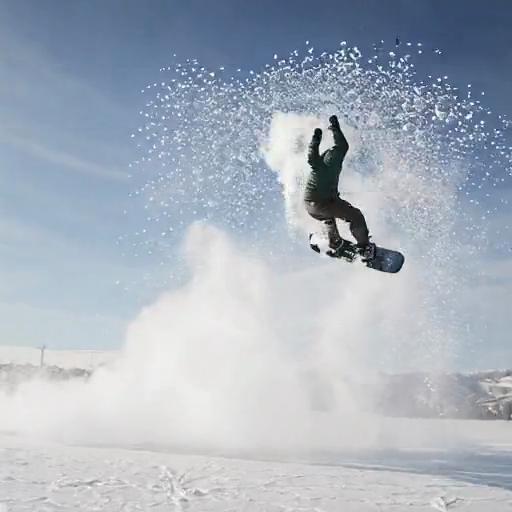}
    \includegraphics[width=0.22\textwidth]{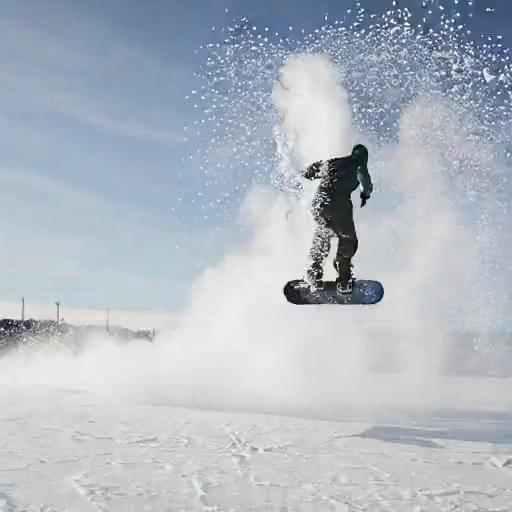}
    \includegraphics[width=0.22\textwidth]{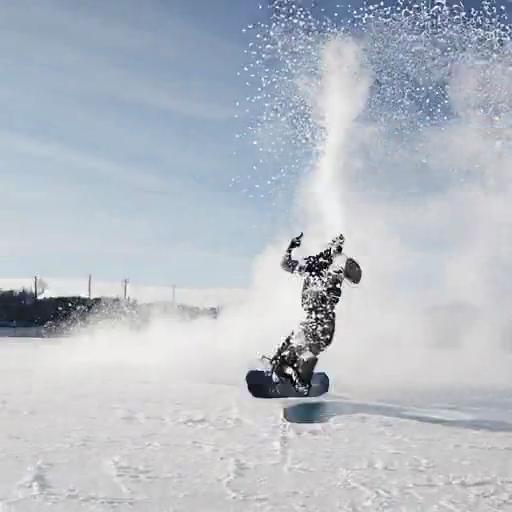}
    \\[1.5em]

    \textbf{Prompt:} \textit{A surfer accelerating on a wave, carving through the water.} \\[0.5em]
        \includegraphics[width=0.22\textwidth]{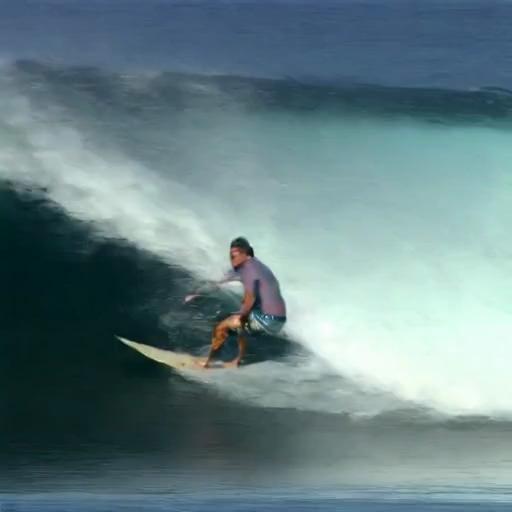}
        \includegraphics[width=0.22\textwidth]{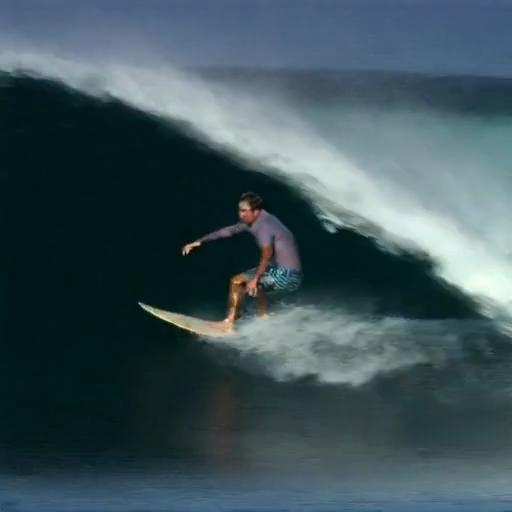}
        \includegraphics[width=0.22\textwidth]{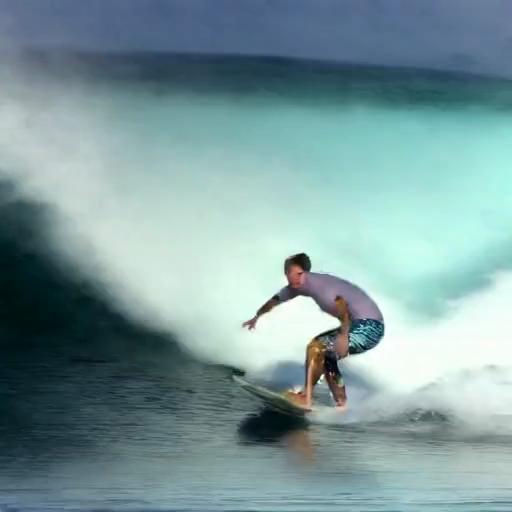}
        \includegraphics[width=0.22\textwidth]{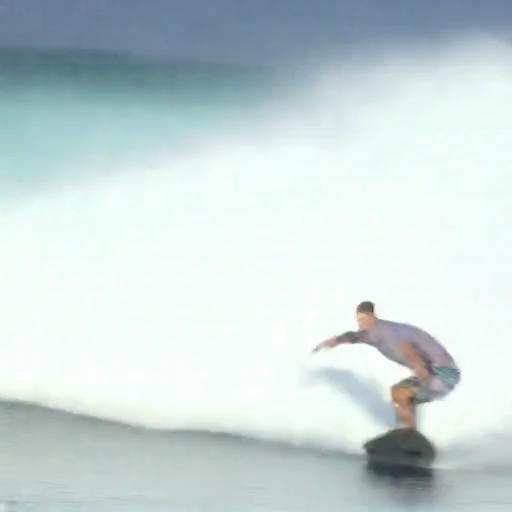}
        \\[1em]
\end{figure*}

\begin{figure*}[htbp]
    \centering
    \begin{minipage}{\textwidth}
        \centering
        \setlength{\fboxrule}{1.5pt}
        \fcolorbox{darkgreen}{lightcream}{
            \parbox{\textwidth}{
                \centering \textbf{\Large \textcolor{darkgreen}{Text-to-Video}}
            }
        } \\[1em]


        \textbf{Prompt:} \textit{A person dancing with their own shadow, which has come to life.} \\[0.5em]
        \includegraphics[width=0.22\textwidth]{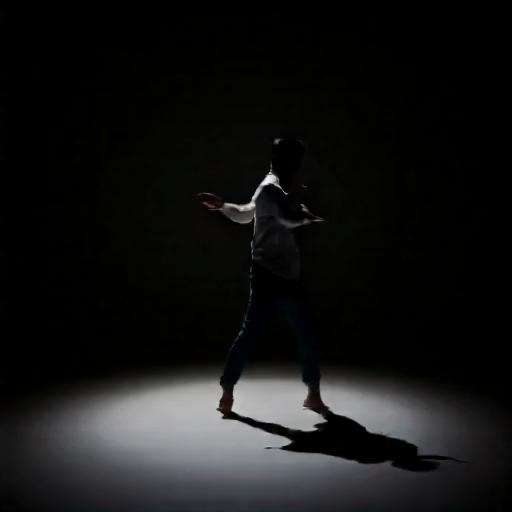}
        \includegraphics[width=0.22\textwidth]{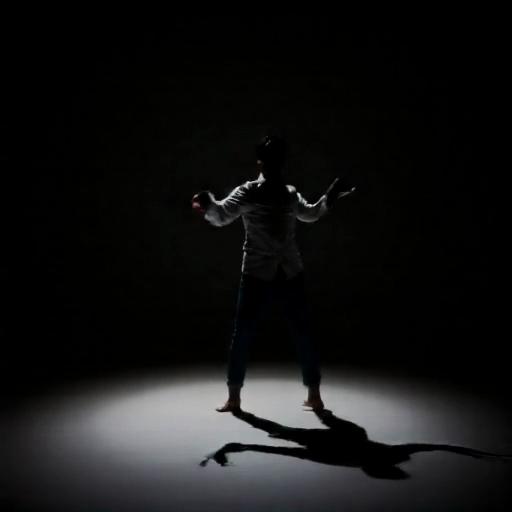}
        \includegraphics[width=0.22\textwidth]{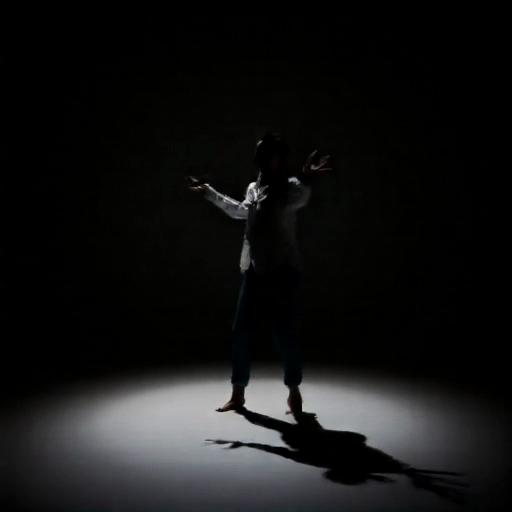}
        \includegraphics[width=0.22\textwidth]{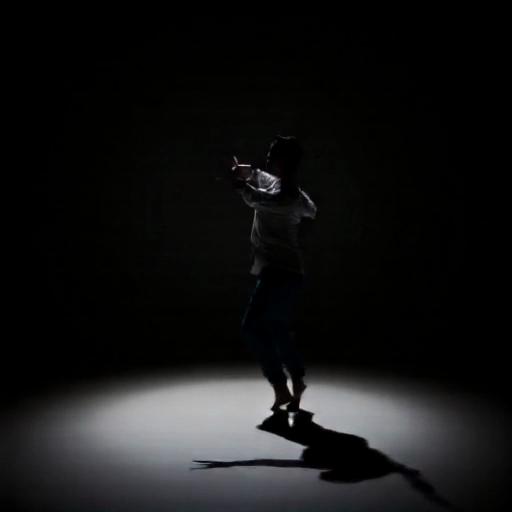}
        \\[1em]
        \textbf{Prompt:} \textit{A bobsled team accelerating down an icy track.} \\[0.5em]
        \includegraphics[width=0.22\textwidth]{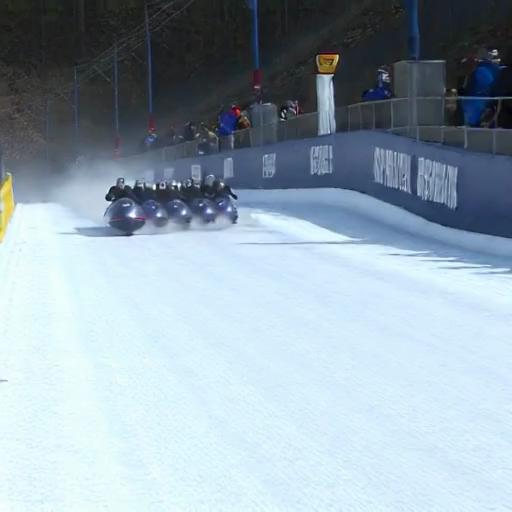}
        \includegraphics[width=0.22\textwidth]{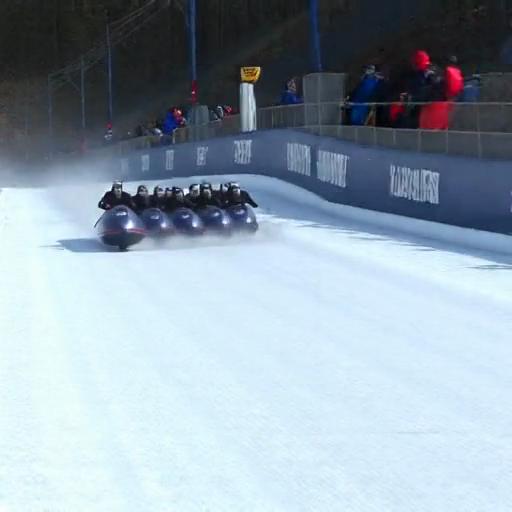}
        \includegraphics[width=0.22\textwidth]{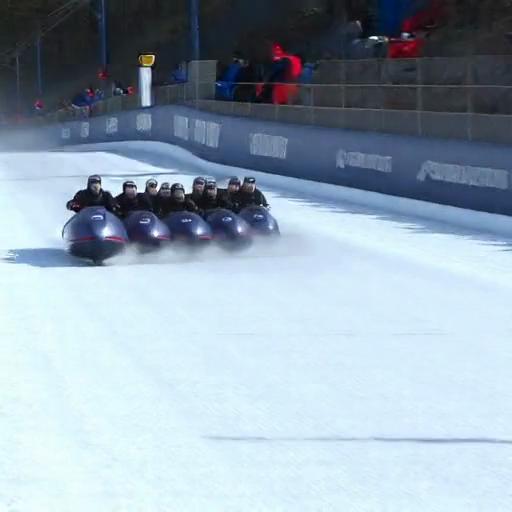}
        \includegraphics[width=0.22\textwidth]{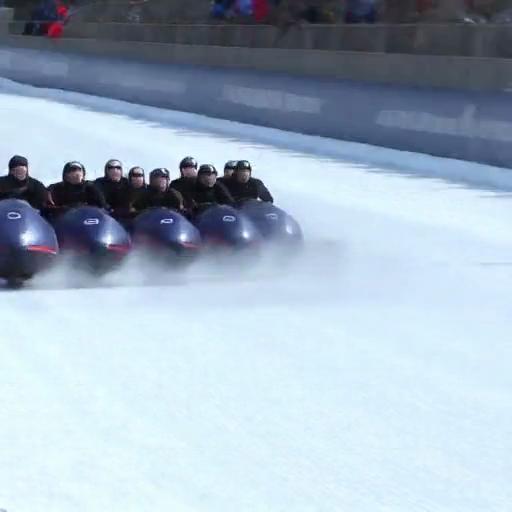}
        \\[1em]
        \textbf{Prompt:} \textit{A cyclist accelerating out of the saddle during a steep climb.} \\[0.5em]
        \includegraphics[width=0.22\textwidth]{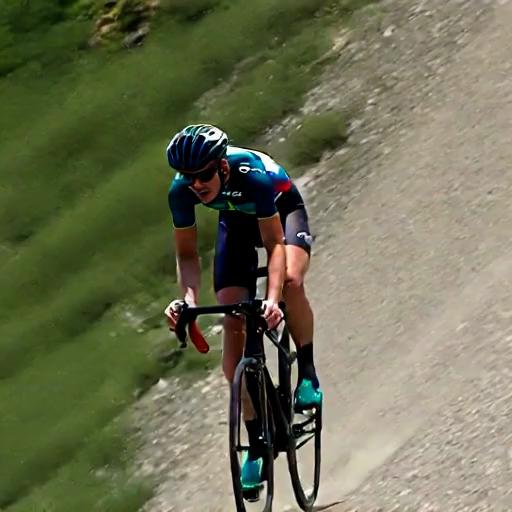}
        \includegraphics[width=0.22\textwidth]{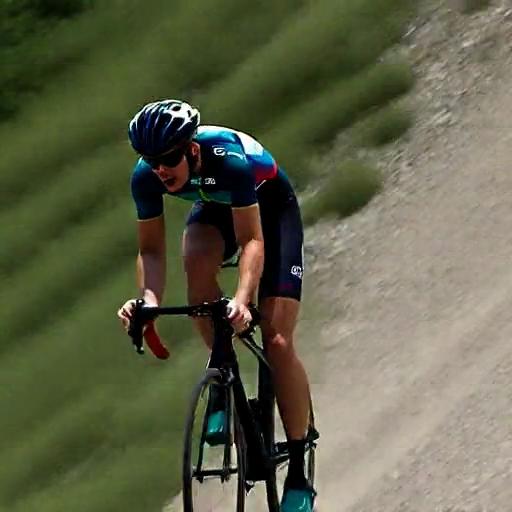}
        \includegraphics[width=0.22\textwidth]{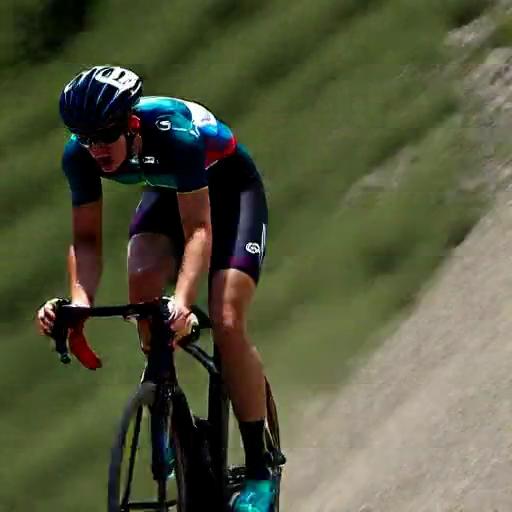}
        \includegraphics[width=0.22\textwidth]{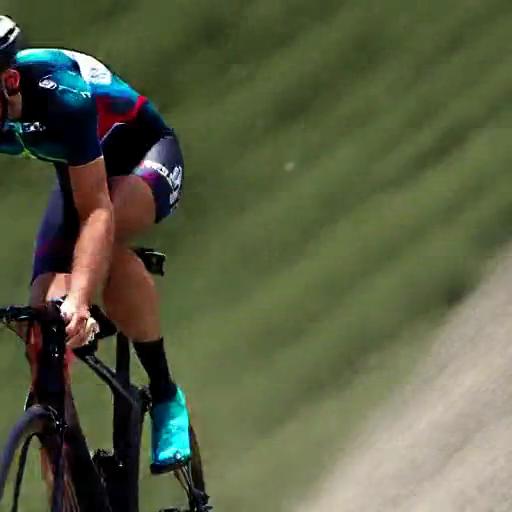}
        \\[1em]
        
        \textbf{Prompt:} \textit{A speed skater accelerating during a short track race.} \\[0.5em]
        \includegraphics[width=0.22\textwidth]{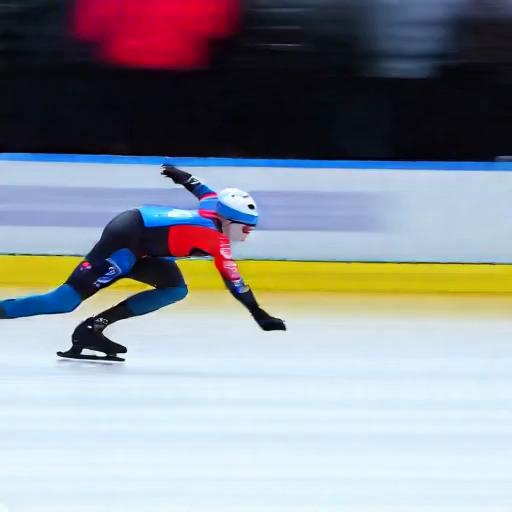}
        \includegraphics[width=0.22\textwidth]{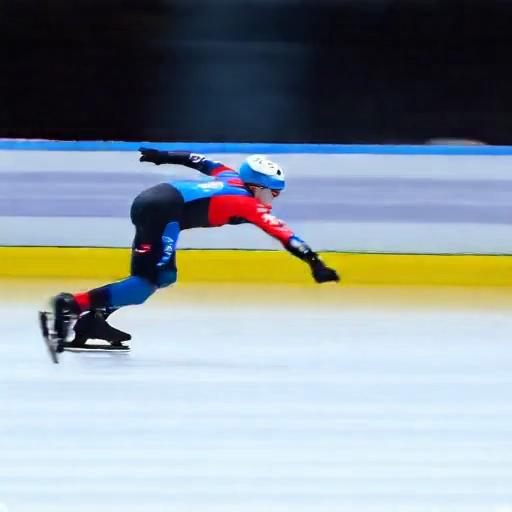}
        \includegraphics[width=0.22\textwidth]{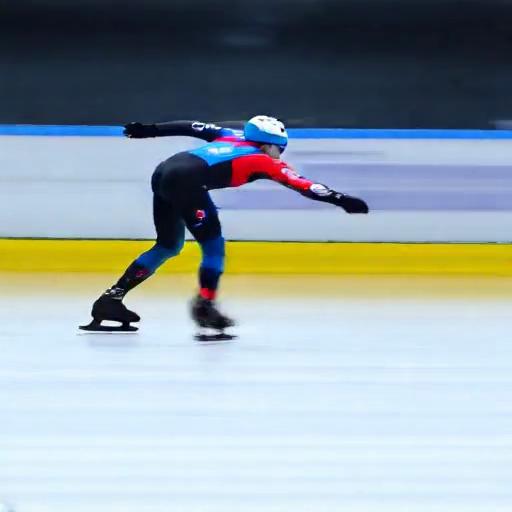}
        \includegraphics[width=0.22\textwidth]{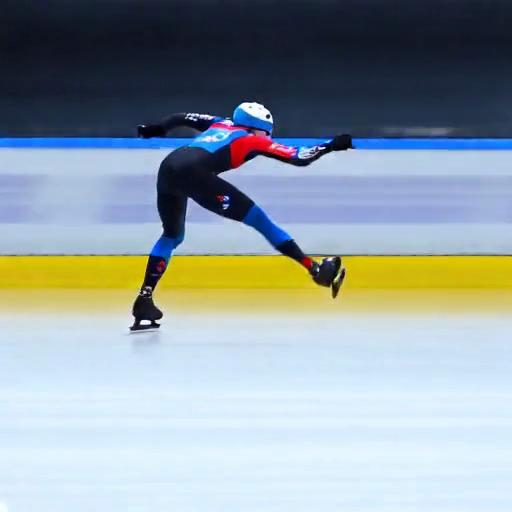}
        \\[1em]
    \end{minipage}
\end{figure*}

\begin{figure*}[htbp]
    \centering
    \centering
    \fcolorbox{darkgreen}{lightcream}{
        \parbox{\textwidth}{
            \centering \textbf{\Large \textcolor{darkgreen}{Text-Image-to-Video}}
        }
    }
    \\[1em]

    \centering
    \textbf{Prompt:} \textit{Reflections in the window of a train traveling through the Tokyo suburbs.} \\[0.5em]
    \begin{minipage}{\textwidth}
        \centering
        \begin{minipage}[t]{0.18\textwidth}
            \centering
            \fcolorbox{blue}{white}{\includegraphics[width=\linewidth]{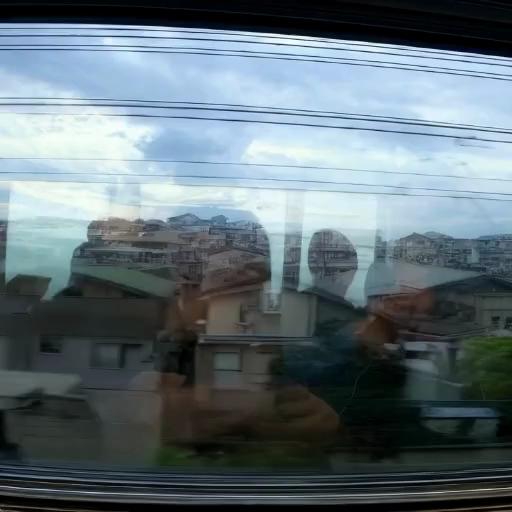}}
            \\[0.5em]
            \textbf{Reference Image}
        \end{minipage}
        \hfill
        \includegraphics[width=0.18\textwidth]{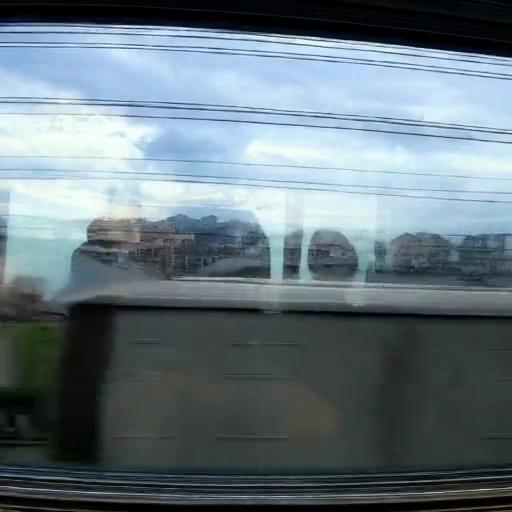}
        \includegraphics[width=0.18\textwidth]{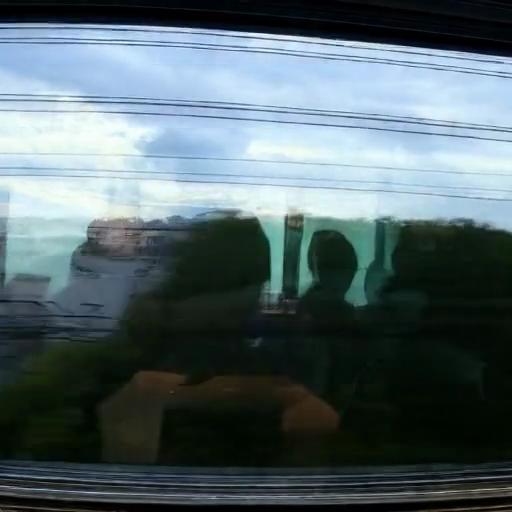}
        \includegraphics[width=0.18\textwidth]{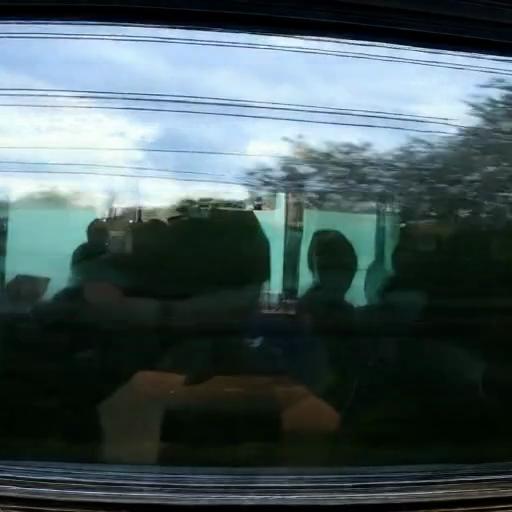}
        \includegraphics[width=0.18\textwidth]{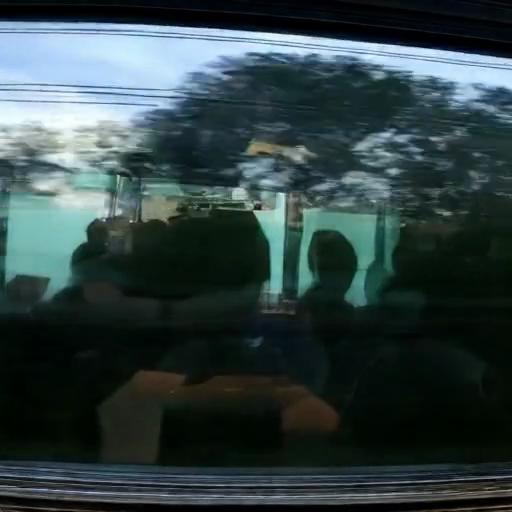}
    \end{minipage}
    \\[1em]


    \centering
    \textbf{Prompt:} \textit{The Glenfinnan Viaduct is a historic railway bridge in Scotland, UK, that crosses over the west highland line between the towns of Mallaig and Fort William. It is a stunning sight as a steam train leaves the bridge...} \\[0.5em]
    \begin{minipage}{\textwidth}
        \centering
        \begin{minipage}[t]{0.18\textwidth}
            \centering
            \fcolorbox{blue}{white}{\includegraphics[width=\linewidth]{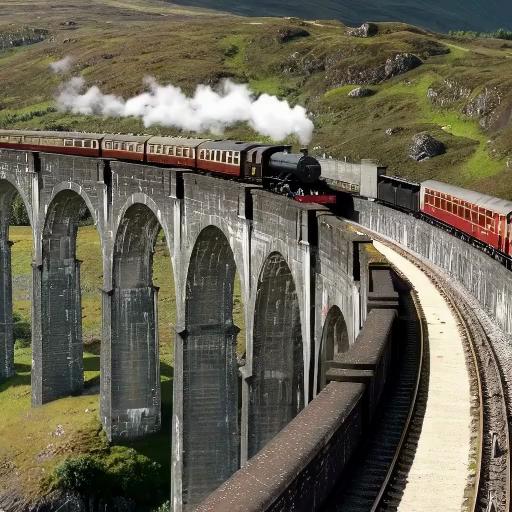}}
            \\[0.5em]
            \textbf{Reference Image}
        \end{minipage}
        \hfill
        \includegraphics[width=0.18\textwidth]{ti2v_figures/0008_bridge/frame_000000.jpg}
        \includegraphics[width=0.18\textwidth]{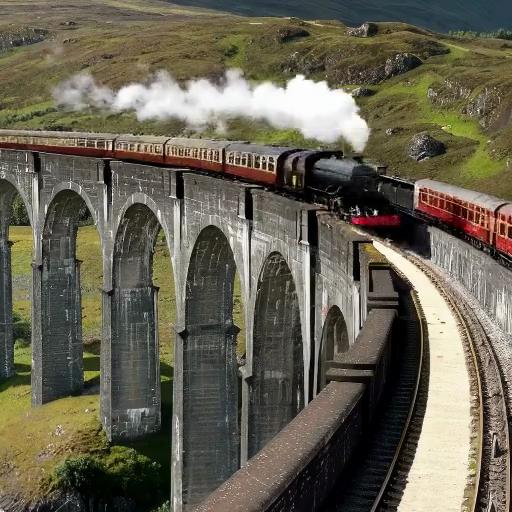}
        \includegraphics[width=0.18\textwidth]{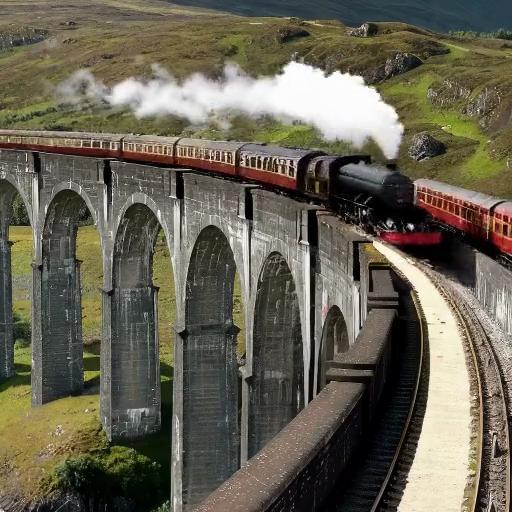}
        \includegraphics[width=0.18\textwidth]{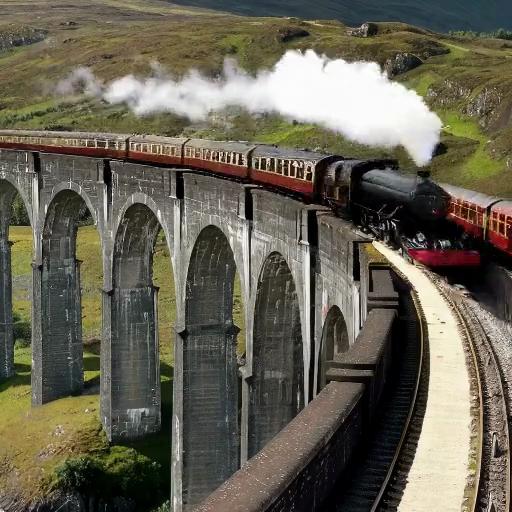}
    \end{minipage}
    \\[1em]

    \centering
    \textbf{Prompt:} \textit{The camera follows behind a white vintage SUV with a black roof rack as it speeds up a steep dirt road surrounded by pine trees on a steep mountain slope, dust kicks up from it's tires, the sunlight shines on the SUV...} \\[0.5em]
    \begin{minipage}{\textwidth}
        \centering
        \begin{minipage}[t]{0.18\textwidth}
            \centering
            \fcolorbox{blue}{white}{\includegraphics[width=\linewidth]{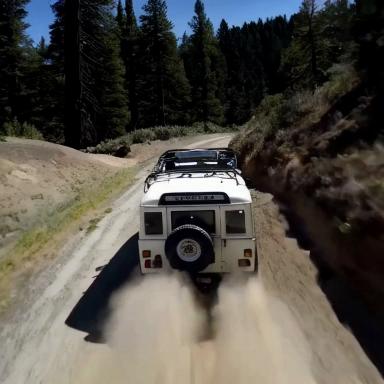}}
            \\[0.5em]
            \textbf{Reference Image}
        \end{minipage}
        \hfill
        \includegraphics[width=0.18\textwidth]{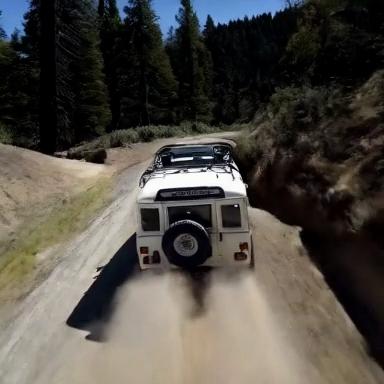}
        \includegraphics[width=0.18\textwidth]{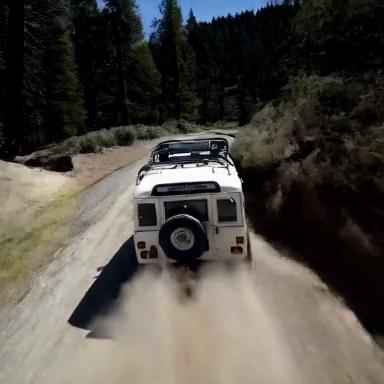}
        \includegraphics[width=0.18\textwidth]{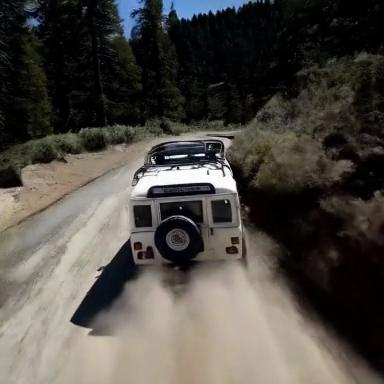}
        \includegraphics[width=0.18\textwidth]{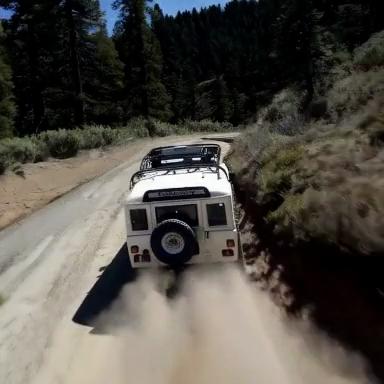}
    \end{minipage}
    \\[1em]

    \centering
    \textbf{Prompt:} \textit{Photorealistic closeup video of two pirate ships battling each other as they sail inside a cup of coffee.} \\[0.5em]
    \begin{minipage}{\textwidth}
        \centering
        \begin{minipage}[t]{0.18\textwidth}
            \centering
            \fcolorbox{blue}{white}{\includegraphics[width=\linewidth]{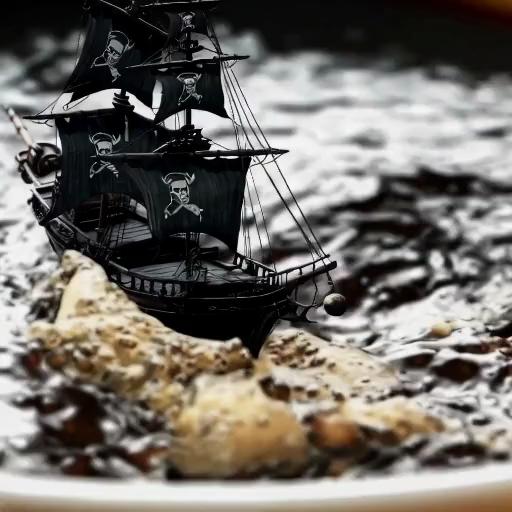}}
            \\[0.5em]
            \textbf{Reference Image}
        \end{minipage}
        \hfill
        \includegraphics[width=0.18\textwidth]{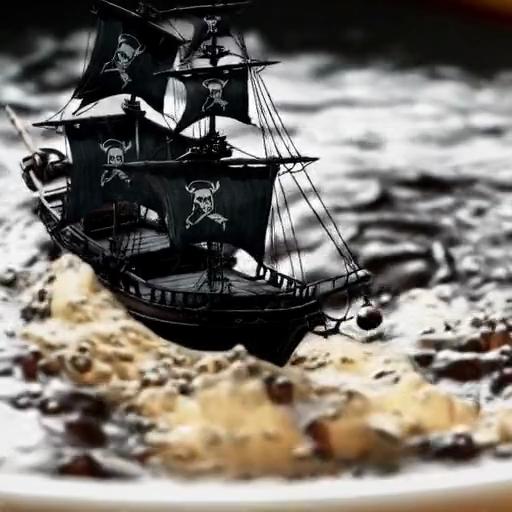}
        \includegraphics[width=0.18\textwidth]{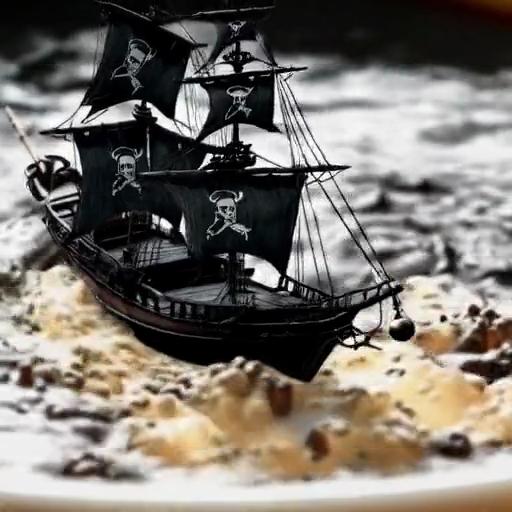}
        \includegraphics[width=0.18\textwidth]{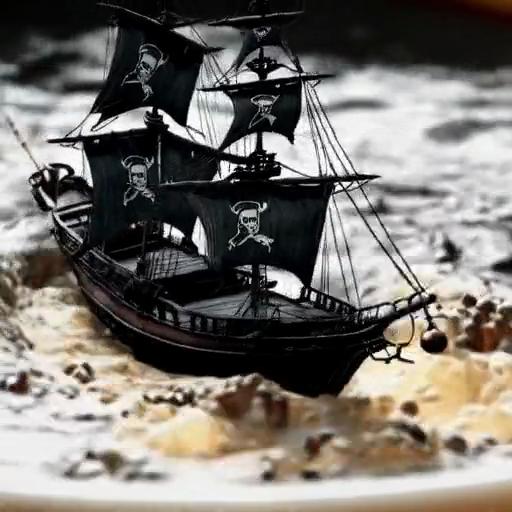}
        \includegraphics[width=0.18\textwidth]{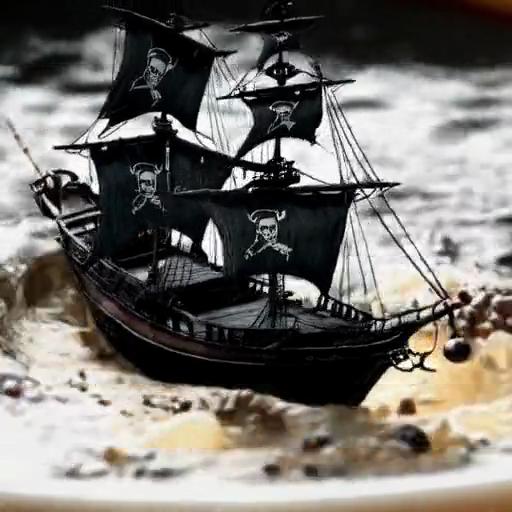}
    \end{minipage}
    \\[1em]
\end{figure*}

\begin{figure*}[htbp]
    \centering
    \centering
    \fcolorbox{darkgreen}{lightcream}{
        \parbox{\textwidth}{
            \centering \textbf{\Large \textcolor{darkgreen}{Text-Image-to-Video}}
        }
    }
    \\[1em]
    
    \centering
    \textbf{Prompt:} \textit{A litter of golden retriever puppies playing in the snow. Their heads pop out of the snow, covered in.} \\[0.5em]
    \begin{minipage}{\textwidth}
        \centering
        \begin{minipage}[t]{0.18\textwidth}
            \centering
            \fcolorbox{blue}{white}{\includegraphics[width=\linewidth]{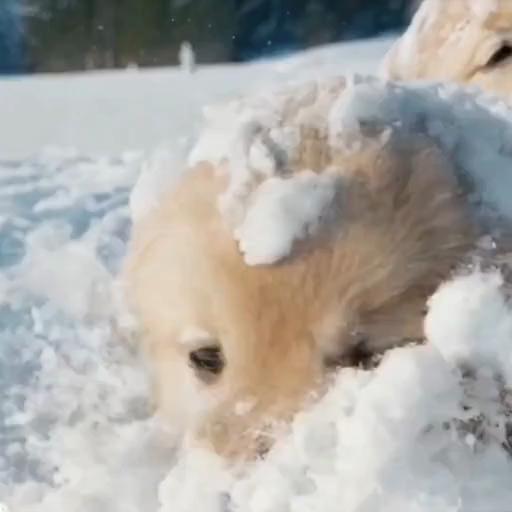}}
            \\[0.5em]
            \textbf{Reference Image}
        \end{minipage}
        \hfill
        \includegraphics[width=0.18\textwidth]{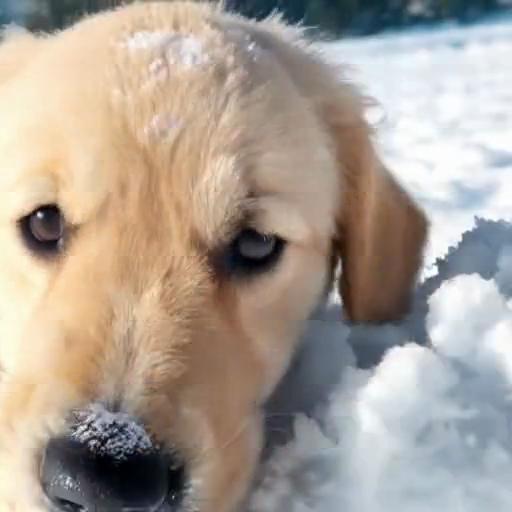}
        \includegraphics[width=0.18\textwidth]{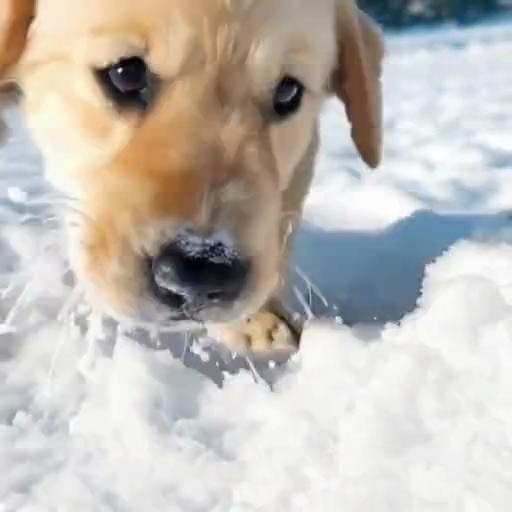}
        \includegraphics[width=0.18\textwidth]{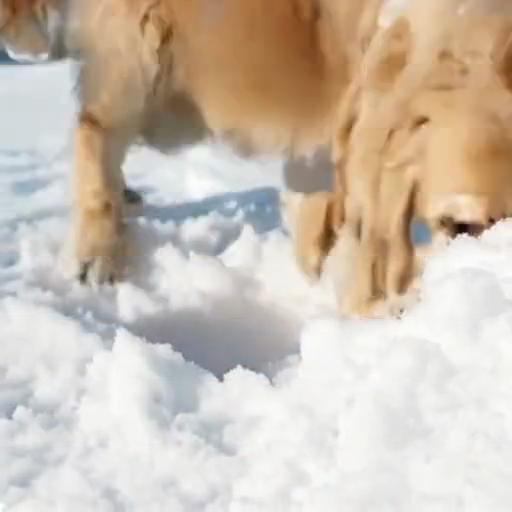}
        \includegraphics[width=0.18\textwidth]{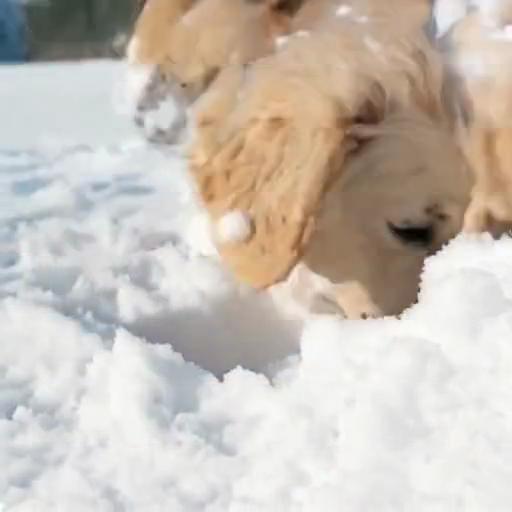}
    \end{minipage}
    \\[1em]
    
    \centering
    \textbf{Prompt:} \textit{An adorable happy otter confidently stands on a surfboard wearing a yellow lifejacket, riding along turquoise tropical waters near lush tropical islands, 3D digital render art style.} \\[0.5em]
    \begin{minipage}{\textwidth}
        \centering
        \begin{minipage}[t]{0.18\textwidth}
            \centering
            \fcolorbox{blue}{white}{\includegraphics[width=\linewidth]{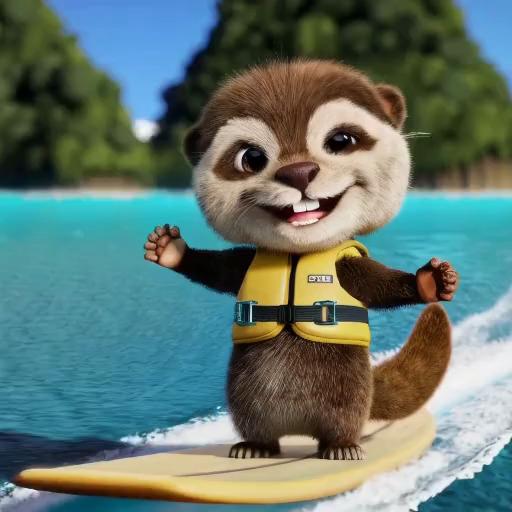}}
            \\[0.5em]
            \textbf{Reference Image}
        \end{minipage}
        \hfill
        \includegraphics[width=0.18\textwidth]{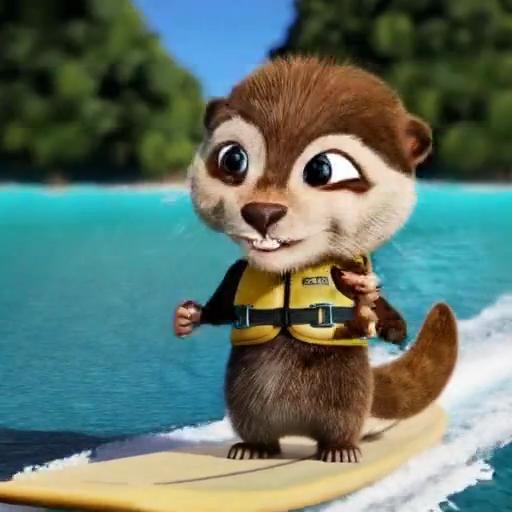}
        \includegraphics[width=0.18\textwidth]{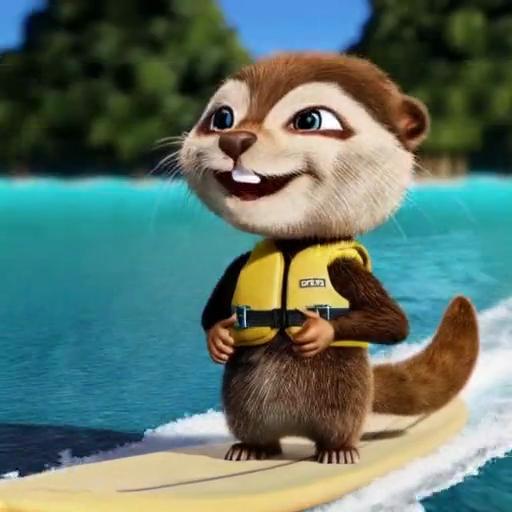}
        \includegraphics[width=0.18\textwidth]{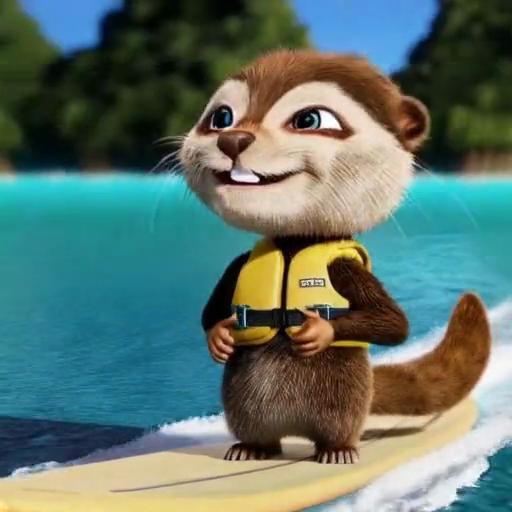}
        \includegraphics[width=0.18\textwidth]{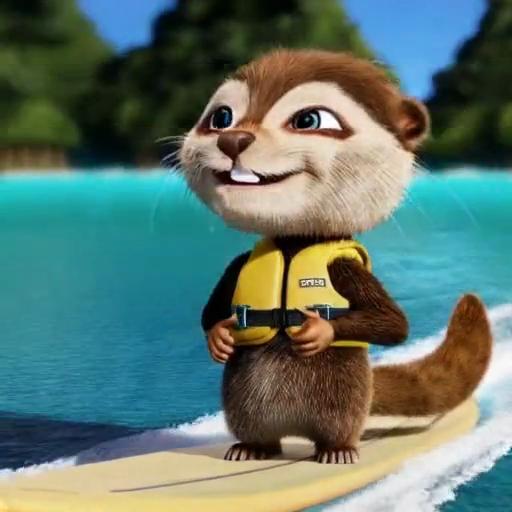}
    \end{minipage}
    \\[1em]
    
    \centering
    \textbf{Prompt:} \textit{A dog dressed as a chef, expertly flipping pancakes in a kitchen.} \\[0.5em]
    \begin{minipage}{\textwidth}
        \centering
        \begin{minipage}[t]{0.18\textwidth}
            \centering
            \fcolorbox{blue}{white}{\includegraphics[width=\linewidth]{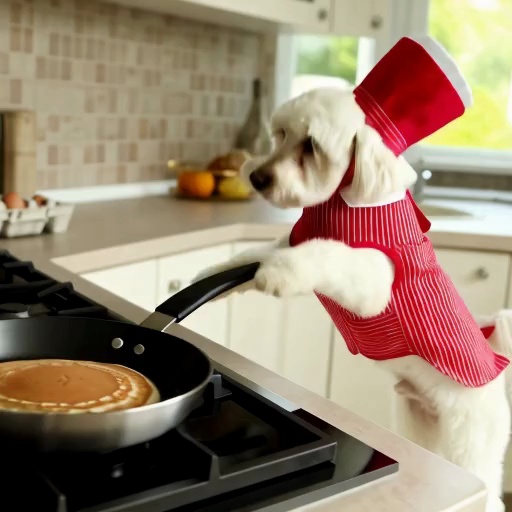}}
            \\[0.5em]
            \textbf{Reference Image}
        \end{minipage}
        \hfill
        \includegraphics[width=0.18\textwidth]{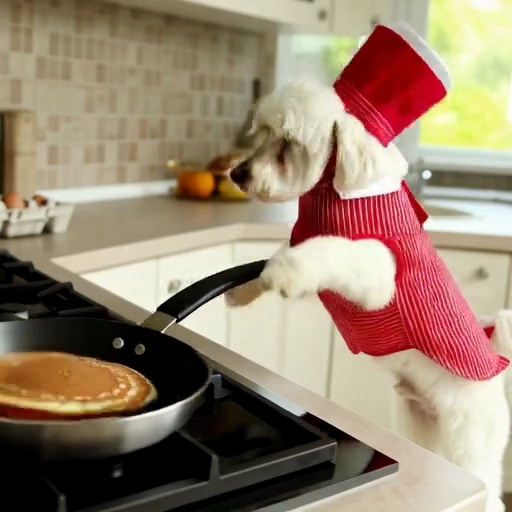}
        \includegraphics[width=0.18\textwidth]{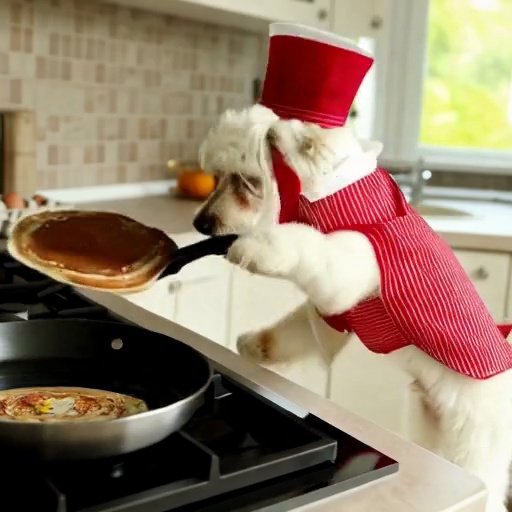}
        \includegraphics[width=0.18\textwidth]{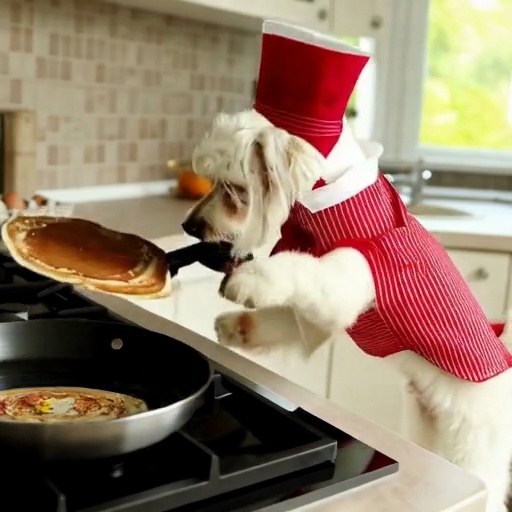}
        \includegraphics[width=0.18\textwidth]{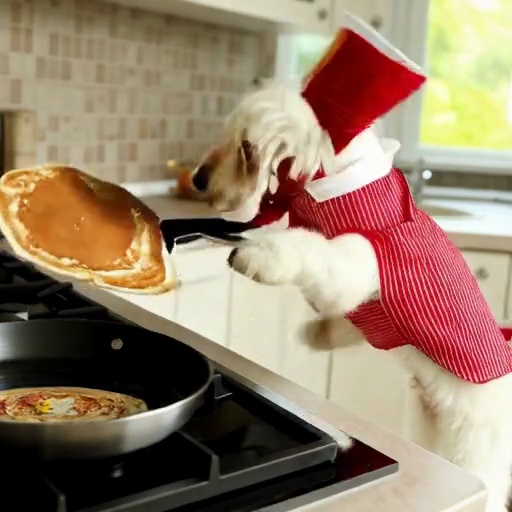}
    \end{minipage}

    \centering
    \textbf{Prompt:} \textit{A skeleton wearing a flower hat and sunglasses dances in the wild at sunset.} \\[0.5em]
    \begin{minipage}{\textwidth}
        \centering
        \begin{minipage}[t]{0.18\textwidth}
            \centering
            \fcolorbox{blue}{white}{\includegraphics[width=\linewidth]{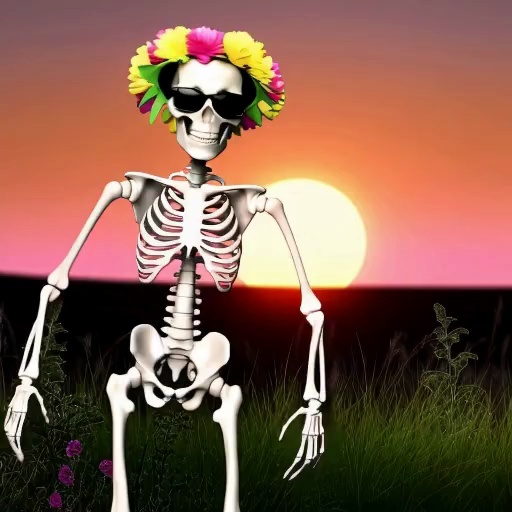}}
            \\[0.5em]
            \textbf{Reference Image}
        \end{minipage}
        \hfill
        \includegraphics[width=0.18\textwidth]{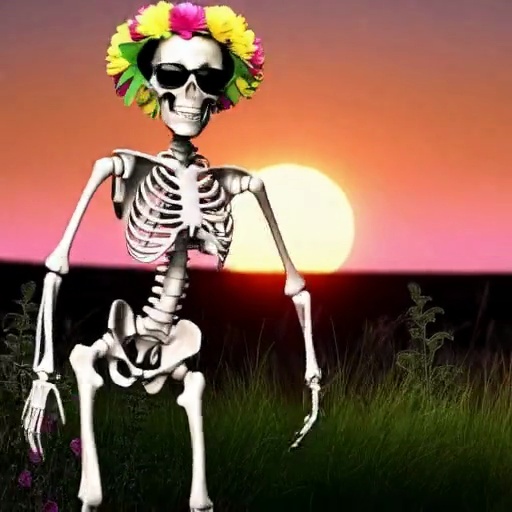}
        \includegraphics[width=0.18\textwidth]{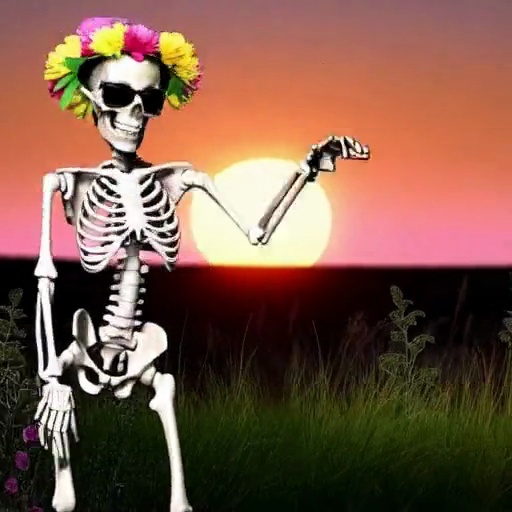}
        \includegraphics[width=0.18\textwidth]{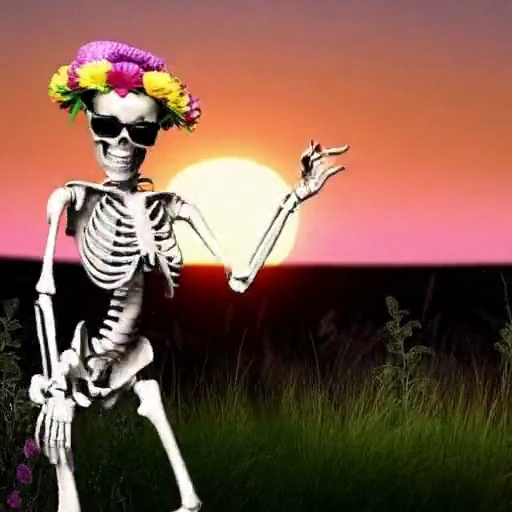}
        \includegraphics[width=0.18\textwidth]{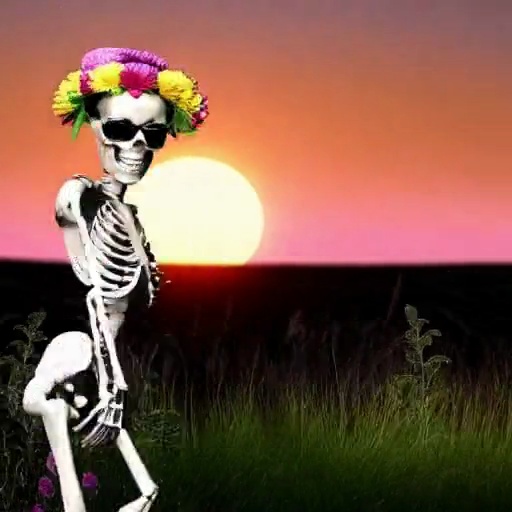}
    \end{minipage}
    \\[1em]
\end{figure*}

\begin{figure*}[htbp]
    \centering
    \fcolorbox{darkgreen}{lightcream}{
        \parbox{\textwidth}{
            \centering \textbf{\Large \textcolor{darkgreen}{Text-Image-to-Video}}
        }
    }
    \\[1em]

    \centering
    \textbf{Prompt:} \textit{The video features a central spacecraft with a predominantly white and gray color scheme, accented with red and black details. It has a sleek, angular design with multiple protruding elements that suggest advanced technology...} \\[0.5em]
    \begin{minipage}{\textwidth}
        \centering
        \begin{minipage}[t]{0.18\textwidth}
            \centering
            \fcolorbox{blue}{white}{\includegraphics[width=\linewidth]{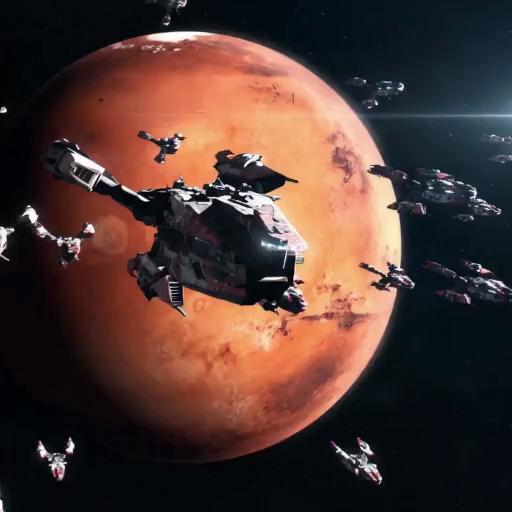}}
            \\[0.5em]
            \textbf{Reference Image}
        \end{minipage}
        \hfill
        \includegraphics[width=0.18\textwidth]{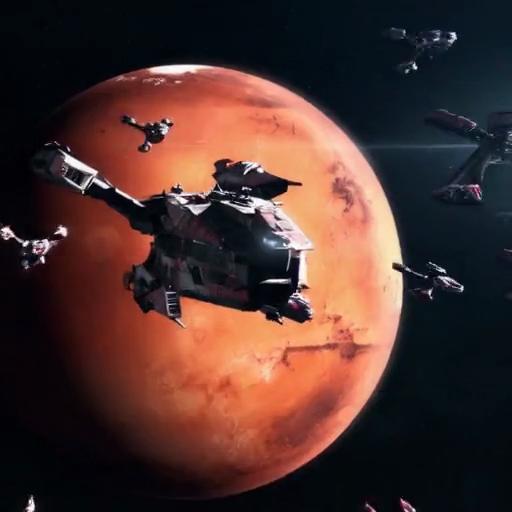}
        \includegraphics[width=0.18\textwidth]{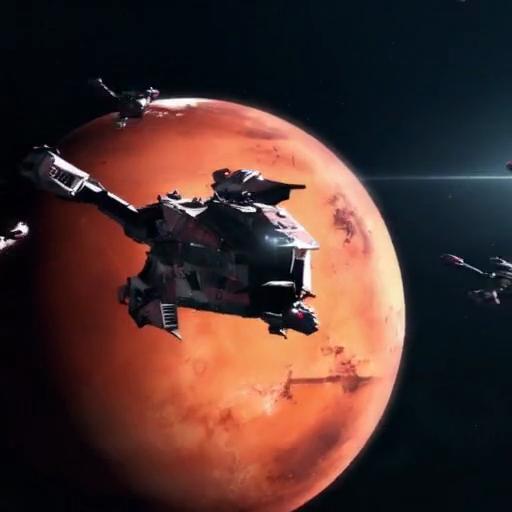}
        \includegraphics[width=0.18\textwidth]{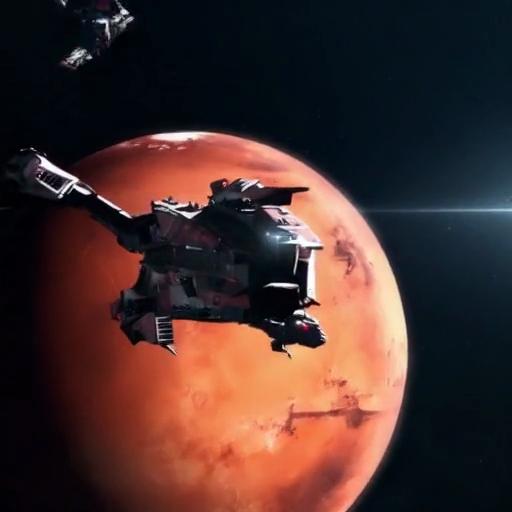}
        \includegraphics[width=0.18\textwidth]{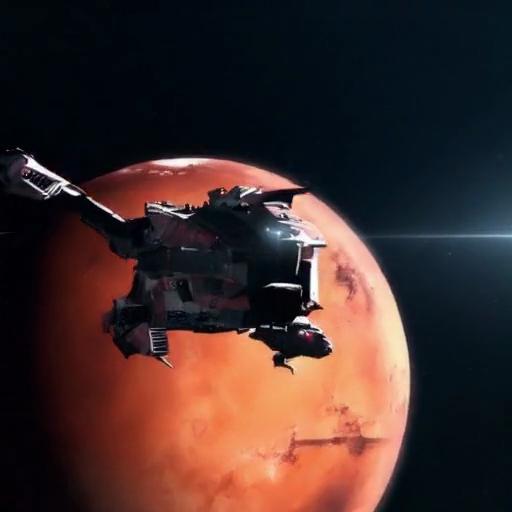}
    \end{minipage}
    \\[1em]

    \centering
    \textbf{Prompt:} \textit{The video begins with a dark space background, dotted with stars, and a central object that appears to be a spacecraft with a glowing blue light at its core. The spacecraft is detailed with various components...} \\[0.5em]
    \begin{minipage}{\textwidth}
        \centering
        \begin{minipage}[t]{0.18\textwidth}
            \centering
            \fcolorbox{blue}{white}{\includegraphics[width=\linewidth]{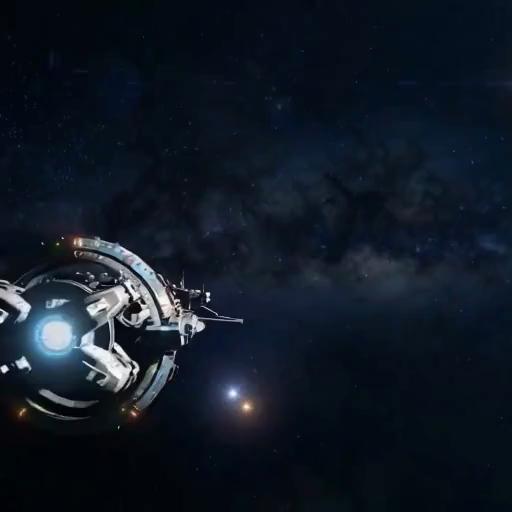}}
            \\[0.5em]
            \textbf{Reference Image}
        \end{minipage}
        \hfill
        \includegraphics[width=0.18\textwidth]{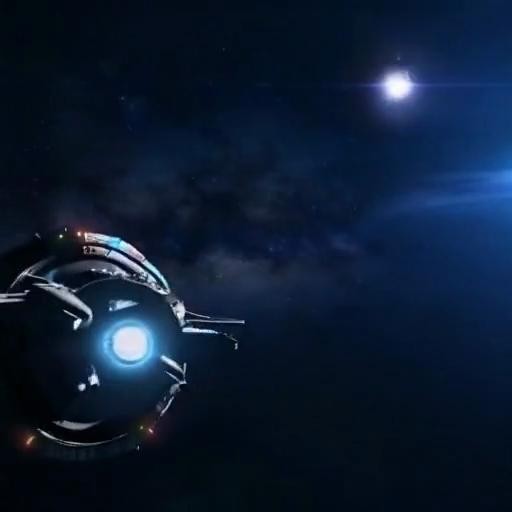}
        \includegraphics[width=0.18\textwidth]{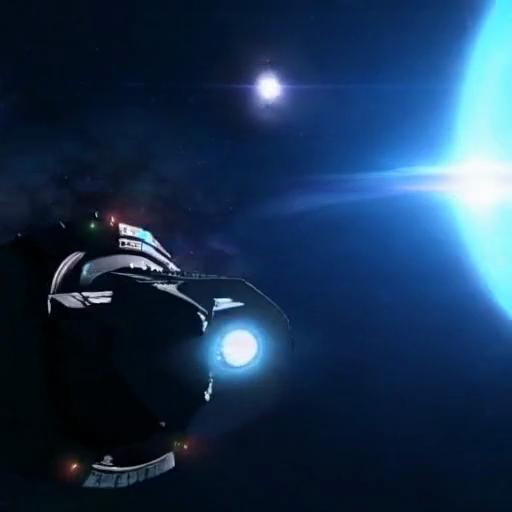}
        \includegraphics[width=0.18\textwidth]{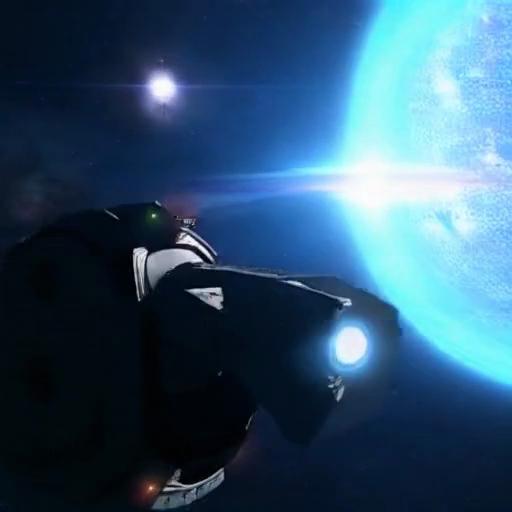}
        \includegraphics[width=0.18\textwidth]{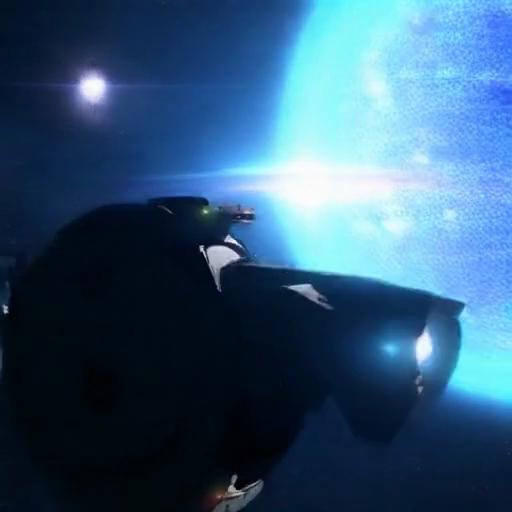}
    \end{minipage}
    \\[1em]

    \textbf{Prompt:} \textit{Robots move efficiently through a futuristic laboratory, adjusting holographic displays and conducting experiments, while scientists observe and interact with the high-tech equipment.} \\[0.5em]
    \begin{minipage}{\textwidth}
        \centering
        \begin{minipage}[t]{0.18\textwidth}
            \centering
            \fcolorbox{blue}{white}{\includegraphics[width=\linewidth]{ti2v_figures/0220/frame_000000.jpg}}
            \\[0.5em]
            \textbf{Reference Image}
        \end{minipage}
        \hfill
        \includegraphics[width=0.18\textwidth]{ti2v_figures/0220/frame_000005.jpg}
        \includegraphics[width=0.18\textwidth]{ti2v_figures/0220/frame_000010.jpg}
        \includegraphics[width=0.18\textwidth]{ti2v_figures/0220/frame_000015.jpg}
        \includegraphics[width=0.18\textwidth]{ti2v_figures/0220/frame_000019.jpg}
    \end{minipage}

    \centering
    \textbf{Prompt:} \textit{The video presents a serene scene with a group of camels walking in a line across a desert landscape. The camels are adorned with colorful saddles and are led by a person wearing a green garment. The background features a clear sky...} \\[0.5em]
    \begin{minipage}{\textwidth}
        \centering
        \begin{minipage}[t]{0.18\textwidth}
            \centering
            \fcolorbox{blue}{white}{\includegraphics[width=\linewidth]{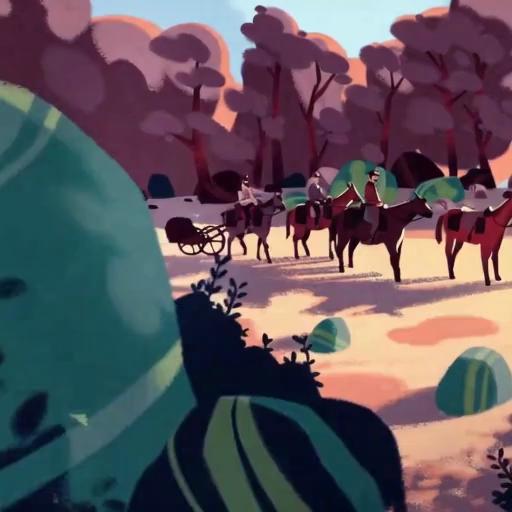}}
            \\[0.5em]
            \textbf{Reference Image}
        \end{minipage}
        \hfill
        \includegraphics[width=0.18\textwidth]{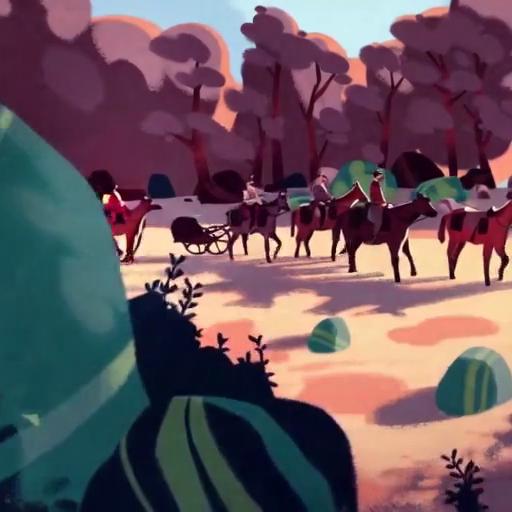}
        \includegraphics[width=0.18\textwidth]{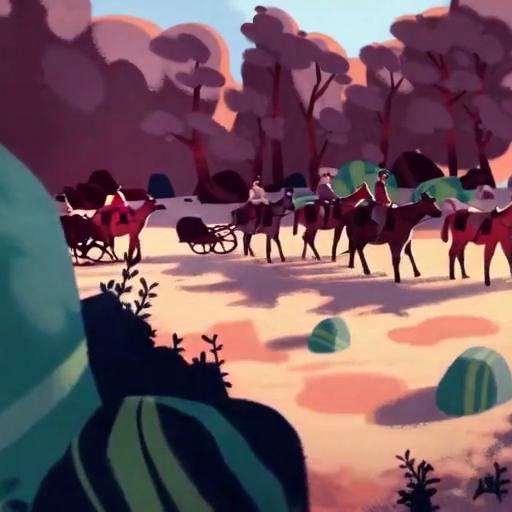}
        \includegraphics[width=0.18\textwidth]{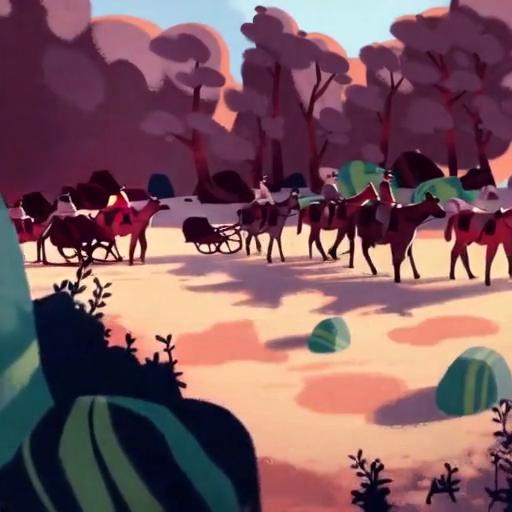}
        \includegraphics[width=0.18\textwidth]{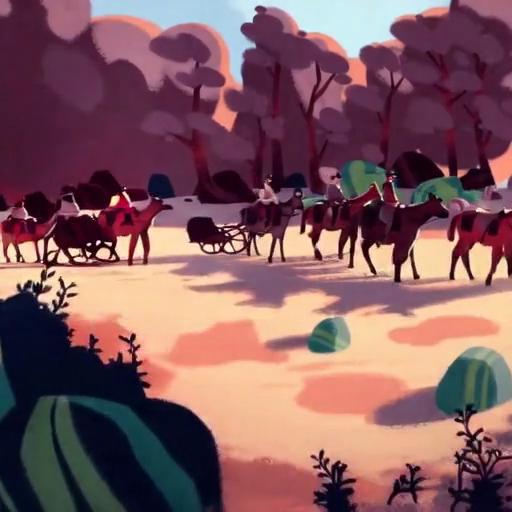}
    \end{minipage}
    \\[1em]
\end{figure*}

\begin{figure*}[htbp]
    \centering
    \centering
    \fcolorbox{darkgreen}{lightcream}{
        \parbox{\textwidth}{
            \centering \textbf{\Large \textcolor{darkgreen}{Text-Image-to-Video}}
        }
    }
    \\[1em]

    \centering
    \textbf{Prompt:} \textit{A crab made of different jewlery is walking on the beach. As it walks, it drops different jewelry pieces like diamonds, pearls, etc.} \\[0.5em]
    \begin{minipage}{\textwidth}
        \centering
        \begin{minipage}[t]{0.18\textwidth}
            \centering
            \fcolorbox{blue}{white}{\includegraphics[width=\linewidth]{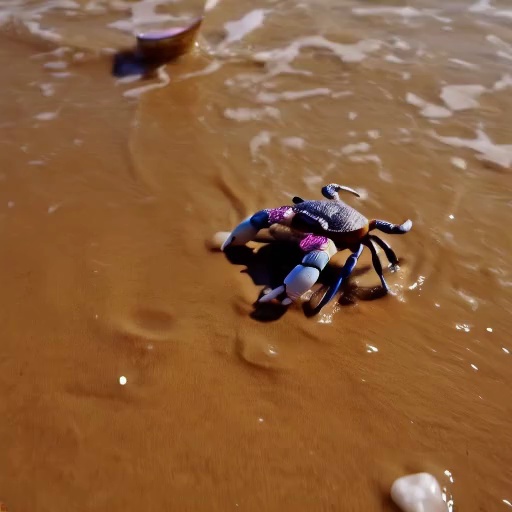}}
            \\[0.5em]
            \textbf{Reference Image}
        \end{minipage}
        \hfill
        \includegraphics[width=0.18\textwidth]{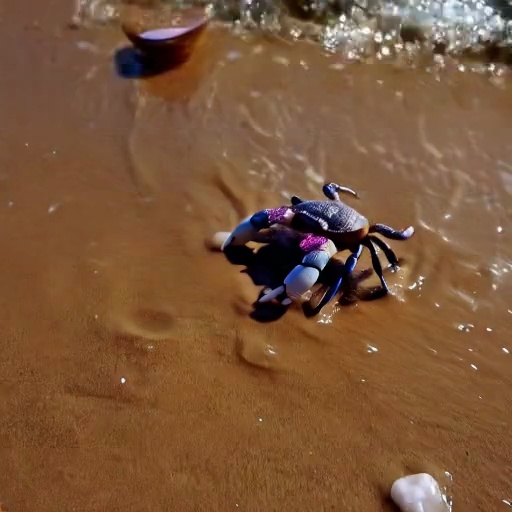}
        \includegraphics[width=0.18\textwidth]{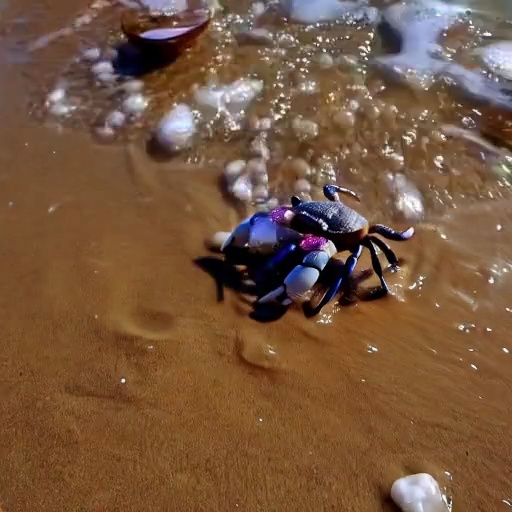}
        \includegraphics[width=0.18\textwidth]{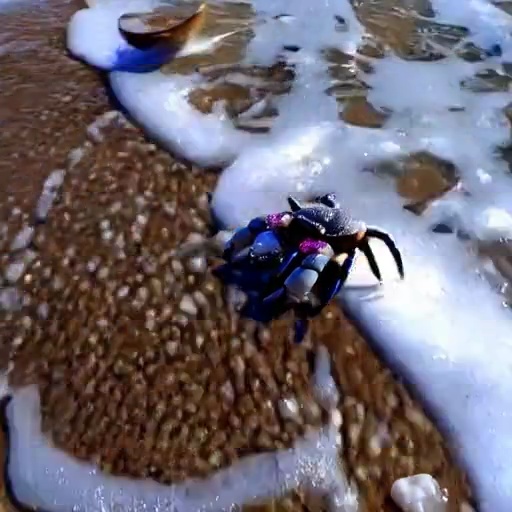}
        \includegraphics[width=0.18\textwidth]{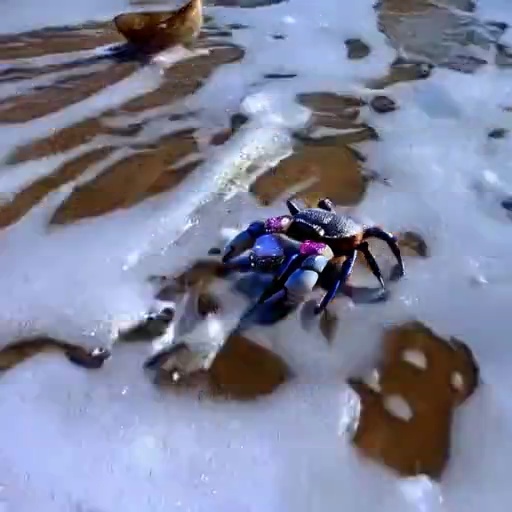}
    \end{minipage}
    \\[1em]

    \centering
    \textbf{Prompt:} \textit{The video captures a single sea turtle with a patterned shell and flippers, swimming in a clear blue underwater environment. The turtle moves gracefully over a bed of coral reefs, which exhibit a variety of colors...} \\[0.5em]
    \begin{minipage}{\textwidth}
        \centering
        \begin{minipage}[t]{0.18\textwidth}
            \centering
            \fcolorbox{blue}{white}{\includegraphics[width=\linewidth]{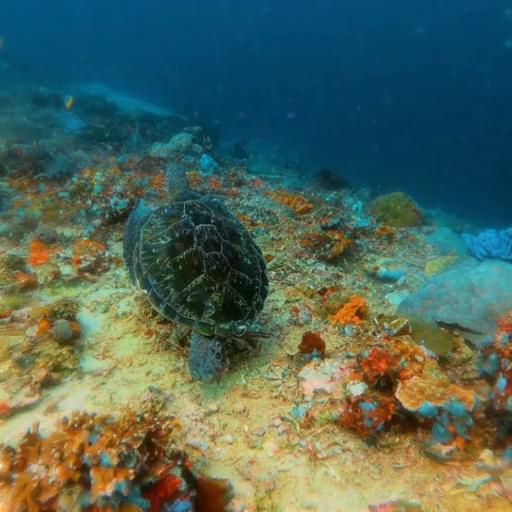}}
            \\[0.5em]
            \textbf{Reference Image}
        \end{minipage}
        \hfill
        \includegraphics[width=0.18\textwidth]{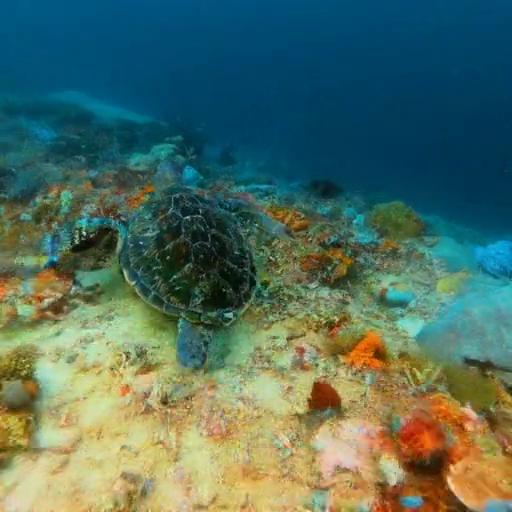}
        \includegraphics[width=0.18\textwidth]{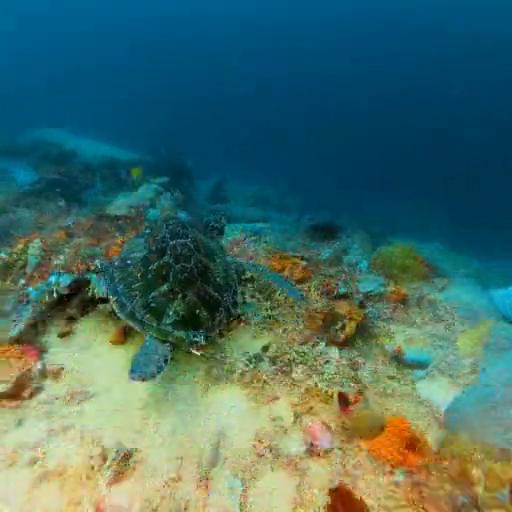}
        \includegraphics[width=0.18\textwidth]{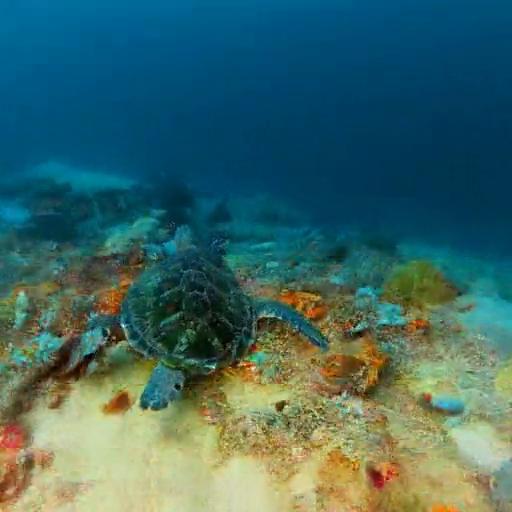}
        \includegraphics[width=0.18\textwidth]{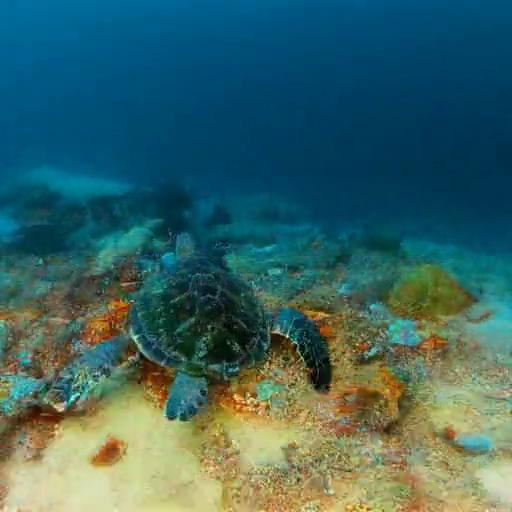}
    \end{minipage}
    \\[1em]

    \centering
    \textbf{Prompt:} \textit{A mesmerizing video of a jellyfish moving through water, with its tentacles flowing gracefully.} \\[0.5em]
    \begin{minipage}{\textwidth}
        \centering
        \begin{minipage}[t]{0.18\textwidth}
            \centering
            \fcolorbox{blue}{white}{\includegraphics[width=\linewidth]{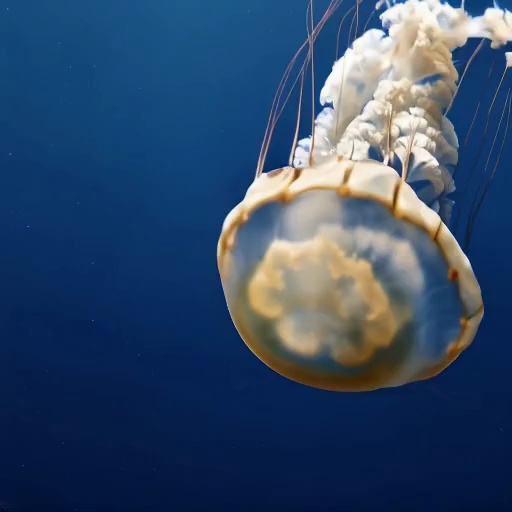}}
            \\[0.5em]
            \textbf{Reference Image}
        \end{minipage}
        \hfill
        \includegraphics[width=0.18\textwidth]{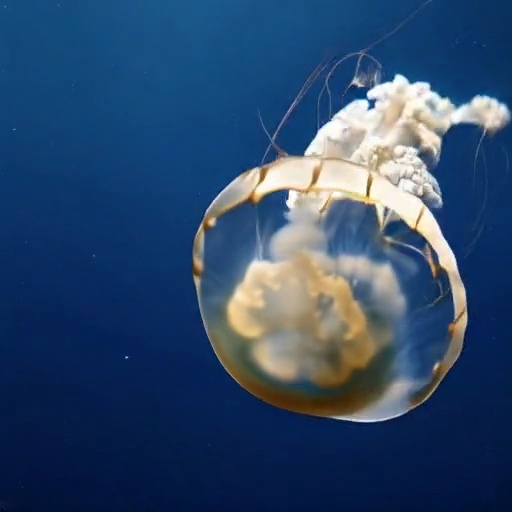}
        \includegraphics[width=0.18\textwidth]{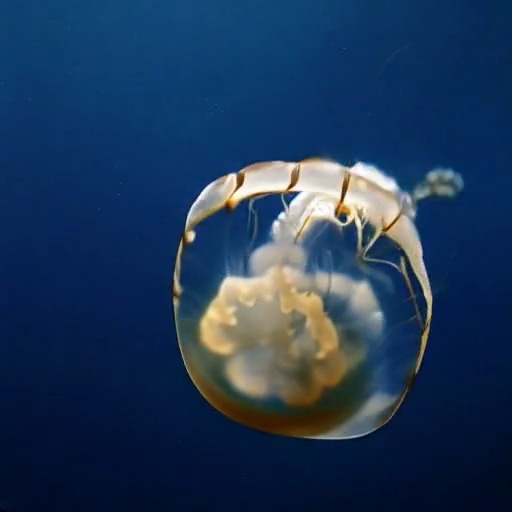}
        \includegraphics[width=0.18\textwidth]{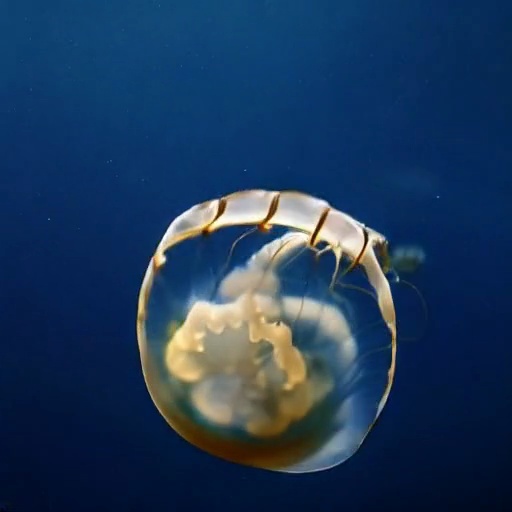}
        \includegraphics[width=0.18\textwidth]{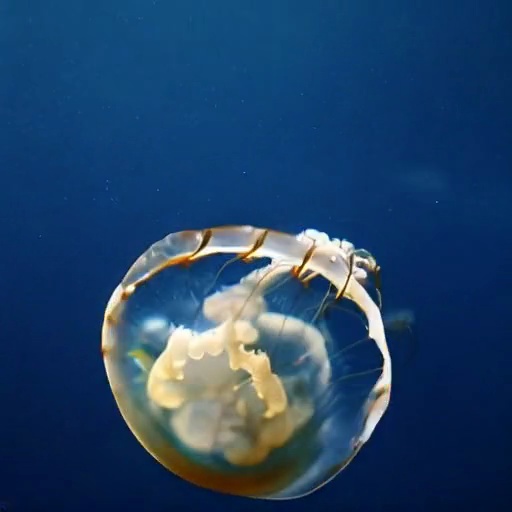}
    \end{minipage}
    \\[1em]


    \centering
    \textbf{Prompt:} \textit{A video of a diver creating bubbles underwater, with bubbles rising and interacting with each other.} \\[0.5em]
    \begin{minipage}{\textwidth}
        \centering
        \begin{minipage}[t]{0.18\textwidth}
            \centering
            \fcolorbox{blue}{white}{\includegraphics[width=\linewidth]{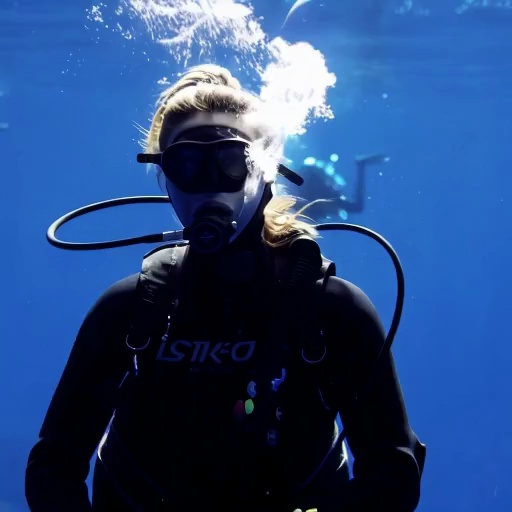}}
            \\[0.5em]
            \textbf{Reference Image}
        \end{minipage}
        \hfill
        \includegraphics[width=0.18\textwidth]{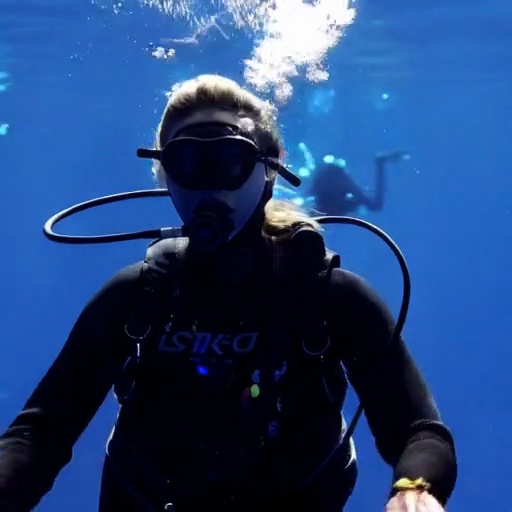}
        \includegraphics[width=0.18\textwidth]{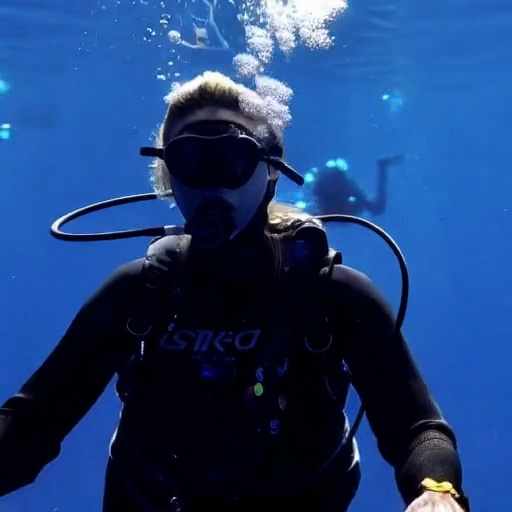}
        \includegraphics[width=0.18\textwidth]{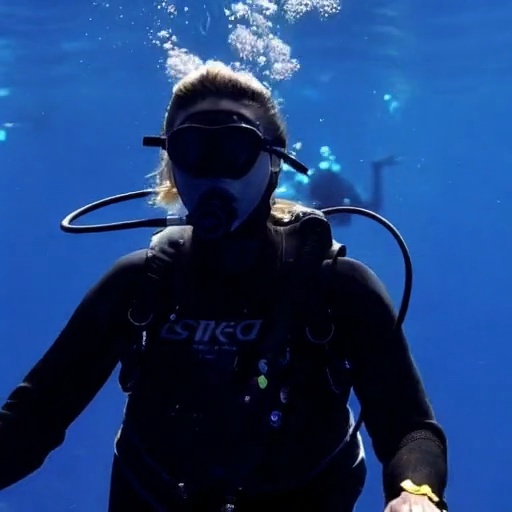}
        \includegraphics[width=0.18\textwidth]{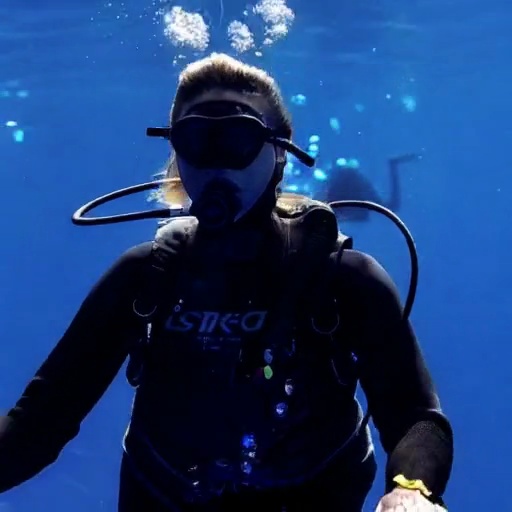}
    \end{minipage}
    \\[1em]

\end{figure*}

\begin{figure*}[htbp]
    \centering
    \centering
    \fcolorbox{darkgreen}{lightcream}{
        \parbox{\textwidth}{
            \centering \textbf{\Large \textcolor{darkgreen}{Text-Image-to-Video}}
        }
    }
    \\[1em]

    \centering
    \textbf{Prompt:} \textit{The individual in the video is dressed in a blue protective suit with a hood, a mask with a filter, and white gloves. They are holding a spray bottle in one hand and a spray nozzle in the other...} \\[0.5em]
    \begin{minipage}{\textwidth}
        \centering
        \begin{minipage}[t]{0.18\textwidth}
            \centering
            \fcolorbox{blue}{white}{\includegraphics[width=\linewidth]{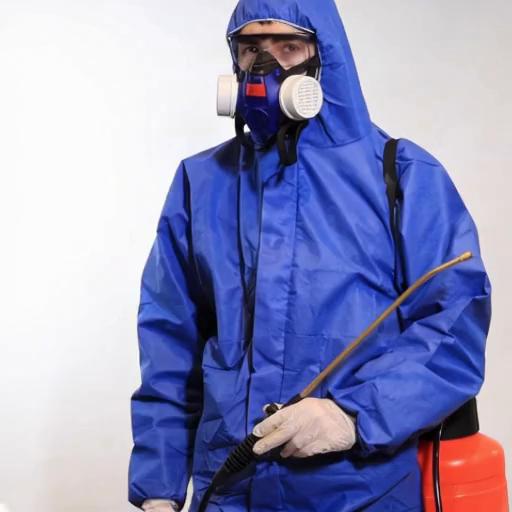}}
            \\[0.5em]
            \textbf{Reference Image}
        \end{minipage}
        \hfill
        \includegraphics[width=0.18\textwidth]{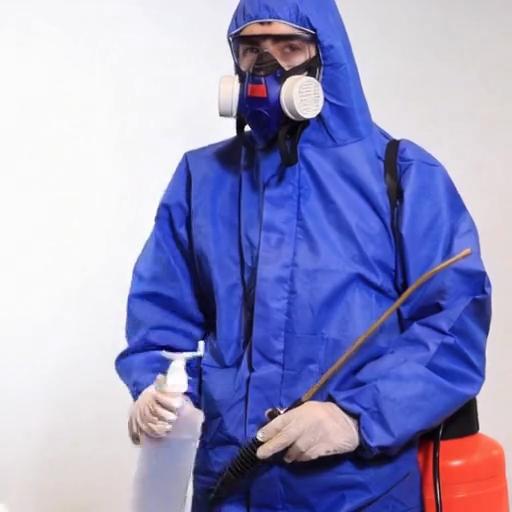}
        \includegraphics[width=0.18\textwidth]{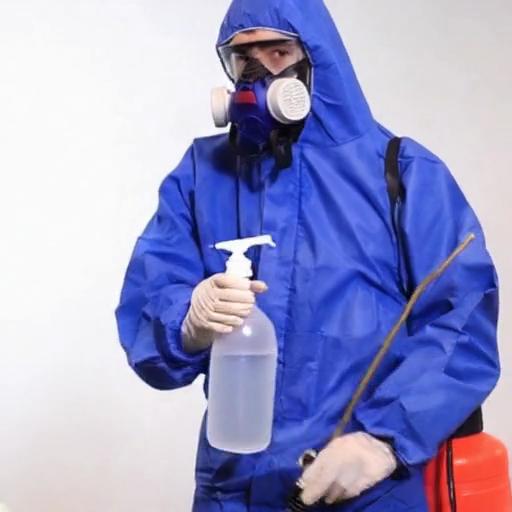}
        \includegraphics[width=0.18\textwidth]{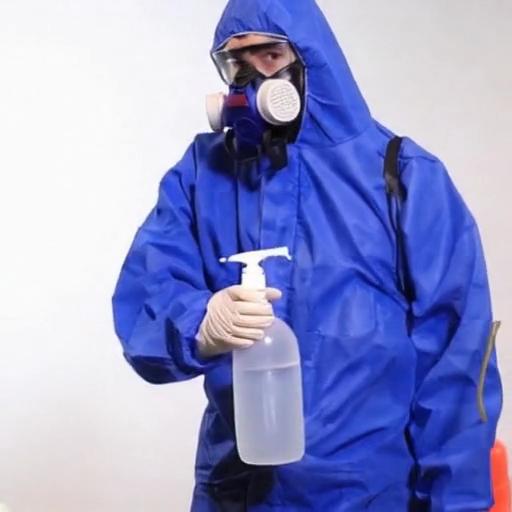}
        \includegraphics[width=0.18\textwidth]{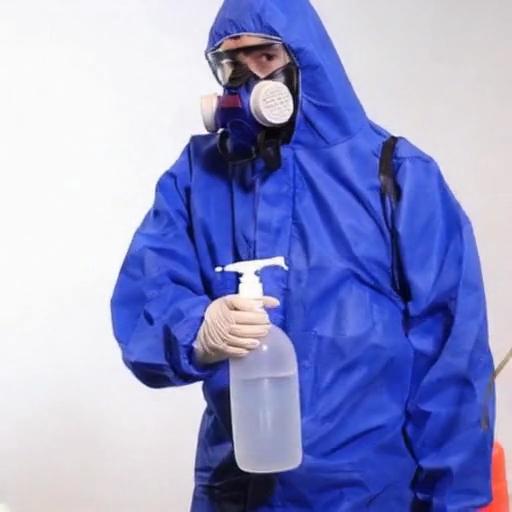}
    \end{minipage}
    \\[1em]
    
    \centering
    \textbf{Prompt:} \textit{The video captures a bustling city street scene during the evening. The sky is overcast, and the street is wet, reflecting the lights from the vehicles and buildings. The buildings are tall with modern architecture...} \\[0.5em]
    \begin{minipage}{\textwidth}
        \centering
        \begin{minipage}[t]{0.18\textwidth}
            \centering
            \fcolorbox{blue}{white}{\includegraphics[width=\linewidth]{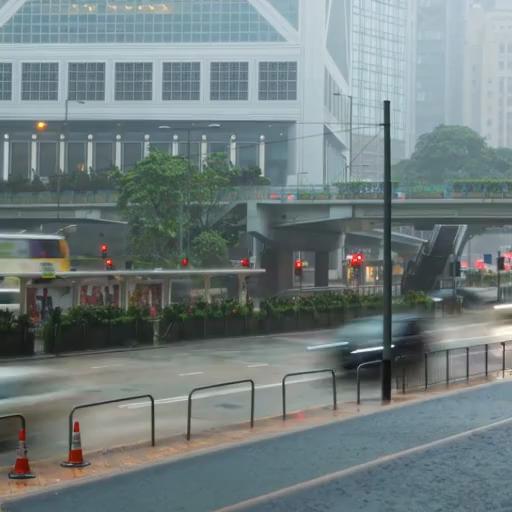}}
            \\[0.5em]
            \textbf{Reference Image}
        \end{minipage}
        \hfill
        \includegraphics[width=0.18\textwidth]{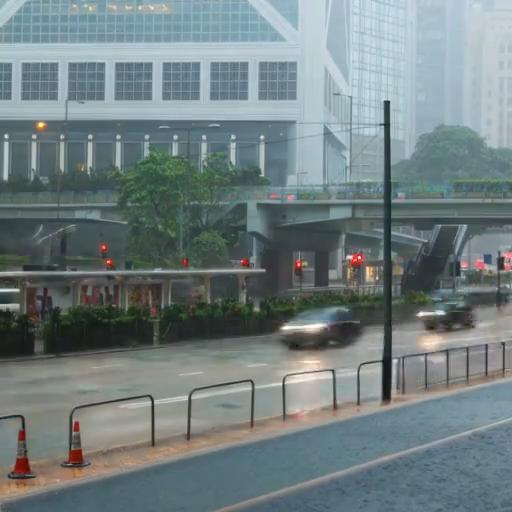}
        \includegraphics[width=0.18\textwidth]{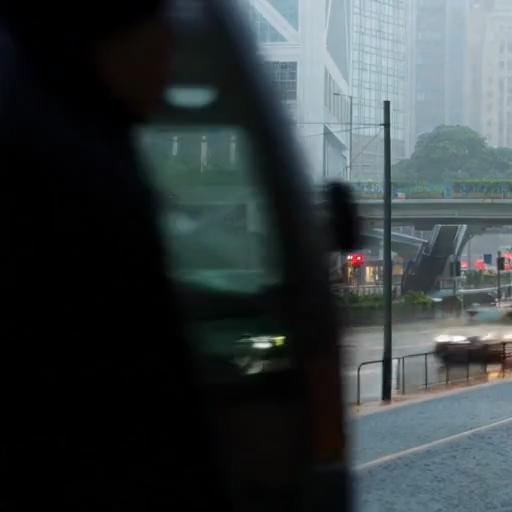}
        \includegraphics[width=0.18\textwidth]{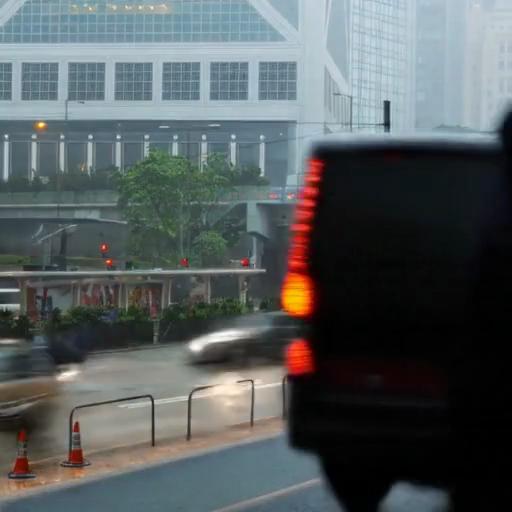}
        \includegraphics[width=0.18\textwidth]{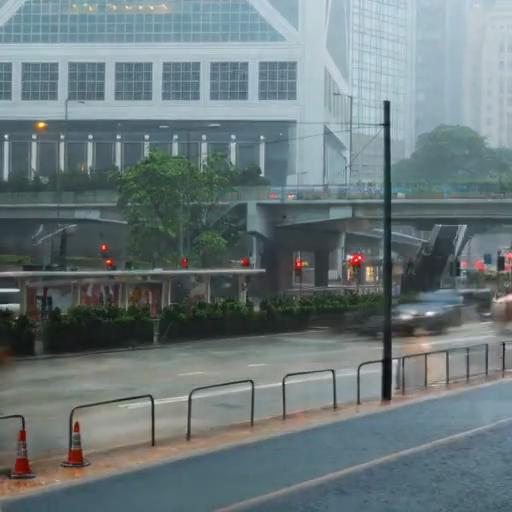}
    \end{minipage}
    \\[1em]

    \centering
    \textbf{Prompt:} \textit{The video presents a series of images capturing the Colosseum from an aerial perspective during the evening. The ancient amphitheater is illuminated by artificial lighting, which highlights its circular shape and the arches...} \\[0.5em]
    \begin{minipage}{\textwidth}
        \centering
        \begin{minipage}[t]{0.18\textwidth}
            \centering
            \fcolorbox{blue}{white}{\includegraphics[width=\linewidth]{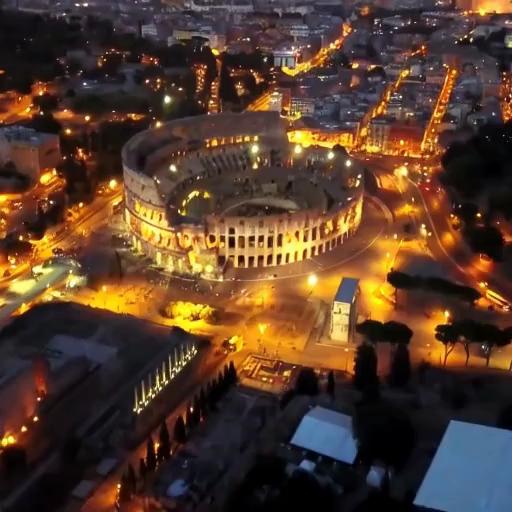}}
            \\[0.5em]
            \textbf{Reference Image}
        \end{minipage}
        \hfill
        \includegraphics[width=0.18\textwidth]{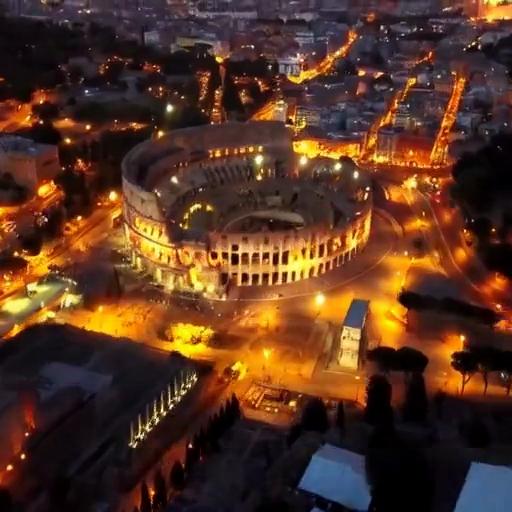}
        \includegraphics[width=0.18\textwidth]{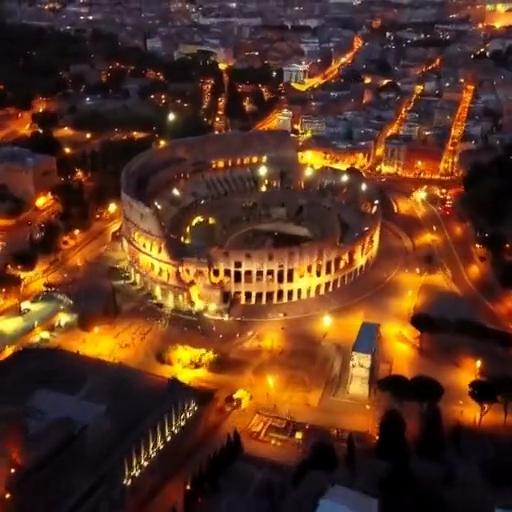}
        \includegraphics[width=0.18\textwidth]{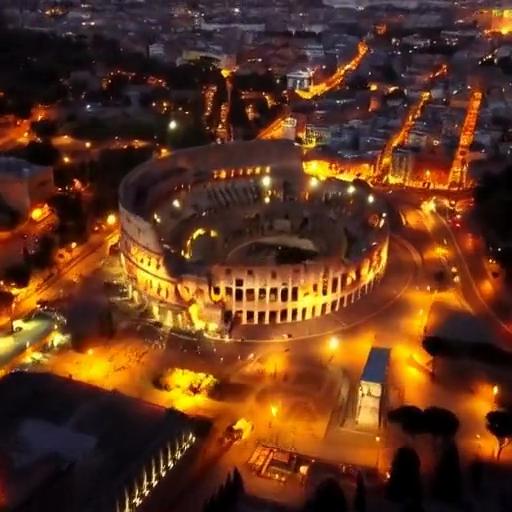}
        \includegraphics[width=0.18\textwidth]{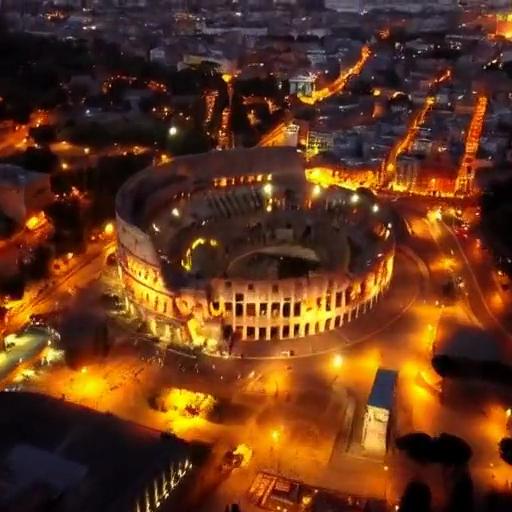}
    \end{minipage}
    \\[1em]

    \centering
    \textbf{Prompt:} \textit{The video features two dogs, one with a predominantly white coat and the other with a mix of black, brown, and white fur. Both dogs are adorned with accessories; the white dog wears a red tie, while the other sports a purple bow tie...} \\[0.5em]
    \begin{minipage}{\textwidth}
        \centering
        \begin{minipage}[t]{0.18\textwidth}
            \centering
            \fcolorbox{blue}{white}{\includegraphics[width=\linewidth]{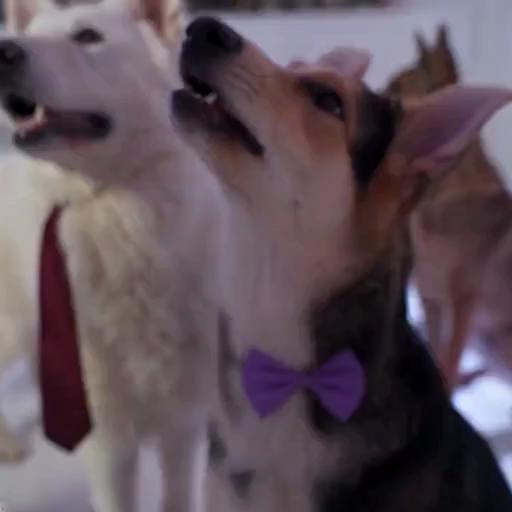}}
            \\[0.5em]
            \textbf{Reference Image}
        \end{minipage}
        \hfill
        \includegraphics[width=0.18\textwidth]{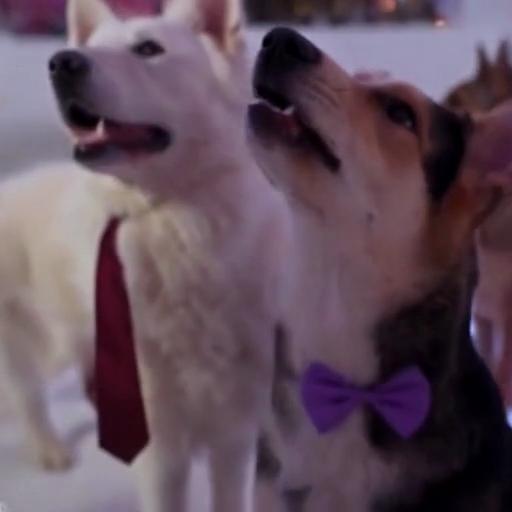}
        \includegraphics[width=0.18\textwidth]{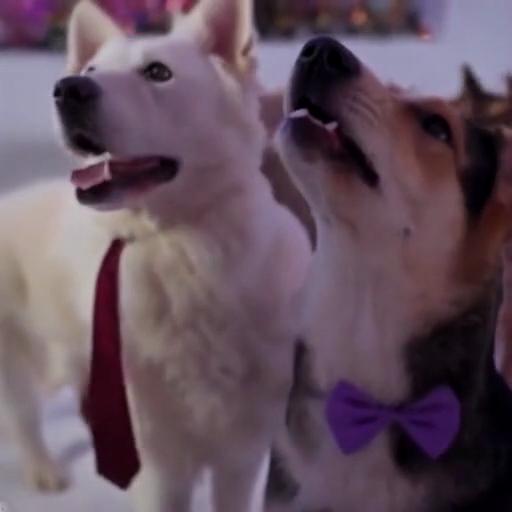}
        \includegraphics[width=0.18\textwidth]{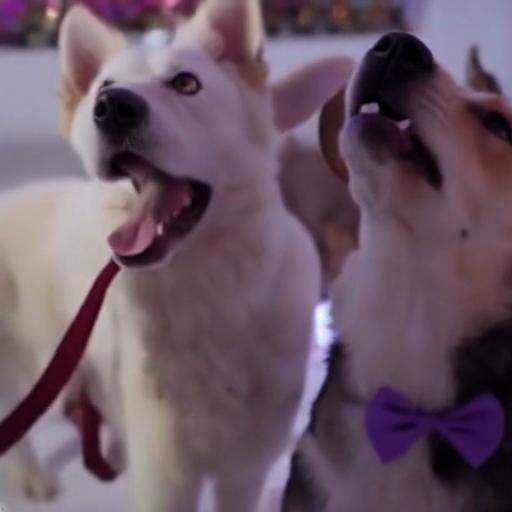}
        \includegraphics[width=0.18\textwidth]{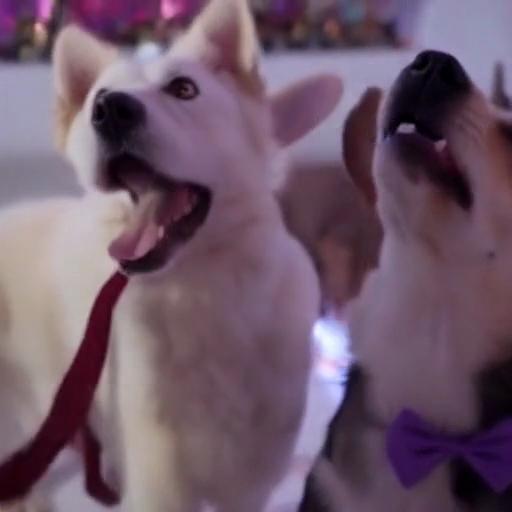}
    \end{minipage}
    \\[1em]
\end{figure*}

\end{document}